\documentclass{article}

\usepackage{subfig}

\usepackage{hyperref}
\usepackage[accepted]{icml2024}

\usepackage{mathtools}
\usepackage[capitalize,noabbrev]{cleveref}

\usepackage{url}

\usepackage{algorithm}
\usepackage{algorithmic}

\usepackage{multirow}
\usepackage{float}
\usepackage{multicol}
\usepackage{emptypage}
\usepackage{changepage}

\usepackage{booktabs}
\usepackage[switch]{lineno}
\usepackage{multirow}
\usepackage{graphicx}
\usepackage{xcolor}

\usepackage{nicefrac}       % compact symbols for 1/2, etc.
\usepackage{microtype}      % microtypography
\usepackage{tcolorbox}
\usepackage{amsmath,amssymb,amsfonts}
\usepackage{pifont}

\usepackage{overpic}
\usepackage{array} % for the 'm' column specifier
\usepackage{tabularx}
\usepackage{ltablex}
\newcolumntype{b}{X}
\newcolumntype{s}{>{\hsize=2.0\hsize}X}

\usepackage[textsize=tiny]{todonotes}

\hypersetup{
    colorlinks=true,
    linkcolor=black,
    citecolor=black,
    urlcolor=black
}

\icmltitlerunning{Vision-Language Instruction Tuning for Semiconductor Electron Micrograph Analysis}

\begin{document}

\twocolumn[
\icmltitle{Parameter-Efficient Quantized Mixture-of-Experts Meets Vision-Language Instruction Tuning for Semiconductor Electron Micrograph Analysis}



\icmlsetsymbol{equal}{*}

\begin{icmlauthorlist}
\icmlauthor{Sakhinana Sagar Srinivas}{yyy}
\icmlauthor{Chidaksh Ravuru}{comp}
\icmlauthor{Geethan Sannidhi}{sch}
\icmlauthor{Venkataramana Runkana}{yyy}
\end{icmlauthorlist}

\icmlaffiliation{yyy}{TCS Research, Bangalore}
\icmlaffiliation{comp}{IIT Dharwad}
\icmlaffiliation{sch}{IIIT Pune}

\icmlcorrespondingauthor{Sakhinana Sagar Srinivas}{sagar.sakhinana@tcs.com}
\icmlkeywords{Machine Learning, ICML}

\vskip 0.3in
]



\printAffiliationsAndNotice{\icmlEqualContribution} % otherwise use the standard text.

\begin{abstract}
\vspace{-2mm}
Semiconductors, crucial to modern electronics, are generally under-researched in foundational models. It highlights the need for research to enhance the semiconductor device technology portfolio and aid in high-end device fabrication. In this paper, we introduce \texttt{sLAVA}, a small-scale vision-language assistant tailored for semiconductor manufacturing, with a focus on electron microscopy image analysis. It addresses challenges of data scarcity and acquiring high-quality, expert-annotated data. We employ a teacher-student paradigm, using a foundational vision-language model like GPT-4 as a teacher to create instruction-following multimodal data for customizing the student model, \texttt{sLAVA}, for  electron microscopic image analysis tasks on consumer hardware with limited budgets. Our approach allows enterprises to further fine-tune the proposed framework with their proprietary data securely within their own infrastructure, protecting intellectual property. Rigorous experiments validate that our framework surpasses traditional methods, handles data shifts, and enables high-throughput screening.
\vspace{-5mm}
\end{abstract}

\vspace{-5mm}
\section{Introduction} %  where accurate and trustworthy analysis of microscopic images is crucial for technological advancements.
\vspace{-2mm}
The semiconductor multi-step fabrication process is highly complex and involves specialized firms. Fabless chip designers like Apple, Qualcomm, and NVIDIA create complex integrated circuit designs but outsource manufacturing to foundries like TSMC and Samsung. Foundries use expensive, high-tech fabrication techniques, including photolithography and chemical vapor deposition, to produce intricate integrated circuits (ICs) on silicon wafers. The chips then undergo rigorous quality assurance, followed by packaging and assembly into devices such as microprocessors and memory chips. These semiconductor devices are then integrated into various electronic systems, such as consumer electronics and automotive technologies.
Sub-7nm technology marks a significant leap in chip miniaturization, enabling the creation of smaller, more powerful, and efficient devices. However, the industry faces challenges in achieving this miniaturization, such as strictly adhering to the design specifications and tolerances to consistently produce reliable, high-performance sub-7nm chips with minimal variation and high precision. Overcoming these challenges requires thorough testing using sophisticated imaging techniques and analysis to achieve high-quality, large-scale production of semiconductor chips. Advanced microscopy tools like Scanning Electron Microscopy (SEM) and Transmission Electron Microscopy (TEM) generate electron micrographs (nanoscale images), critical for quality control, failure analysis, and subsequent process optimization or design adjustments to help mitigate defects and ensures chips conform to specifications.
Current deep learning methods for characterizing materials are insufficient for the semiconductor industry's specialized needs for accurately analyzing electron micrographs. More effective technology is critical to support ongoing technological innovations. Recent advancements in Artificial Intelligence (AI), such as Large Multimodal Models (LMMs) like GPT-4 Turbo with Vision \cite{gpt4v}, Google Gemini\cite{team2023gemini} have the potential to impact semiconductor manufacturing by accurately analyzing microscopic images for various tasks, including zero/few-shot classification, image captioning, and visual question answering (VQA) tasks. GPT-4's combination of advanced natural language processing, image processing capabilities, and logical reasoning abilities could enable it to interpret and answer natural language questions about the microscopic images being analyzed. The insightful responses generated for end-user questions would allow human users to better evaluate the rationale behind GPT-4's image analysis and, consequently, trust its responses. Using proprietary multimodal vision-language models raises legitimate data privacy concerns, as intellectual property leaks could potentially undermine the cutting-edge technological portfolio of semiconductor firms and jeopardize future innovation. Additionally, their large size and complexity limit the adaptability to tailor them for specialized tasks. On the other hand, open-source, smaller models like LLaVA\cite{liu2023visual} and MiniGPT-4\cite{zhu2023minigpt} offer the benefits of customizable and interpretable microscopic image analysis of nanomaterials, but they may not match the reasoning and generalization capabilities of larger closed-source proprietary models.

\vspace{-2mm}
\begin{figure}[htbp]
     \centering
     \subfloat[High intra-class(within the same class) dissimilarity in electron micrographs of MEMS devices.]
     {\includegraphics[width=0.11\textwidth]{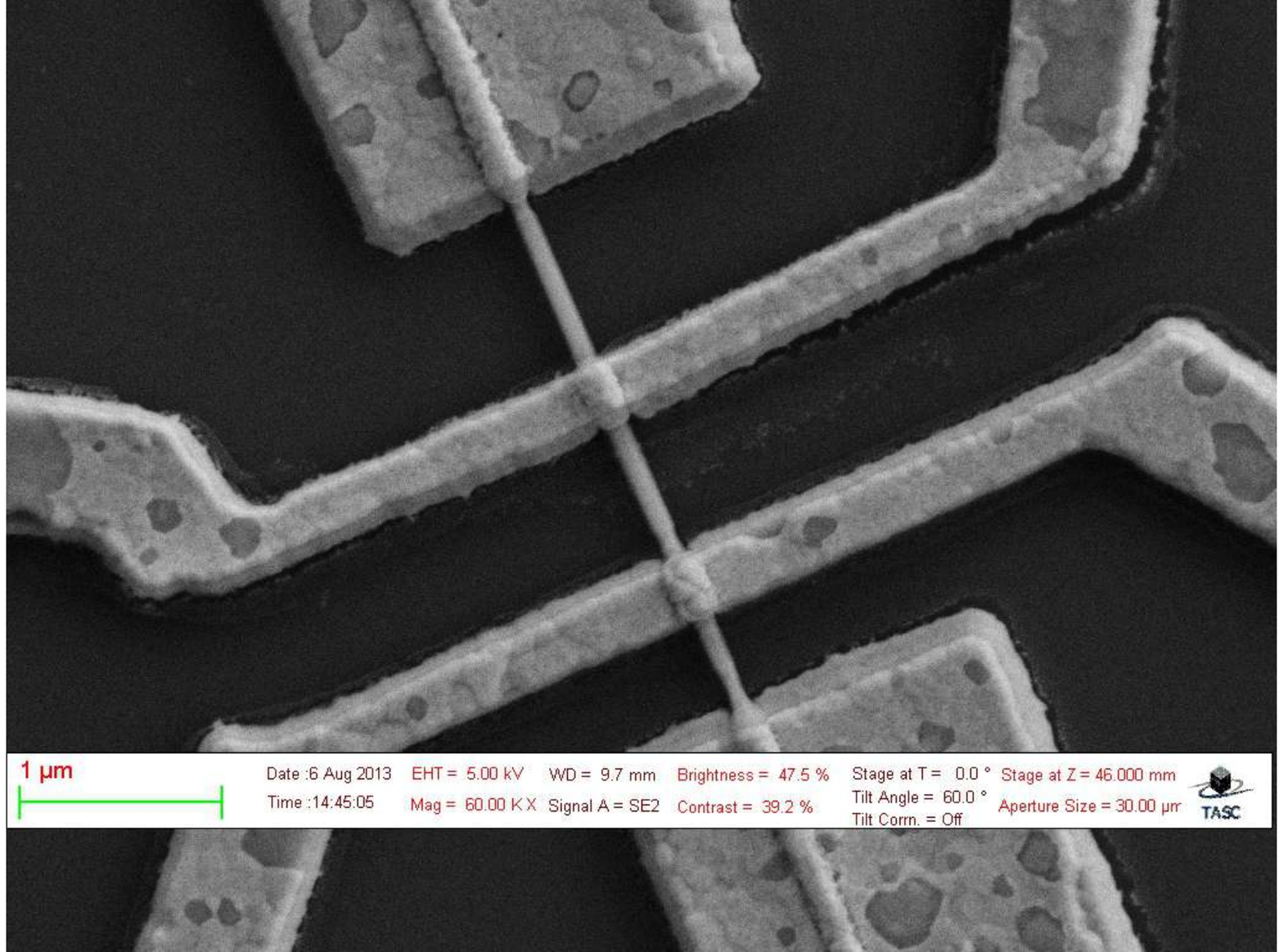}
     \includegraphics[width=0.11\textwidth]{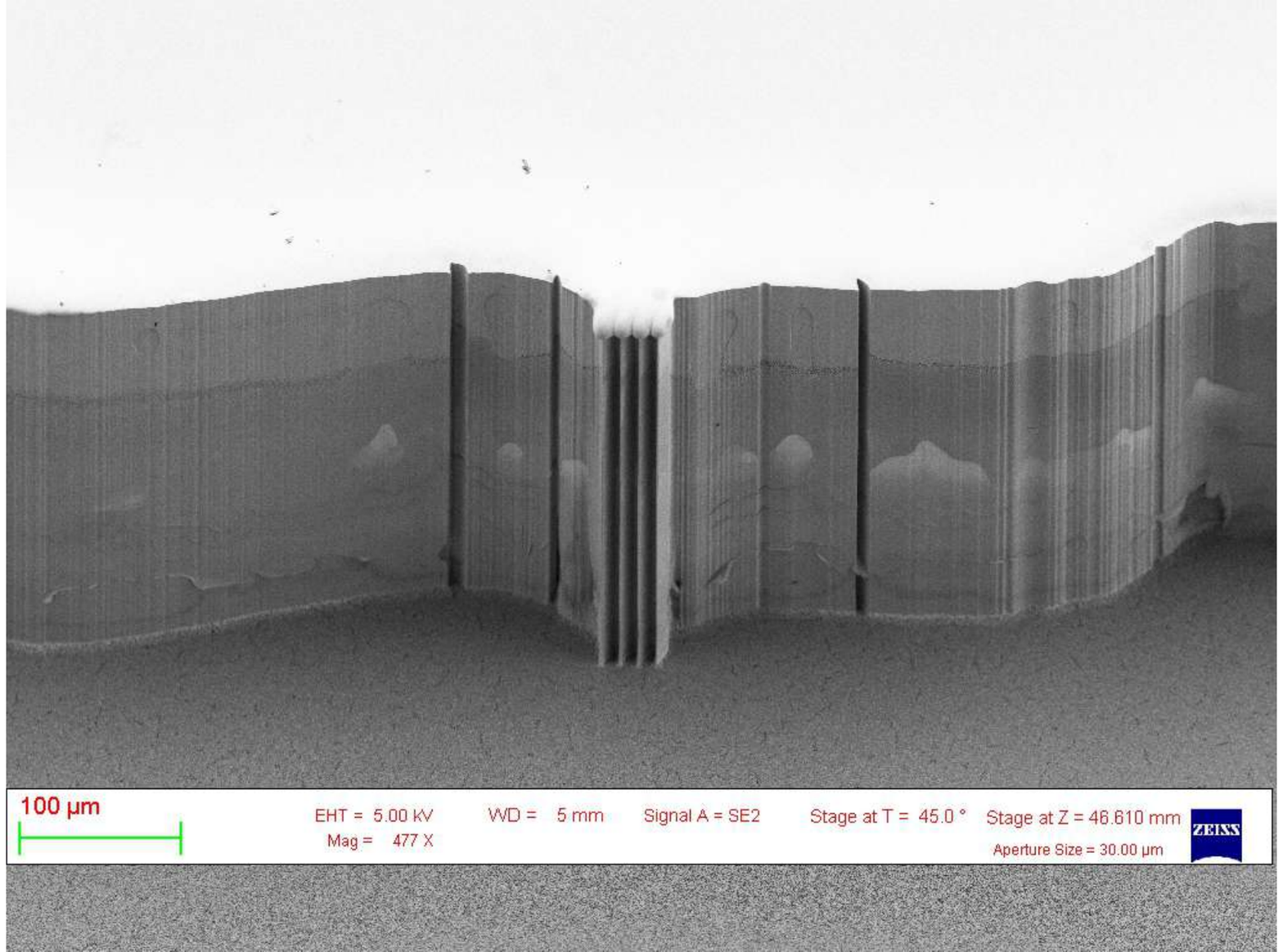}
     \includegraphics[width=0.11\textwidth]{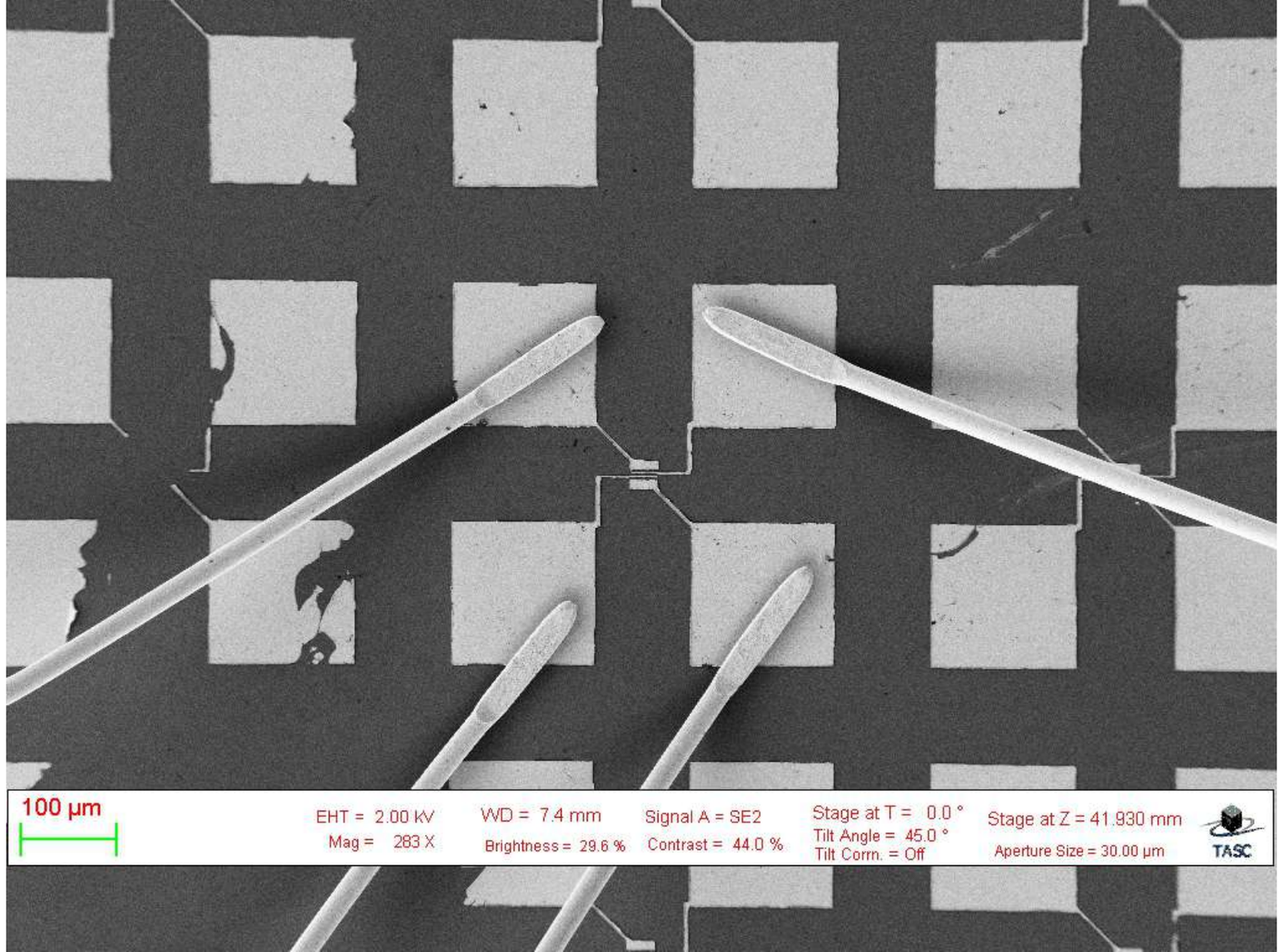}
     \includegraphics[width=0.11\textwidth]{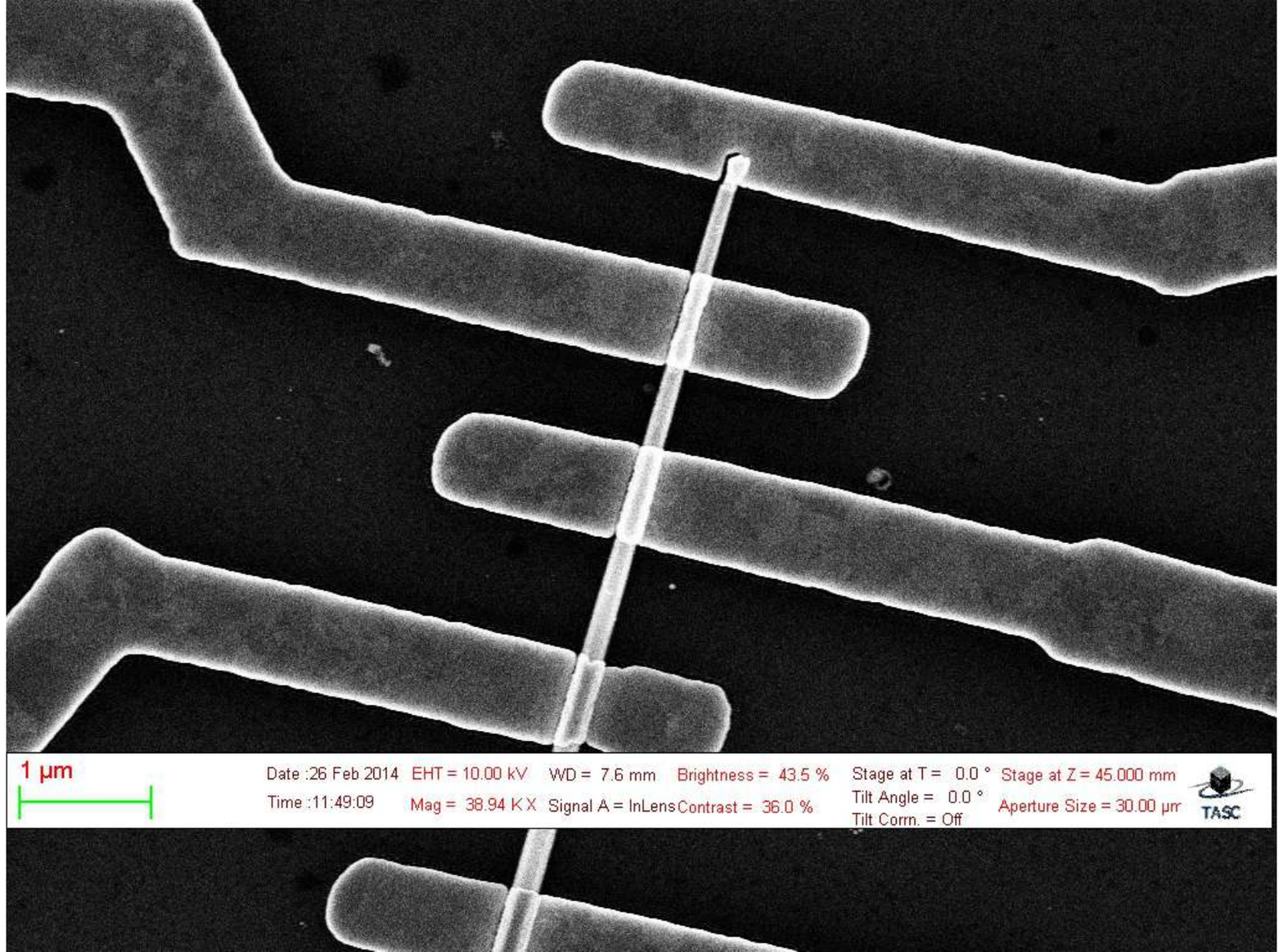}
     }
     \vspace{-3mm}
     \qquad
     \subfloat[High inter-class similarity(between classes): Electron micrographs of various nanomaterials (\textit{porous, particles, powders, films}) exhibit significant resemblance or high homogeneity.]{\includegraphics[width=0.11\textwidth]{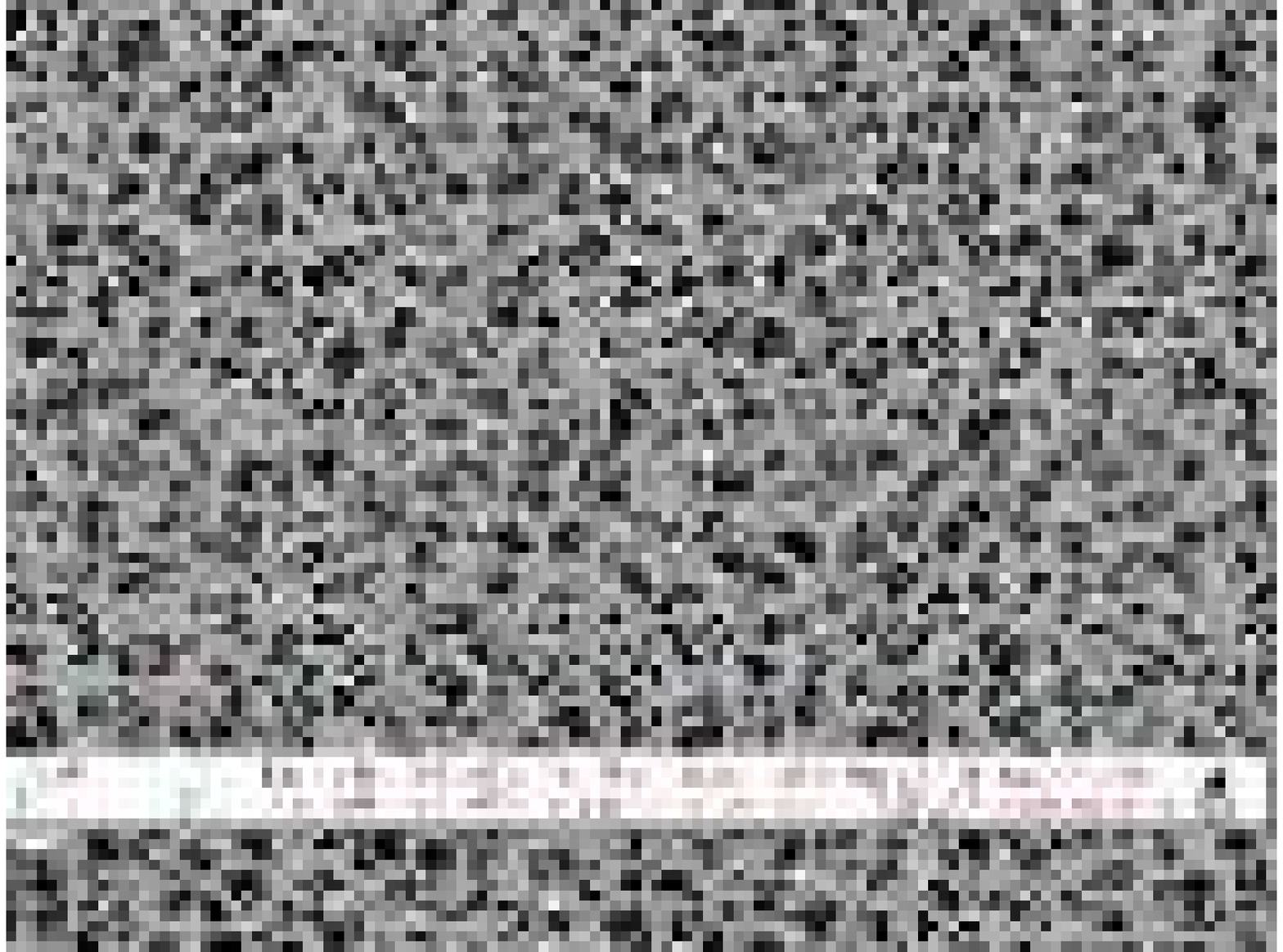}
     \includegraphics[width=0.11\textwidth]{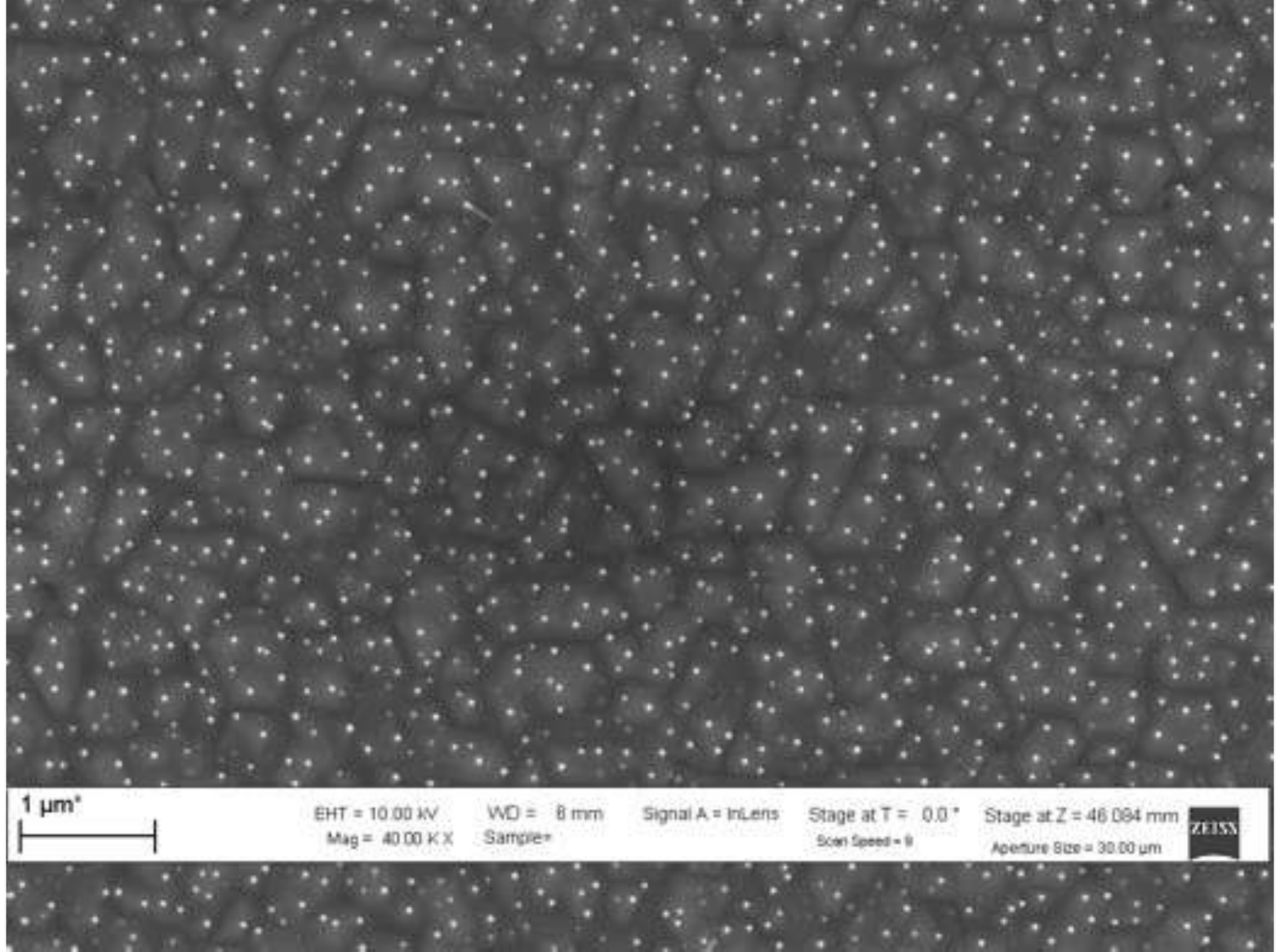}
     \includegraphics[width=0.11\textwidth]{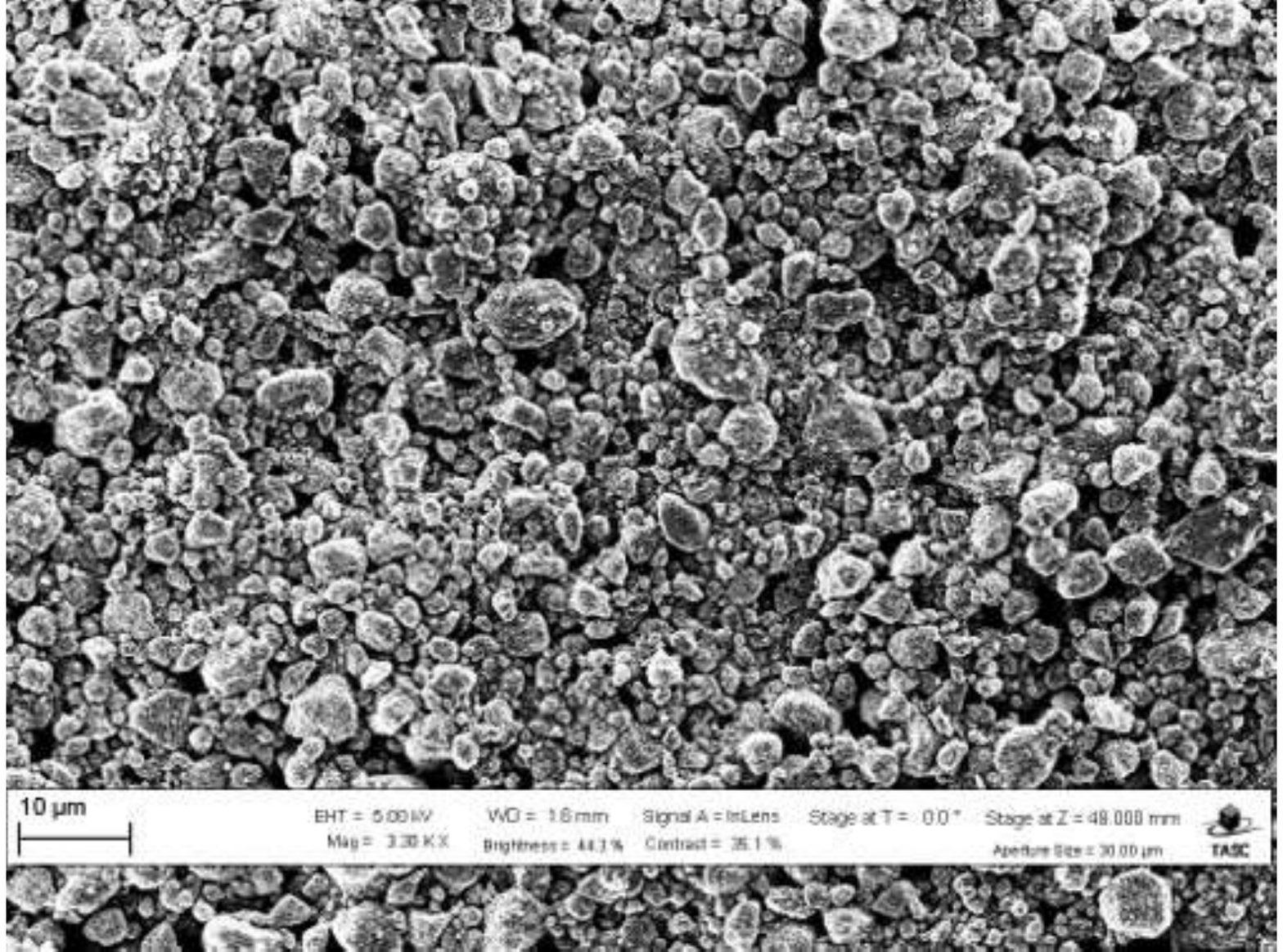}
     \includegraphics[width=0.11\textwidth]{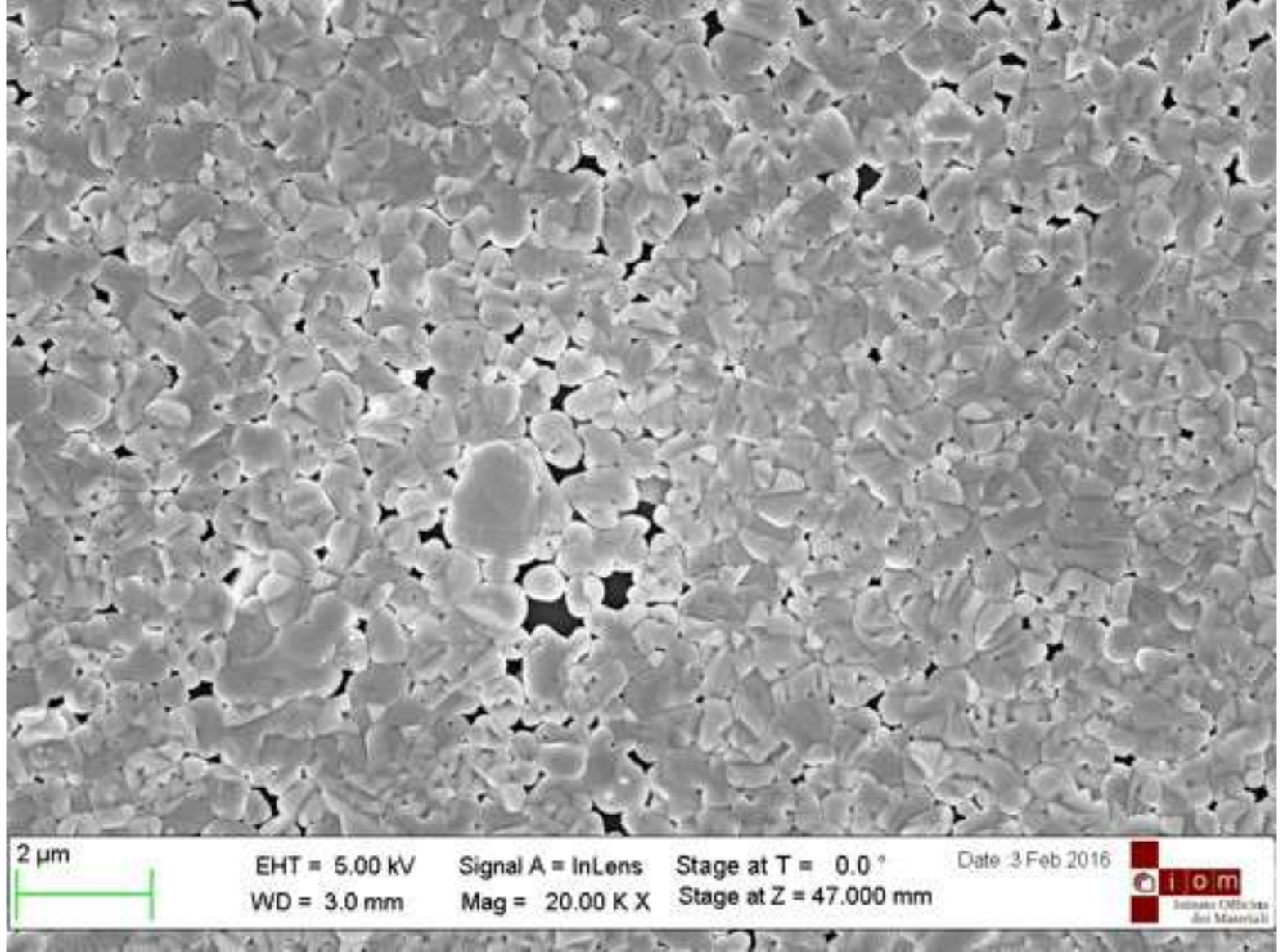}
     }
     \vspace{-3mm}
     \qquad
     \subfloat[Multi-scale patterns of spatial heterogeneity (e.g., size or shape variations) in nanoparticle micrographs.]
     {\includegraphics[width=0.11\textwidth]{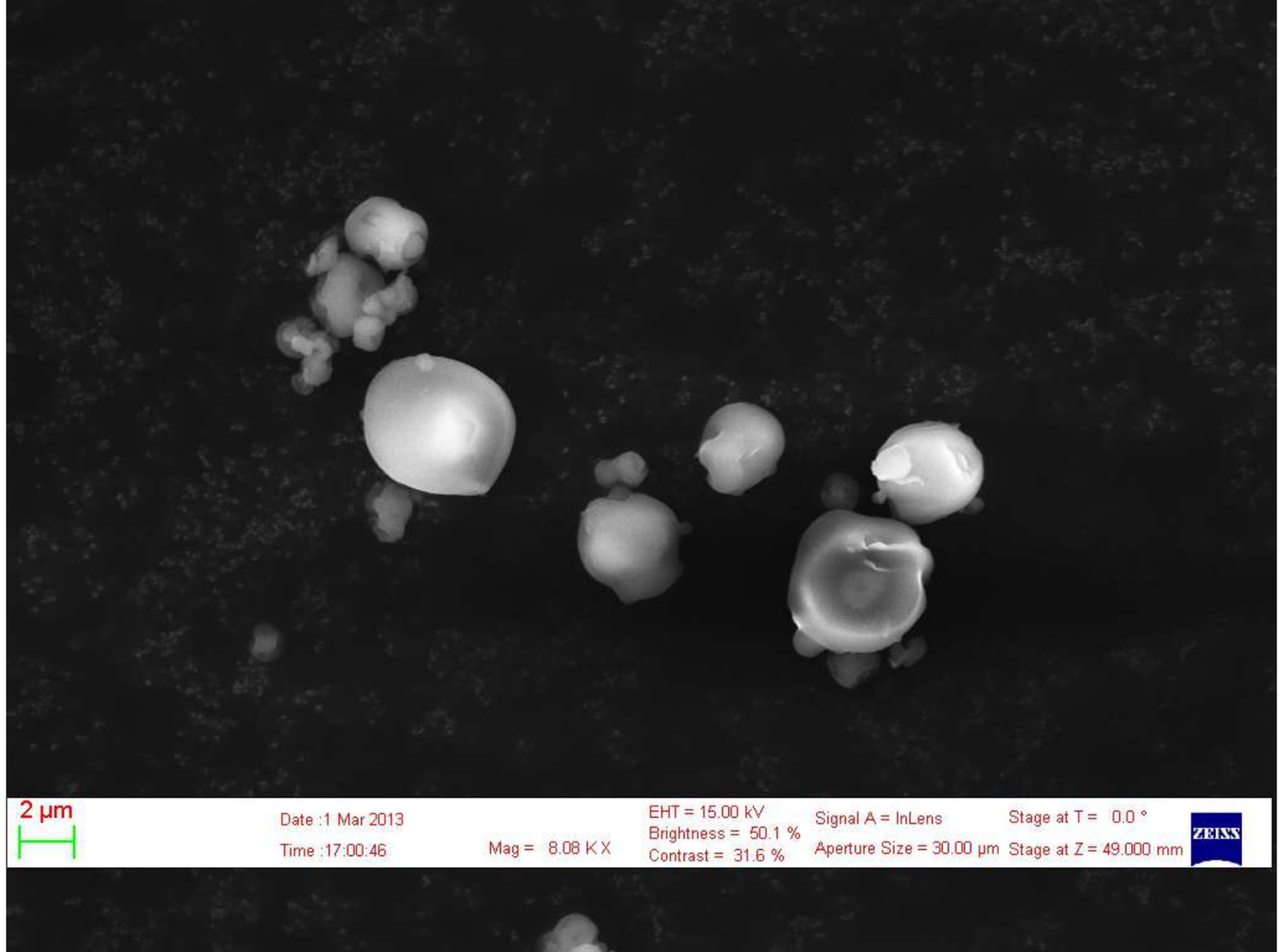}
     \includegraphics[width=0.11\textwidth]{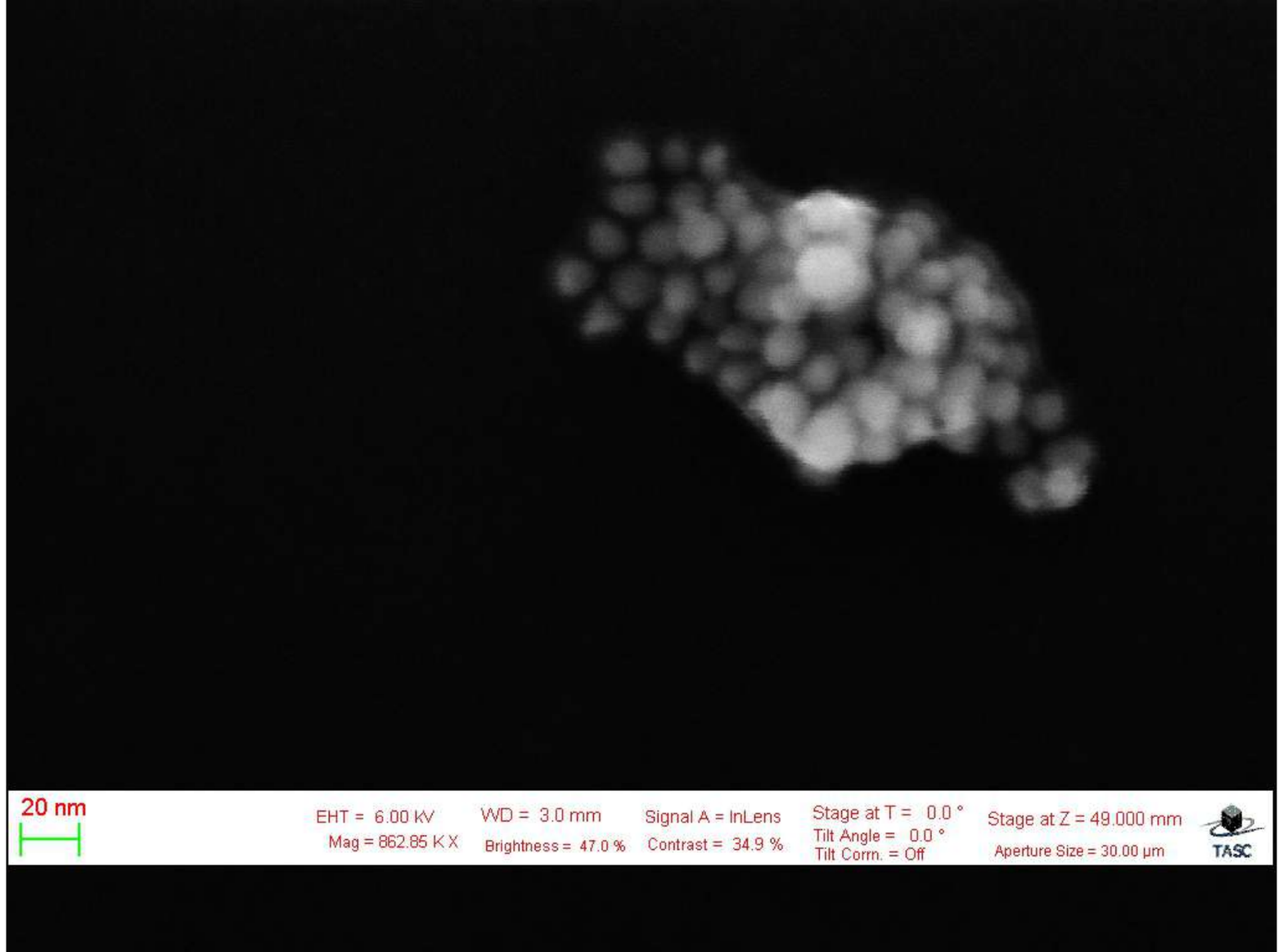}
     \includegraphics[width=0.11\textwidth]{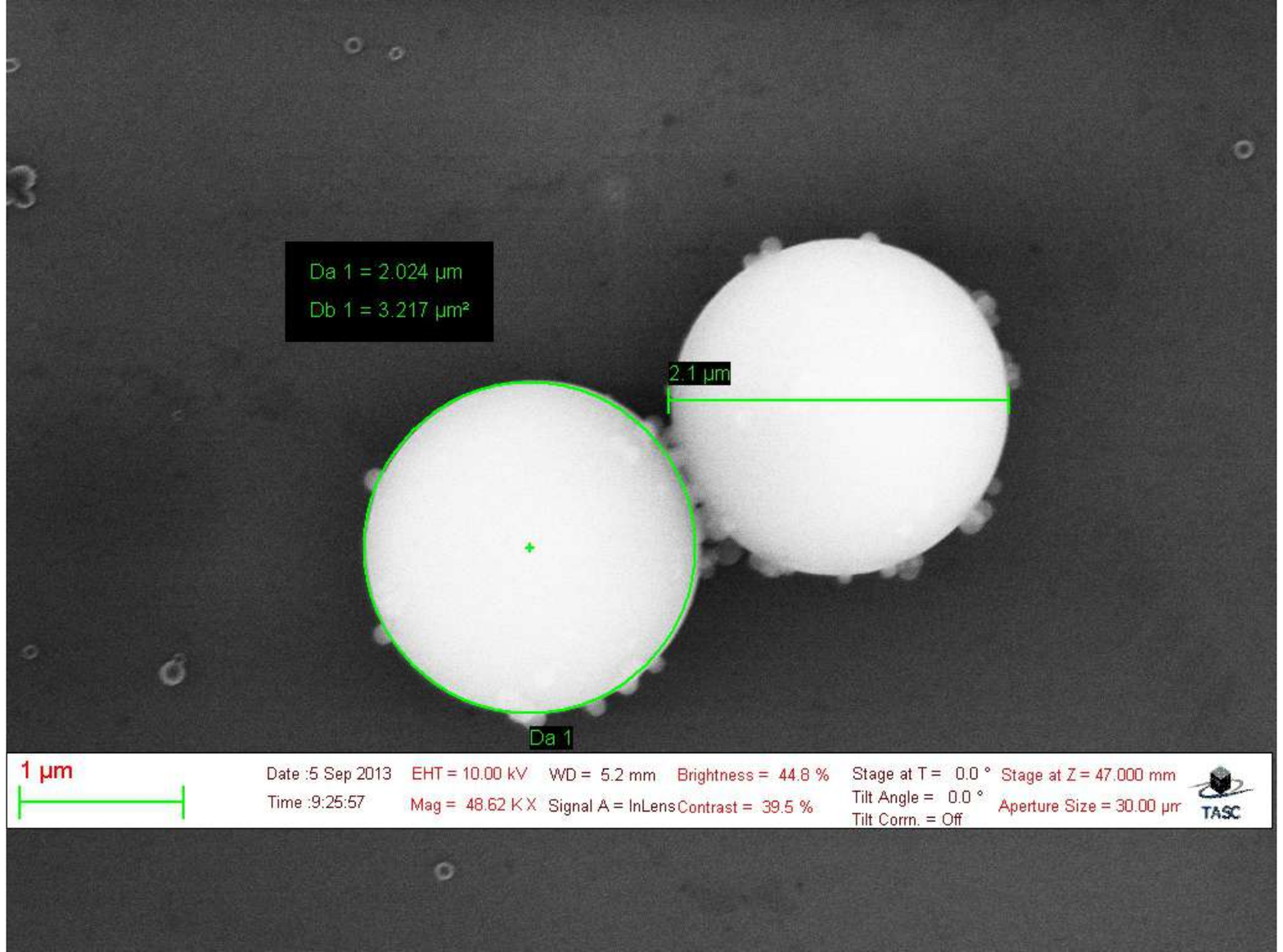}
     \includegraphics[width=0.11\textwidth]{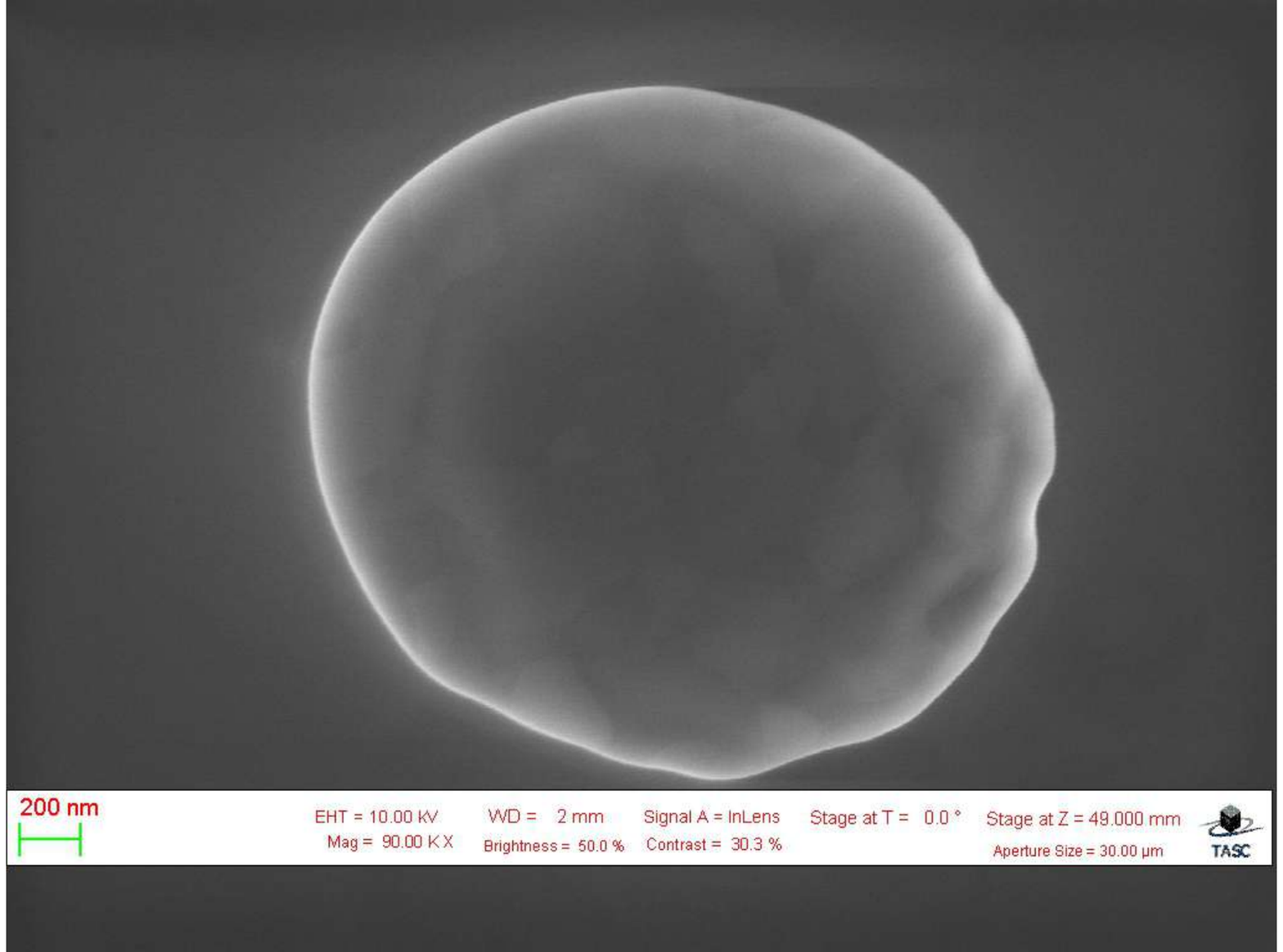}}
     \vspace{-2mm}
     \caption{Challenges in Visual Question Answering (VQA)  task on electron micrographs from the SEM dataset \cite{aversa2018first}.}
     \vspace{0mm}
     \label{fig:figure1}
\end{figure}

\vspace{-3mm}
Creating secure, on-premises small-scale vision-language models for electron micrograph analysis offers several advantages for enterprise adoption, such as improved efficiency, data privacy, cost-effectiveness, interpretability, and customizability. However, this approach also presents significant challenges. Firstly, the scarcity and high cost of high-quality datasets tailored for customizing small-scale multimodal models(SMMs) on electron micrograph analysis for Visual Question Answering (VQA) tasks hinder the collection of necessary data. Secondly, annotating these microscopic images demands expert knowledge and specialized tools, resulting in a resource-intensive and time-consuming process. Finally, the diverse characteristics and representations of microscopic images generated by different imaging techniques pose the most significant obstacle to developing a versatile multimodal model that performs effectively across various electron micrograph-based datasets. In addition, electron micrograph-based zero/few-shot multi-class classification, image-captioning, and, VQA tasks offer powerful insights despite facing challenges. These challenges, highlighted in Figure~\ref{fig:figure1}, include: (1) high intra-class dissimilarity, (2) high inter-class similarity, and (3) the presence of multi-scale visual intricacies (spatial heterogeneity). These factors complicate both accurate image understanding and question answering. 
To address these limitations, we propose a novel framework that utilizes a unique teacher-student paradigm. In this paradigm, a pre-trained foundational large multimodal model (LMM), such as GPT-4, serves as a robust 'teacher' to generate instruction-tuning data (image-question-answer pairs) for customizing a 'student' --- a small-scale, autoregressive, language-and-vision assistant (\texttt{sLAVA}) (hereafter referred to as a small-scale multimodal model or SMM) --- to perform various zero/few-shot multimodal tasks (such as multi-class classification, image captioning, or VQA) for electron microscopy image analysis. Building upon this instruction-following dataset, we introduce vision-language instruction tuning for the smaller multimodal models (SMMs) designed for electron micrograph analysis, thereby eliminating the need for high-quality, human-annotated question-answer pairs for domain-specific customization. Our method efficiently transfers knowledge from a large teacher model to a smaller student model, enhancing its grounded language generation and visual reasoning capabilities to understand the visual content and generate natural language descriptions or responses that accurately reflect the visual information for the end-user question. By distilling the teacher's knowledge, the student achieves performance on par with the original, large-scale proprietary models, demonstrating the efficacy of our approach.
Enterprises can further fine-tune our pre-trained language-and-vision assistant, specifically trained for micrograph analysis tasks, using their proprietary data within their own infrastructure. This empowers them with a secure, on-premises solution for electron micrograph analysis, offering enhanced data privacy, sovereignty, and security, thereby democratizing access to advanced micrograph analysis capabilities. Overall, it accelerates adoption and fosters innovation in semiconductor manufacturing. The proposed small-scale vision-language framework is a visually conditioned autoregressive language generation model with an encoder-decoder architecture, designed for zero-shot or few-shot multiclass classification, image captioning, and VQA tasks. The multimodal model takes as input an interleaved multimodal prompt containing a target microscopic image, supplementary image information, and an end-user question. It then process and aligns the complementary multimodal information to achieve integration of knowledge and semantics, ultimately outputting an open-ended text response grounded in the visual content of the microscopic image. In zero-shot settings, it relies on the domain-specific knowledge acquired during pre-training to answer user questions on unseen images. For few-shot settings, it additionally requires a small set of examples involving microscopic images and the corresponding question-answer pairs (input-output mappings) to tailor its responses for interpreting new, unseen images. \texttt{sLAVA}, a small-scale multimodal model that integrates image processing with language modeling, can answer questions about specific microscopic image characteristics. \texttt{sLAVA} includes the following components: (a) The \textbf{vision encoder} is implemented with a vision transformer\cite{dosovitskiy2020image} to capture the long-range dependencies between microscopic image regions with an expanded receptive field. Consequently, the vision encoder captures salient and global information of the microscopic 

\vspace{0mm}
\begin{figure*}[ht!]
\centering
\resizebox{0.805\linewidth}{!}{ 
\includegraphics[keepaspectratio,height=4.5cm,trim=0.0cm 0.0cm 0cm 0.175cm,clip]{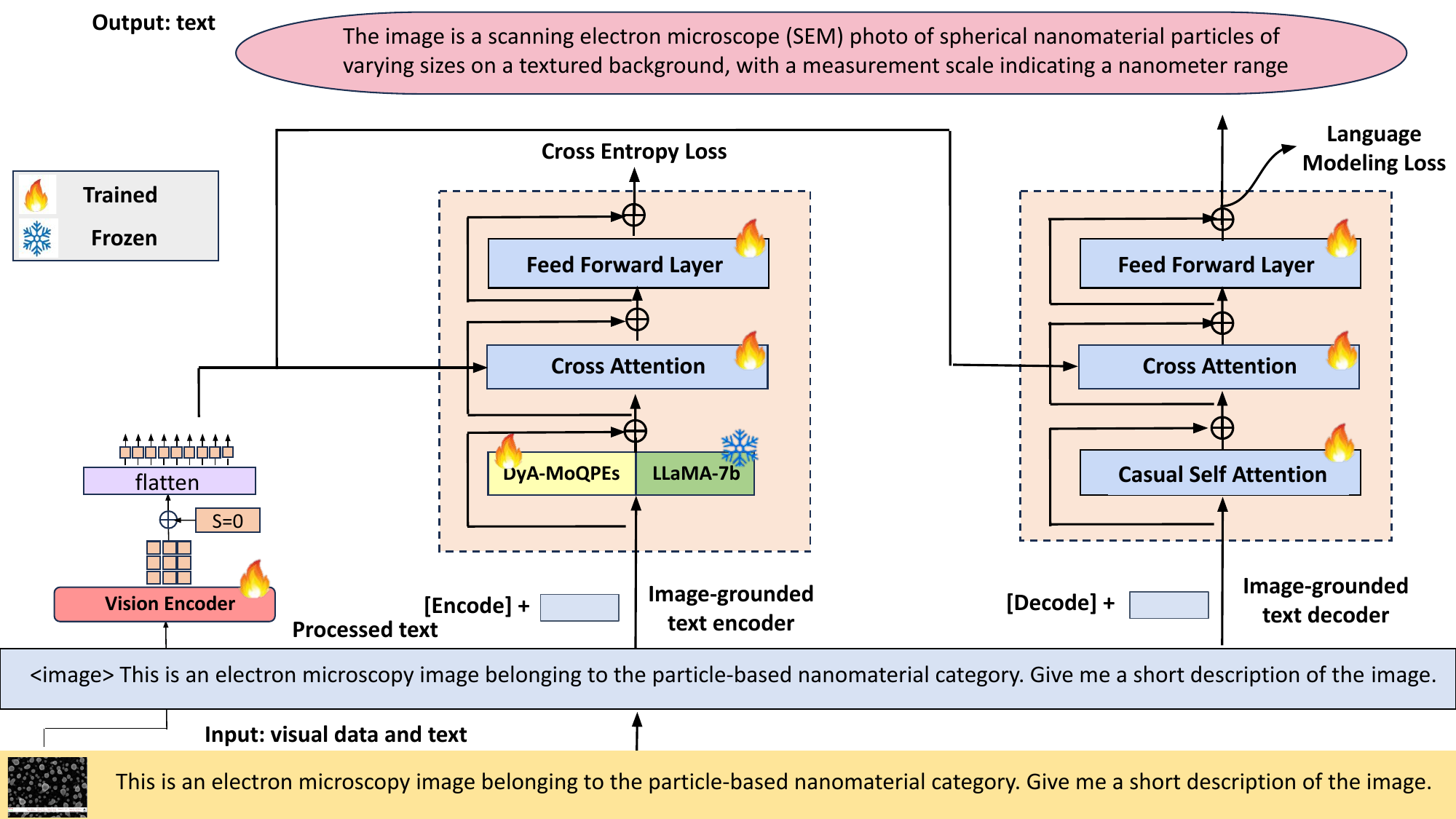} % left, bottom, right, top
}
\vspace{-3mm}
\caption{The schematic illustrates a variant of \texttt{sLAVA}, a small-scale, visually conditioned, autoregressive text generation model that takes prompts combining visual and textual information as input and outputs free-form text for the image captioning task. The input multimodal prompt includes a microscopic image combined with a supplementary user-provided image description, along with the end-user's question. In this zero-shot setting, the task is to answer the question about the microscopic image solely based on the small-scale model's internal parametric knowledge.  \texttt{sLAVA}  comprises a vision encoder to capture the global context of microscopic images, and a text encoder that interprets end-user questions and the auxiliary user-provided image information. The image-grounded text encoder facilitates cross-modal learning by integrating visual information directly into text understanding, thereby generating a comprehensive multimodal representation grounded in the image's visual content. The image-grounded text decoder then synthesizes coherent and contextually relevant textual outputs based on the generated multimodal representations. Finally, the framework is jointly optimized using the binary cross-entropy loss for positive image-text matching and language modeling loss for contextually relevant text generation to answer end-user questions.
}
\label{fig:figure2}
\vspace{-4mm}
\end{figure*}

image, effectively highlighting more relevant visual regions along with their contextual relationships to understand and ground the questions within visual concepts.
We incorporate a $\textless \textit{cls}\textgreater$ token to attend to and aggregate information from all image regions. The higher-level visual semantic representation of the global ($\textless \textit{cls}\textgreater$) token represent the summary of the input image. (b) The text encoder plays a crucial role in analyzing and interpreting user input to understand the nature of the question. It leverages supporting text descriptions associated with the image to extract key details and provide accurate and relevant answers. We insert $\textless \textit{image}\textgreater$ token at the image location in the interleaved multimodal input. We append a $\textless \textit{Encode}\textgreater$ token to the tokenized text to facilitate multimodal integration, with its output embedding representing the fused image-text representation. To better capture the nuances of language and context, the text encoder leverages a pre-trained language model, Llama-2-7b\cite{touvron2023llama}, to compute a high-level representation that captures the semantic meaning and relationships within the end-user question. We fine-tune Llama-2-7b using Dynamic Adaptation of Mixture of Quantized Parameter-Efficient Experts (DyA-MOQPEs) technique (details in the technical appendix) using the instruction-following dataset generated by GPT-4. This Parameter-Efficient Fine-Tuning (PEFT) technique integrates quantization-aware low-rank adaptation (QLoRA) with Mixture of Experts (MoEs) and employs dynamic rank sampling. This approach enhances our ability to interpret natural language questions. Both unimodal encoders play a crucial role in interpreting an end-user question (textual input) regarding the target microscopic image and then analyzing the target microscopic image (visual input) to aid in generating answers that are not only factually accurate but also consistent with the context of the visual information in the microscopic image. (c) The \textbf{image-grounded text encoder} facilitates cross-modal learning to bridge the gap between visual content and linguistic end-user questions by pairing textual descriptions with visual patterns through a cross-attention mechanism. This allows the encoder to focus on relevant image regions and integrate visual information directly into text understanding, resulting in a contextually relevant text representation grounded in the microscopic image's visual content. We minimize the binary cross-entropy loss to align positive image-text pairs.
(d) \textbf{The image-grounded text decoder} leverages multimodal representations to generate accurate and contextually relevant answers, bridging the gap between visual perception and language understanding. To demarcate the generated text sequence, we insert a special $\textless \textit{Decode}\textgreater$ token at the beginning and an end-of-sequence ($\textless \textit{EOS}\textgreater$) token at the end, acting as brackets for the output. The decoder, trained to ground its text generation in visual information, generates contextually relevant descriptions closely aligned with the 

\begin{figure*}[htbp]
\centering
     \subfloat{\hspace{-0mm}\includegraphics[width=0.09\textwidth]{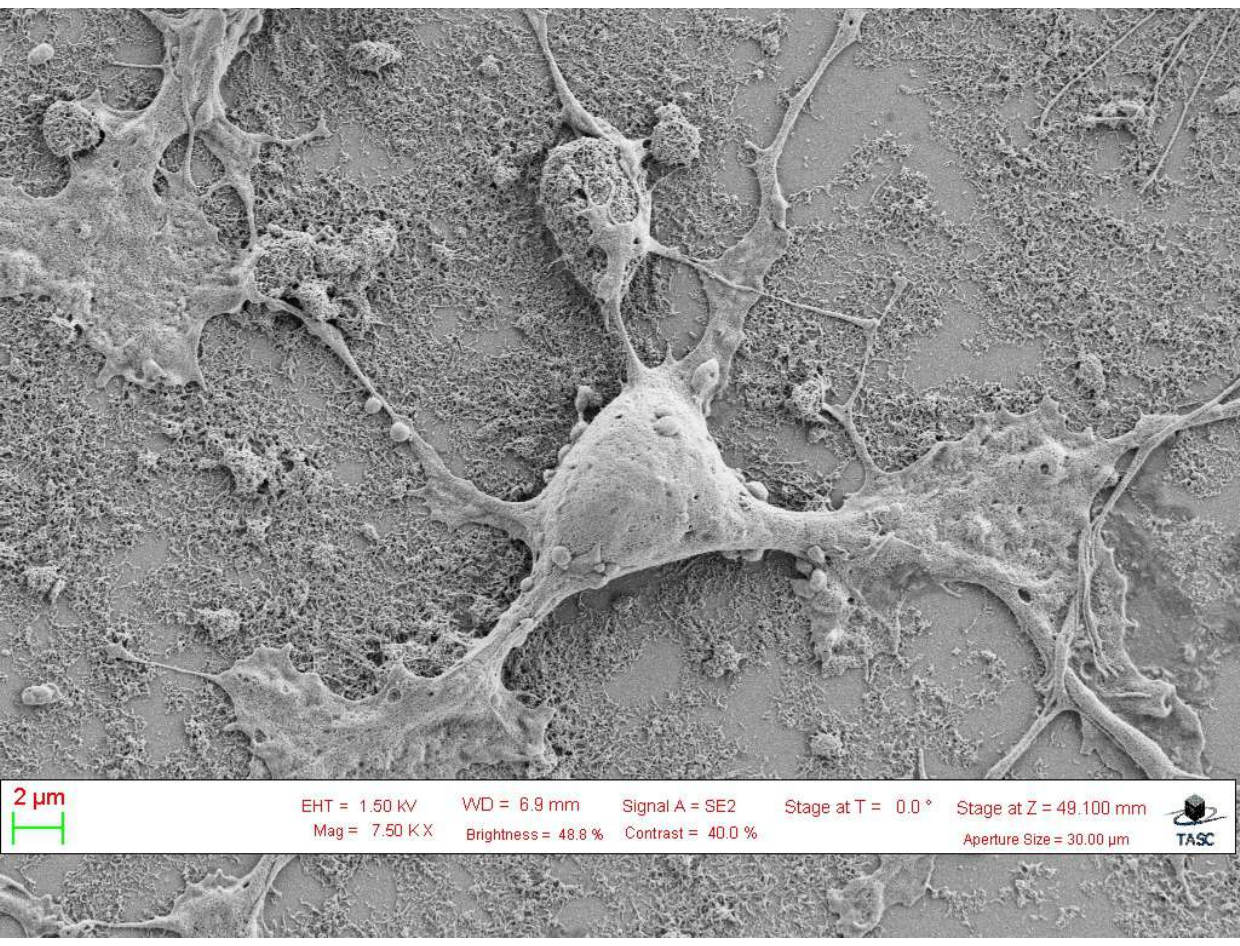}
     \includegraphics[width=0.09\textwidth]{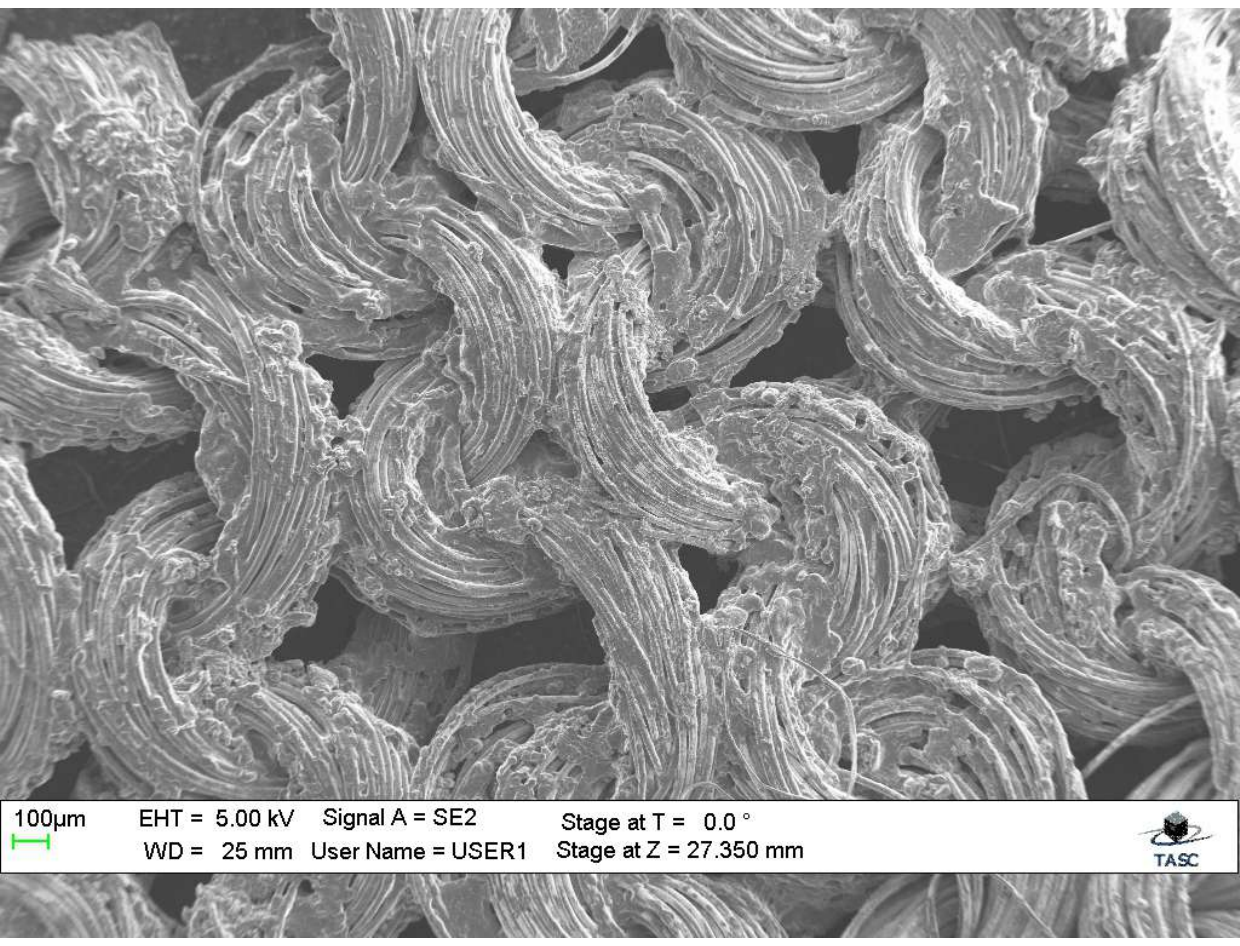}
     \includegraphics[width=0.09\textwidth]{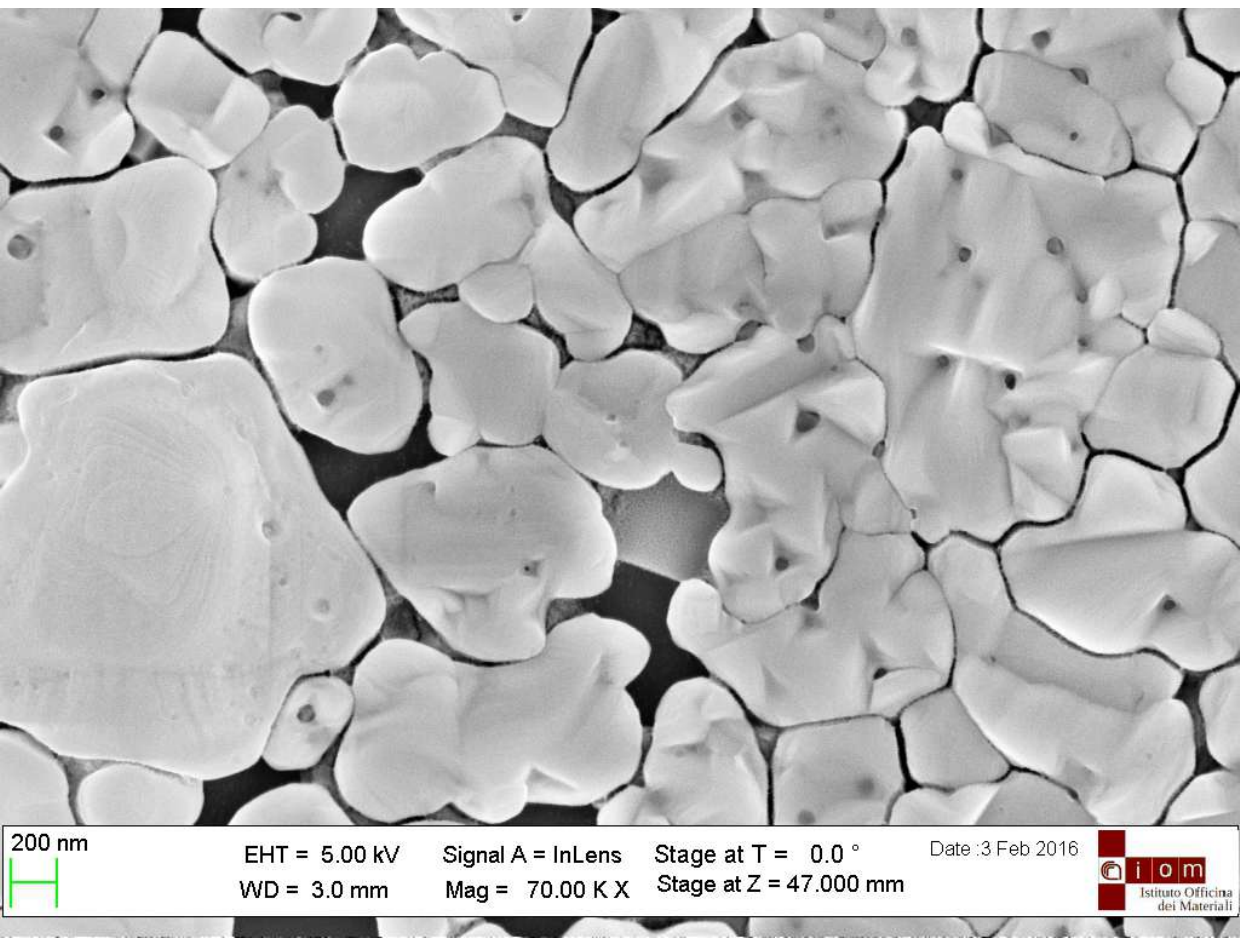}
     \includegraphics[width=0.09\textwidth]{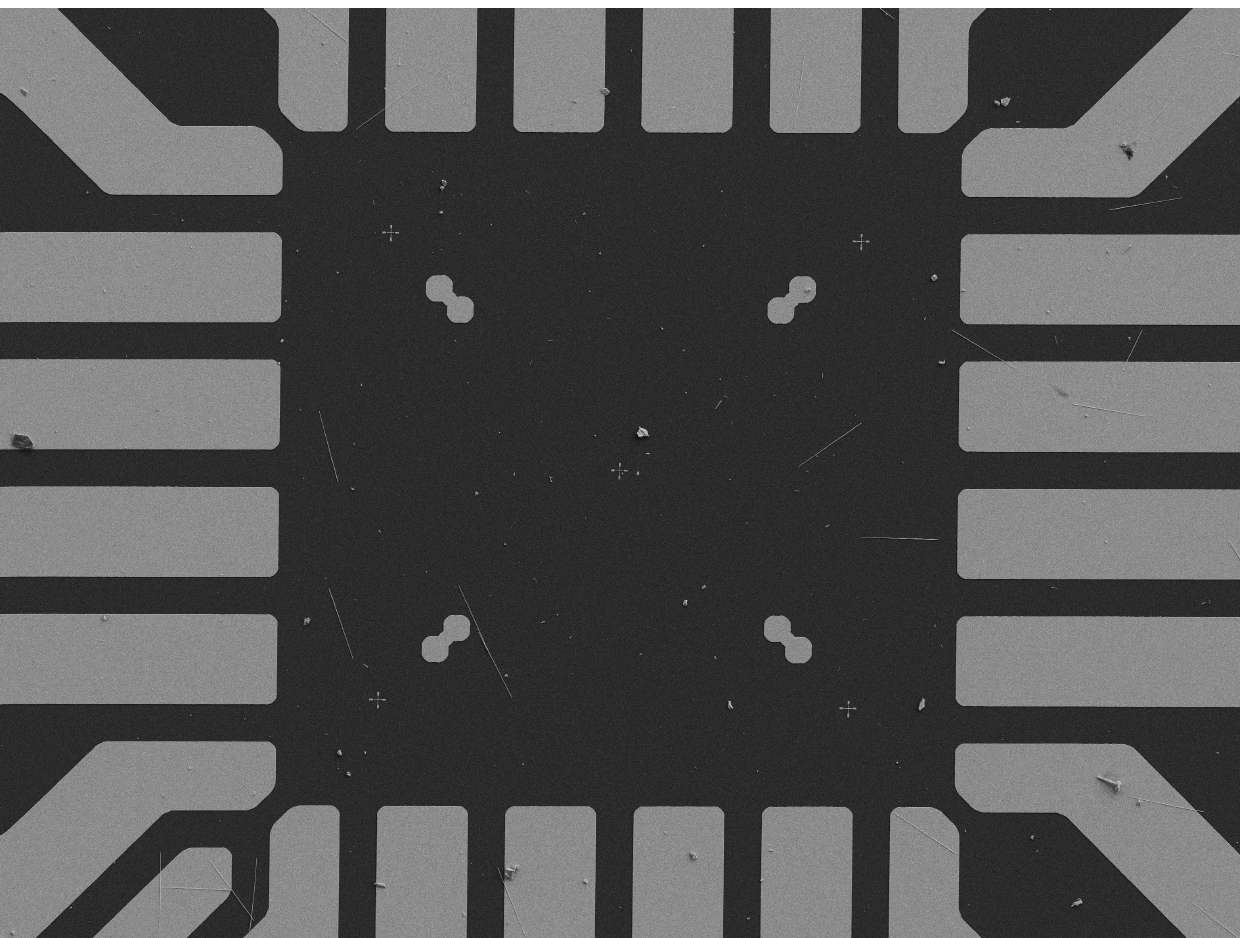}
     \includegraphics[width=0.09\textwidth]{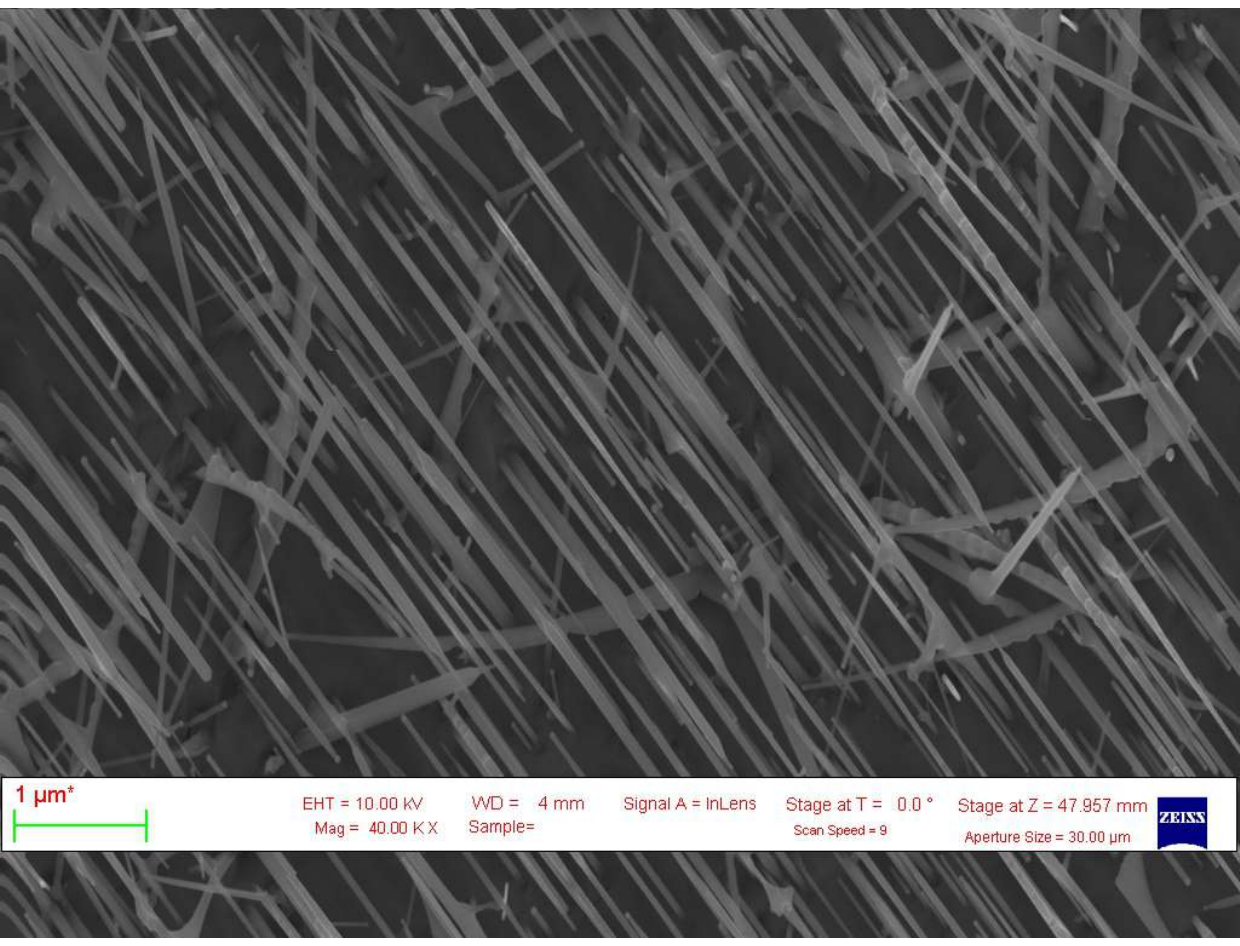}
     }
     \hspace{-7mm}
     \qquad    
     \subfloat{\hspace{-0mm}\includegraphics[width=0.09\textwidth]{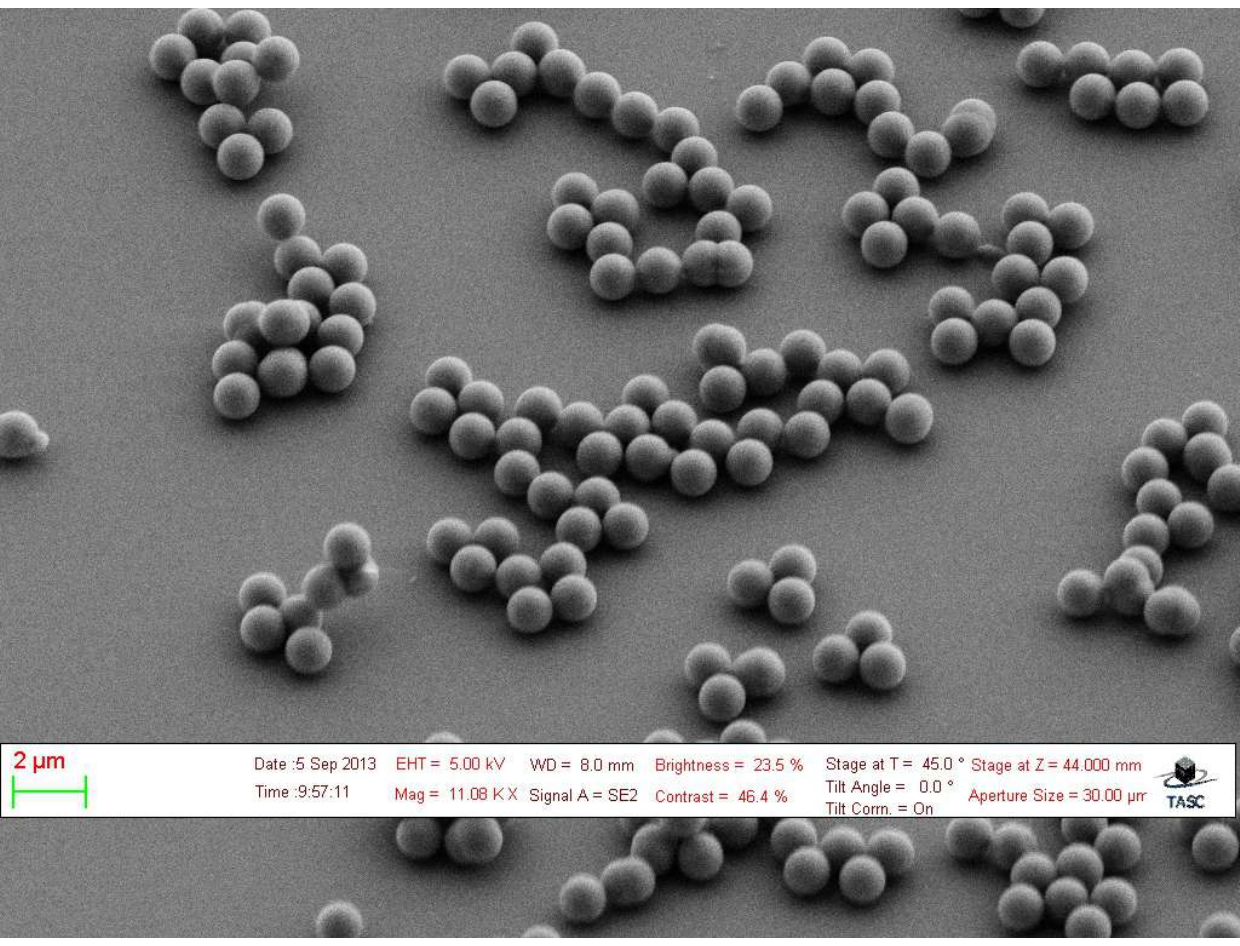}
     \includegraphics[width=0.09\textwidth]{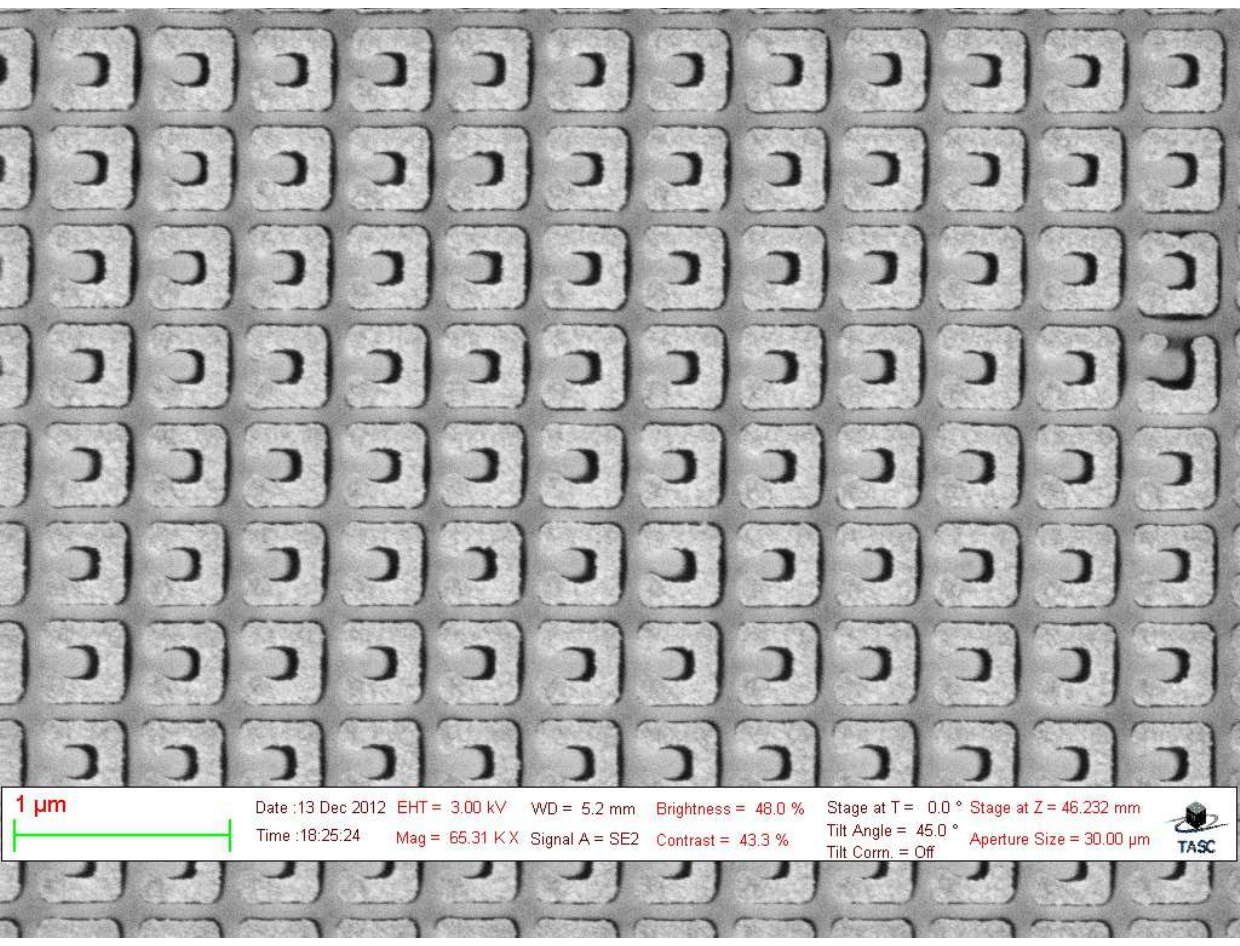}
     \includegraphics[width=0.09\textwidth]{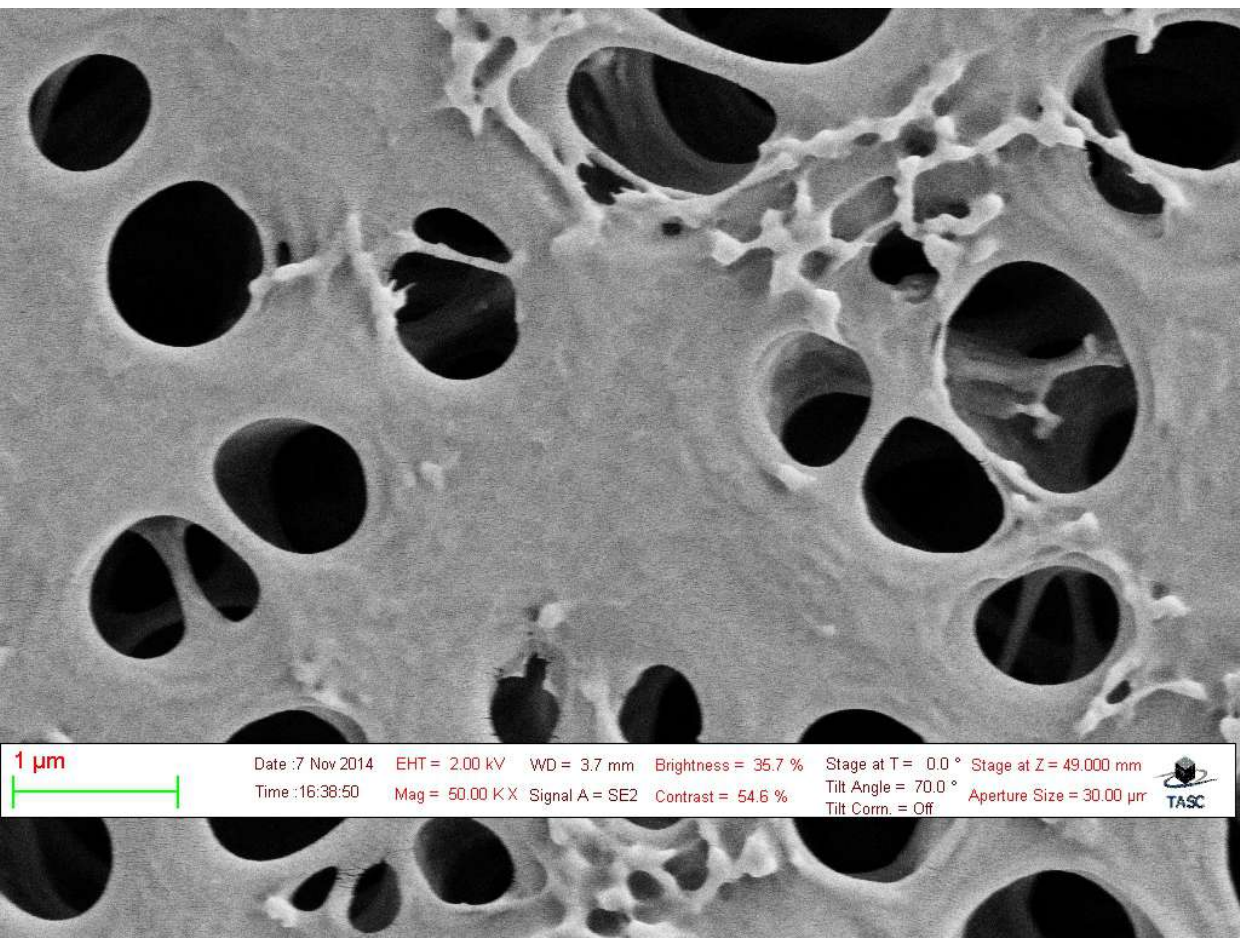}
     \includegraphics[width=0.09\textwidth]{Images/ShortlistedImages/Powder.pdf}
     \includegraphics[width=0.09\textwidth]{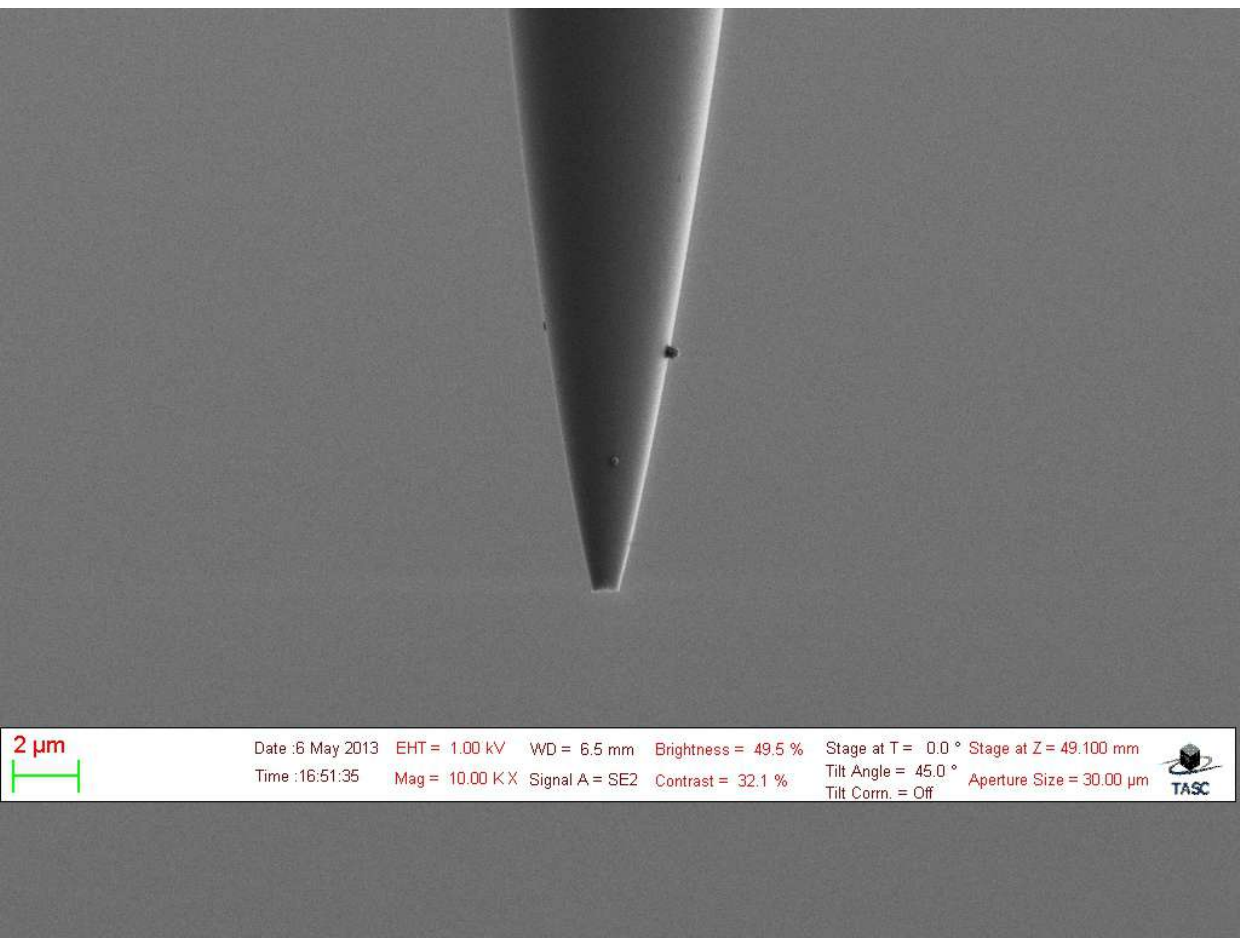}
     }
     \vspace{-8mm}
     \caption{The figure shows SEM images\cite{aversa2018first} showcasing diverse nanomaterial morphologies. Top row: biological structures, fibers, films, MEMS devices, nanowires. Bottom row: nanoparticles, patterned surfaces, porous sponges, powders, tips.}
      \vspace{-5mm}
     \label{fig:illustrationpics}
\end{figure*}

\vspace{-3mm}
\begin{table*}[!ht]
\centering
\caption{The table summarizes the performance of the proposed framework against various methods on the image captioning task.}
\vspace{0mm}
\scalebox{0.75}{
\hspace*{-5mm}\begin{tabular}{l|c|c|c|c|c|c}
\toprule
Method & BLEU-2 & BLEU-4 & ROUGE-1 & ROUGE-2 & ROUGE-L & METEOR \\ 
\midrule
InstructBLIP\cite{dai2305instructblip} & 0.711 $\pm$ 0.032 & 0.660 $\pm$ 0.039 & 0.824 $\pm$ 0.016 & 0.746 $\pm$ 0.005 & 0.814 $\pm$ 0.021 & 0.845 $\pm$ 0.024 \\ 
\midrule
LLaVA\cite{liu2023visual} & 0.715 $\pm$ 0.035 & 0.671 $\pm$ 0.043 & 0.822 $\pm$ 0.016 & 0.757 $\pm$ 0.005 & 0.816 $\pm$ 0.021 & 0.837 $\pm$ 0.023 \\ 
\midrule
MiniGPT-4\cite{zhu2023minigpt} & 0.776 $\pm$ 0.086 & 0.686 $\pm$ 0.100 & 0.839 $\pm$ 0.035 & 0.795 $\pm$ 0.014 & 0.827 $\pm$ 0.047 & 0.864 $\pm$ 0.052 \\ 
\midrule
sLAVA & \textbf{0.819 $\pm$ 0.089} & \textbf{0.727 $\pm$ 0.115} & \textbf{0.939 $\pm$ 0.041} & \textbf{0.876 $\pm$ 0.016} & \textbf{0.880 $\pm$ 0.054} & \textbf{0.906 $\pm$ 0.062} \\ 
\bottomrule
\end{tabular}
}
\label{captioning_results1}
\vspace{-4mm}
\end{table*}

image content. This bridging of visual perception and language generation is achieved through a language modeling loss, ensuring the output accurately captures the image's essence. To achieve robust comprehension and accurate answer generation, our proposed multimodal learning framework employs a two-pronged learning approach. First, we minimize positive image-text pair matching losses to ensure a deep understanding of both visual and textual content. Second, minimization of language modeling loss fosters the generation of accurate and contextually grounded answers. We jointly optimize these objectives through the vision-language instruction tuning of our proposed model, \texttt{sLAVA}, using a multimodal dataset of image-question-answer pairs generated by GPT-4. This enables \texttt{sLAVA} to achieve remarkable expertise in the challenging domain of microscopic image-based question-answering tasks. As illustrated in Figure \ref{fig:figure2}, the proposed framework, \texttt{sLAVA}, is applied for the zero-shot image captioning task. For other tasks, such as zero/few-shot multi-class classification and open-ended VQA, technical details are discussed in the appendix. In summary, the framework outputs free-form text answers to open-ended image-related questions.

\vspace{-3mm} 
\section{Experiments And Results}

\vspace{-2mm}
\subsection{Datasets} 
\vspace{-1mm}
Our study utilized the extensive SEM dataset \cite{aversa2018first} containing over 21,000 electron micrographs across 10 categories of nanomaterials to generate a diverse set of instruction-following multimodal data by GPT-4. We trained our framework for task-specific customization using this machine-generated data only, without relying on any human-annotated data. Unlike previous research \cite{modarres2017neural} that used only a subset of the data, we leveraged the publicly available entire dataset, enabling broader and more robust model training. Since the dataset curator did not provide predefined train/validation/test splits, we randomly divided the dataset into 70$\%$, 10$\%$, and 20$\%$ portions for training, validation, and testing, respectively. Rigorous benchmarking against baseline algorithms demonstrated significant improvements across tasks for the proposed framework, highlighting its effectiveness. Additionally, we tested our framework's generalizability on other open-source material datasets, demonstrating its effectiveness in similar thematic areas. For a detailed discussion on additional benchmark datasets, please refer to the appendix.

\vspace{-3mm}
\subsection{Experimental Studies}
\vspace{-2mm}
We evaluated our framework on various tasks involving microscopic images, including multi-class classification, image captioning, and open-ended VQA, in order to gain a better understanding of the nanomaterials depicted in the electron micrographs. We also explored VQA tasks to evaluate intra-class dissimilarity, inter-class similarity, and spatial heterogeneity, enriching our insights into the nanomaterials depicted in electron micrographs.

\vspace{-3mm}
\subsection{Results} 
\vspace{-2mm}
Table \ref{captioning_results1} presents the experimental results on the image captioning task in terms of evaluation metrics like BLEU, METEOR, and ROUGE, comparing the framework-generated captions with ground-truth captions. Our proposed framework \texttt{sLAVA} surpasses contemporary baseline models, InstructBLIP \cite{dai2305instructblip}, LLaVA \cite{liu2023visual}, and MiniGPT-4 \cite{zhu2023minigpt} on the image captioning task. Table \ref{VQA1} shows representative electron microscope images with their true captions and framework-generated captions, including evaluation metric scores. The experimental results for zero/few-shot classification, open-ended VQA tasks, and others are discussed in the technical appendix.

\vspace{-4mm}
\section{Conclusion}
\vspace{-2mm}
Our research introduces a novel approach to electron micrograph analysis and presents a small-scale, instruction-tuned language-and-vision assistant, customized by a multimodal dataset generated with GPT-4 and optimized for consumer hardware with performance on-par with proprietary LMMs. The pre-trained framework can be further fine-tuned with proprietary data, all without compromising sensitive information to third-party LMMs, making it ideal for secure, efficient, and economically viable enterprise applications.

\bibliography{example_paper}
\bibliographystyle{icml2024}

\begin{table*}[!htb]
    \caption{The table presents randomly sampled electron microscope images alongside their corresponding ground-truth captions and machine-generated captions. It also includes BLEU-2, ROGUE-L, and METEOR metric scores for each caption, which evaluate their similarity to the true captions.}
    \vspace{3mm}
      \centering 
         \begin{tabular}{|>{\centering\arraybackslash}m{2cm}|m{5cm}|m{5cm}|m{2.5cm}|}
        \hline 
        Image & Ground Truth & Answers & BLEU-2/ \ ROGUE-L/ \ METEOR \\ \hline
        \includegraphics[width=2cm,height=1.5cm,keepaspectratio]{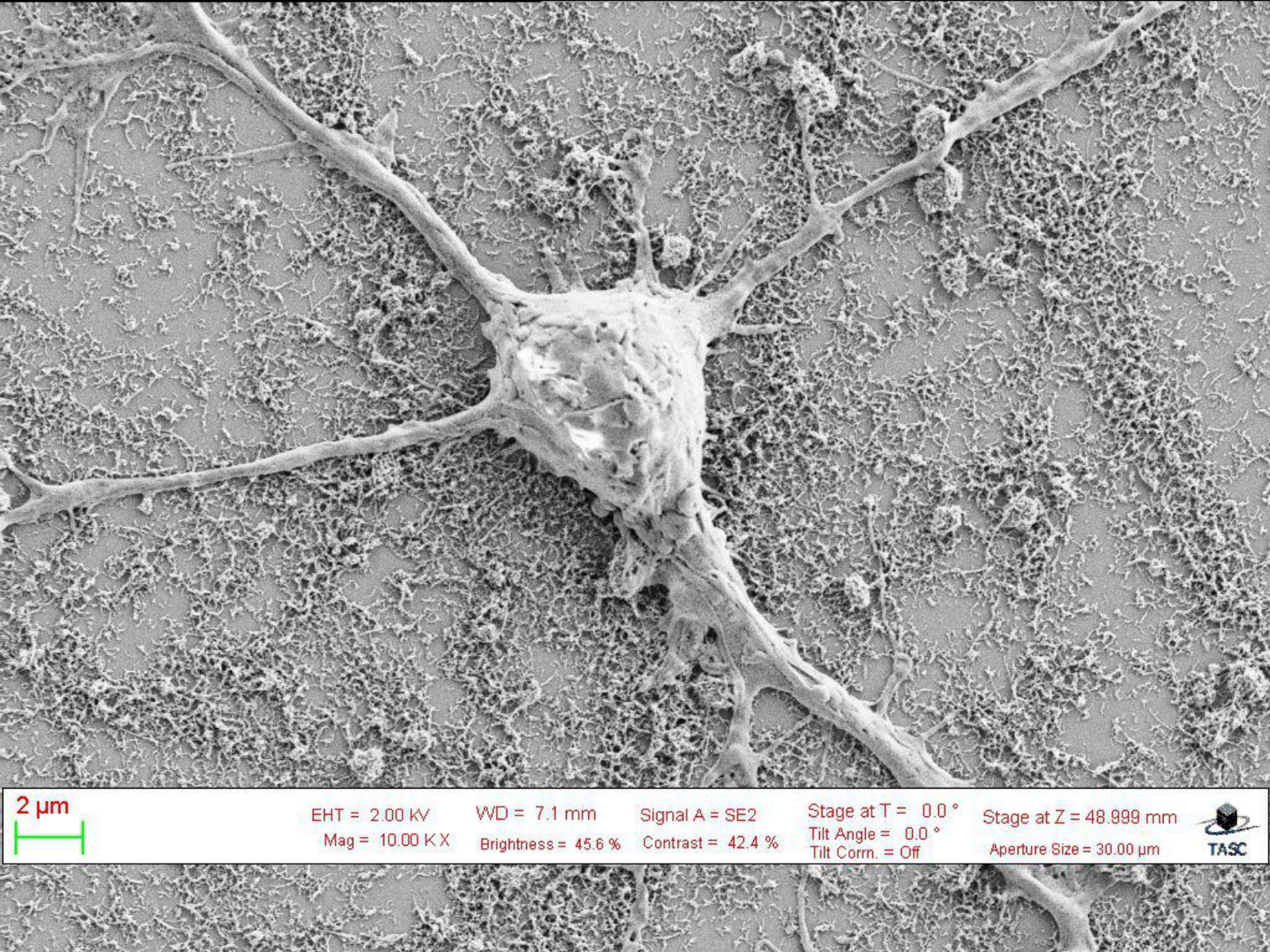} & This electron microscopy image displays a neuron with its dendritic tree and synaptic connections, magnified 10,000 times. & This electron microscopy image shows a neuron with its dendritic tree and synaptic connections magnified 10000 times &\begin{tabular}[c]{@{}c@{}}0.717\\ 0.857\\ 0.879\end{tabular} \\ \hline
        \includegraphics[width=2cm,height=1.5cm,keepaspectratio]{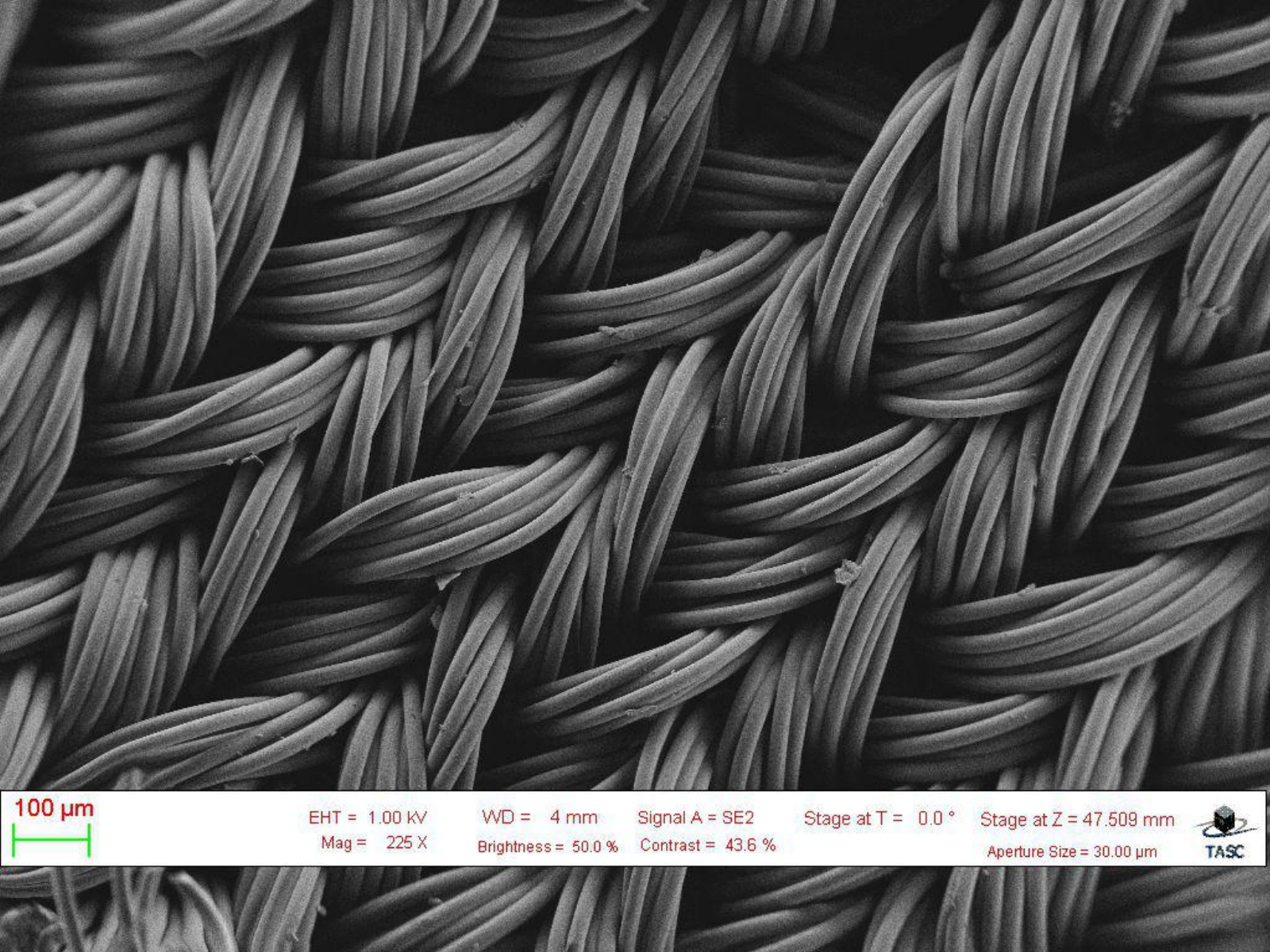} & This SEM image shows tightly woven fibrous material, with each fiber distinctly magnified 225 times to reveal its twisted structure. & This SEM image shows tightly woven fibrous material, with each fiber distinctly magnified 225 times to display its twisted structure &\begin{tabular}[c]{@{}c@{}}0.922\\ 0.950\\ 0.949\end{tabular}  \\ \hline
        \includegraphics[width=2cm,height=1.5cm,keepaspectratio]{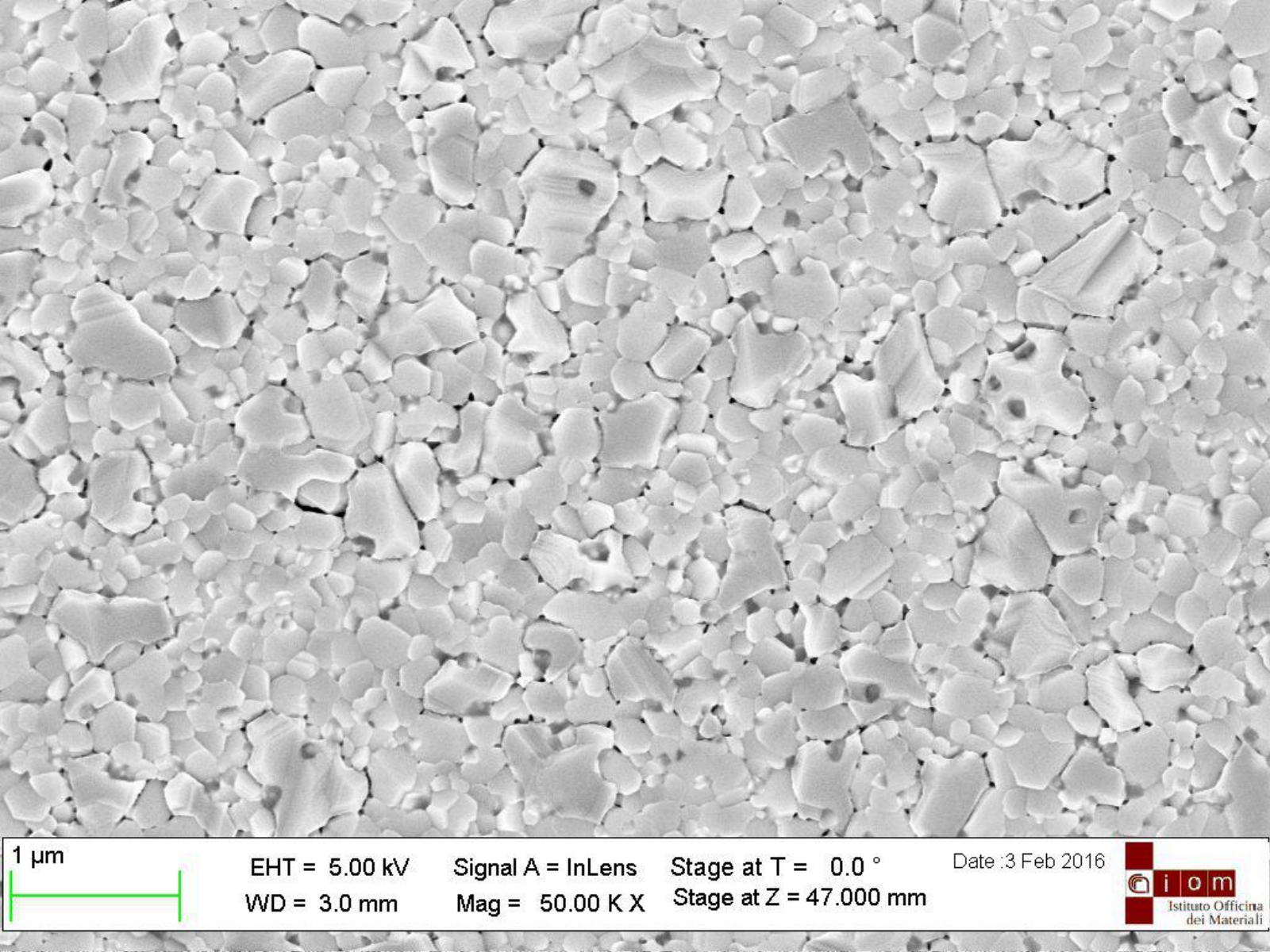} & This SEM image captures a granular film surface with a magnification of 50,000 times, revealing the microstructure of the coated material. & This SEM image shows a granular film surface with a magnification of 50000 times, revealing the microstructure of the coated material. &\begin{tabular}[c]{@{}c@{}}0.851\\ 0.884\\ 0.903\end{tabular}  \\ \hline
        \includegraphics[width=2cm,height=1.5cm,keepaspectratio]{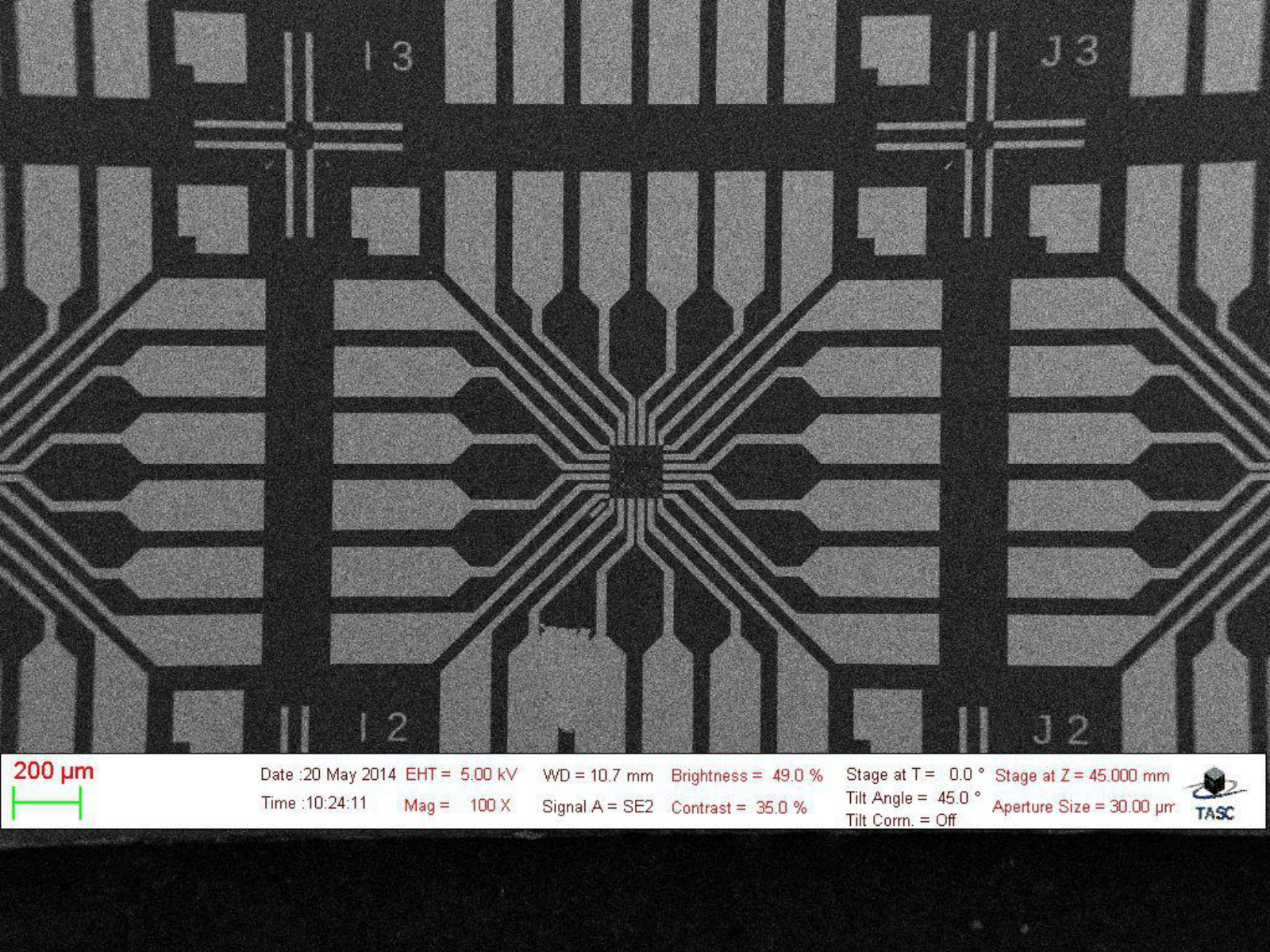} & This SEM image shows a microelectromechanical system (MEMS) with intricate wiring and electrodes, captured at 100 times magnification. & This SEM image displays a microelectromechanical system (MEMS) with intricate wiring and electrodes captured at 100 times magnification. & \begin{tabular}[c]{@{}c@{}}0.824\\ 0.944\\ 0.944\end{tabular}  \\ \hline
        \includegraphics[width=2cm,height=1.5cm,keepaspectratio]{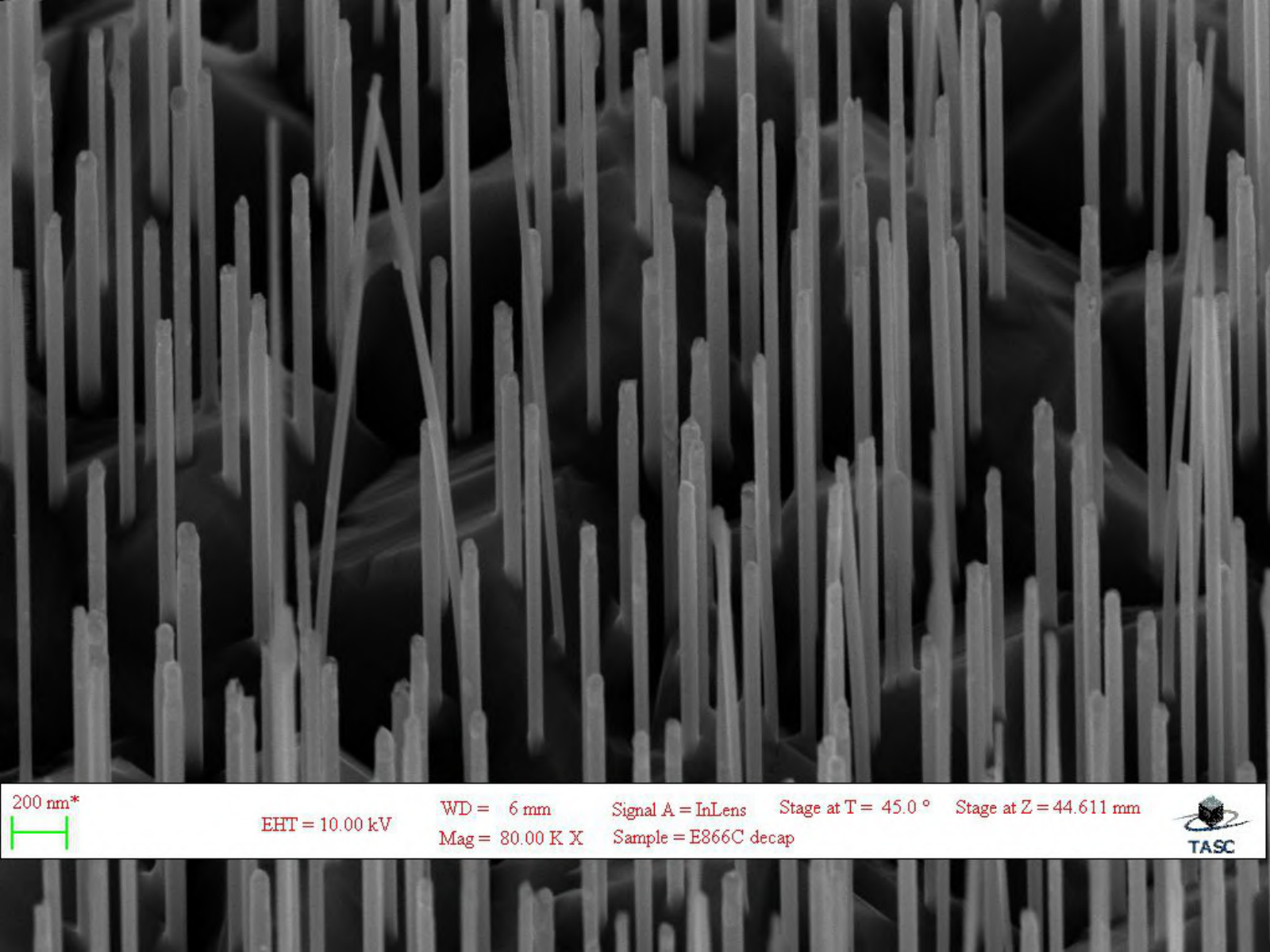} & This SEM image depicts an array of vertical nanowires, showcasing their uniformity and high aspect ratio, magnified at 80,000 times. & This SEM image displays an array of vertical nanowires exhibiting their uniformity and high aspect ratio magnified at 80000 times. & \begin{tabular}[c]{@{}c@{}}0.628\\ 0.829\\ 0.736\end{tabular}  \\ \hline
        \includegraphics[width=2cm,height=1.5cm,keepaspectratio]{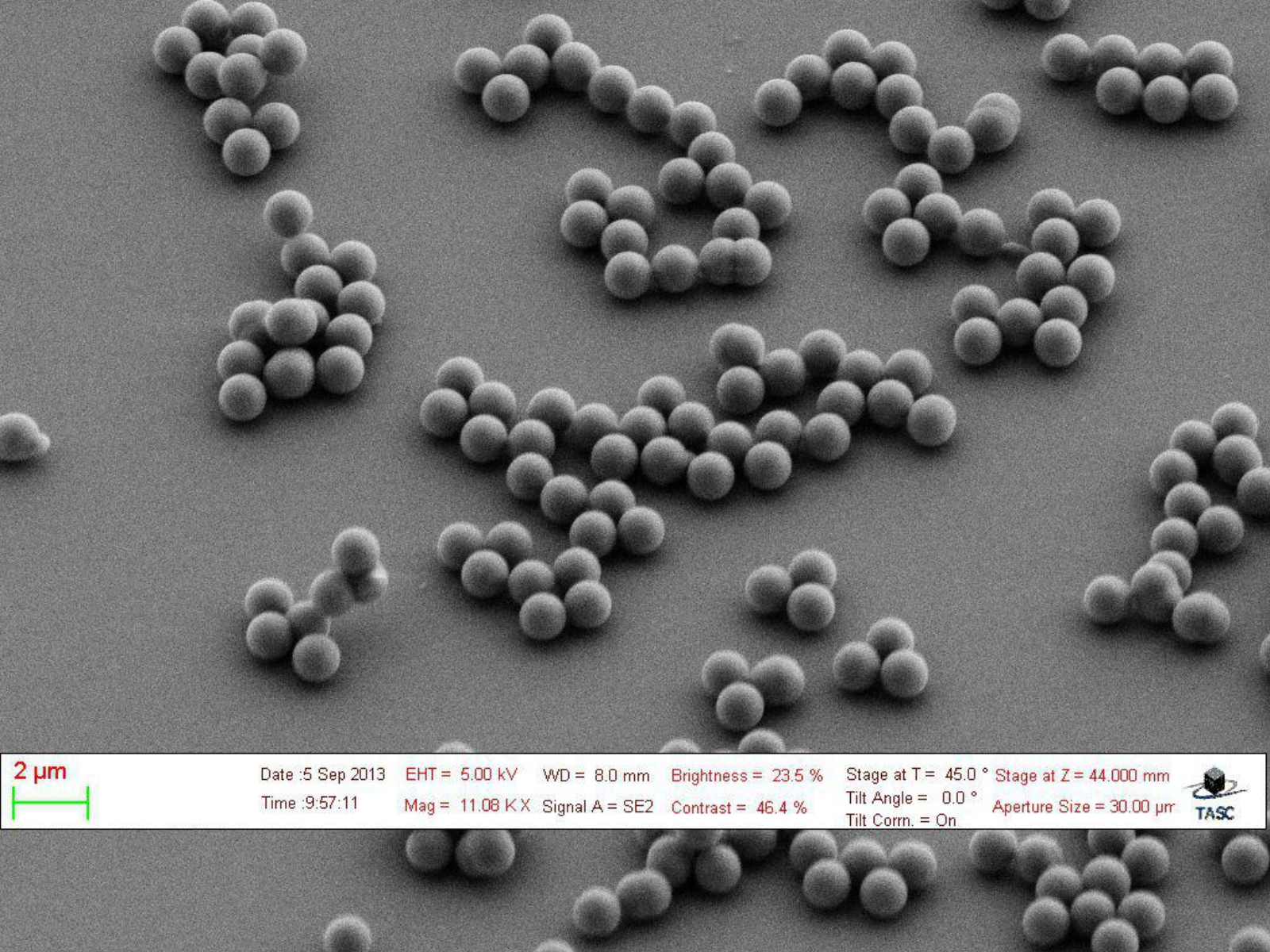} & This SEM image reveals clusters of spherical nanoparticles, each grouping distinct from the others, magnified 11,000 times. & This SEM image shows clusters of spherical nanoparticles, each group distinct from the others, enlarged 11000 times. & \begin{tabular}[c]{@{}c@{}}0.656\\ 0.743\\ 0.819\end{tabular} \\ \hline
        \includegraphics[width=2cm,height=1.5cm,keepaspectratio]{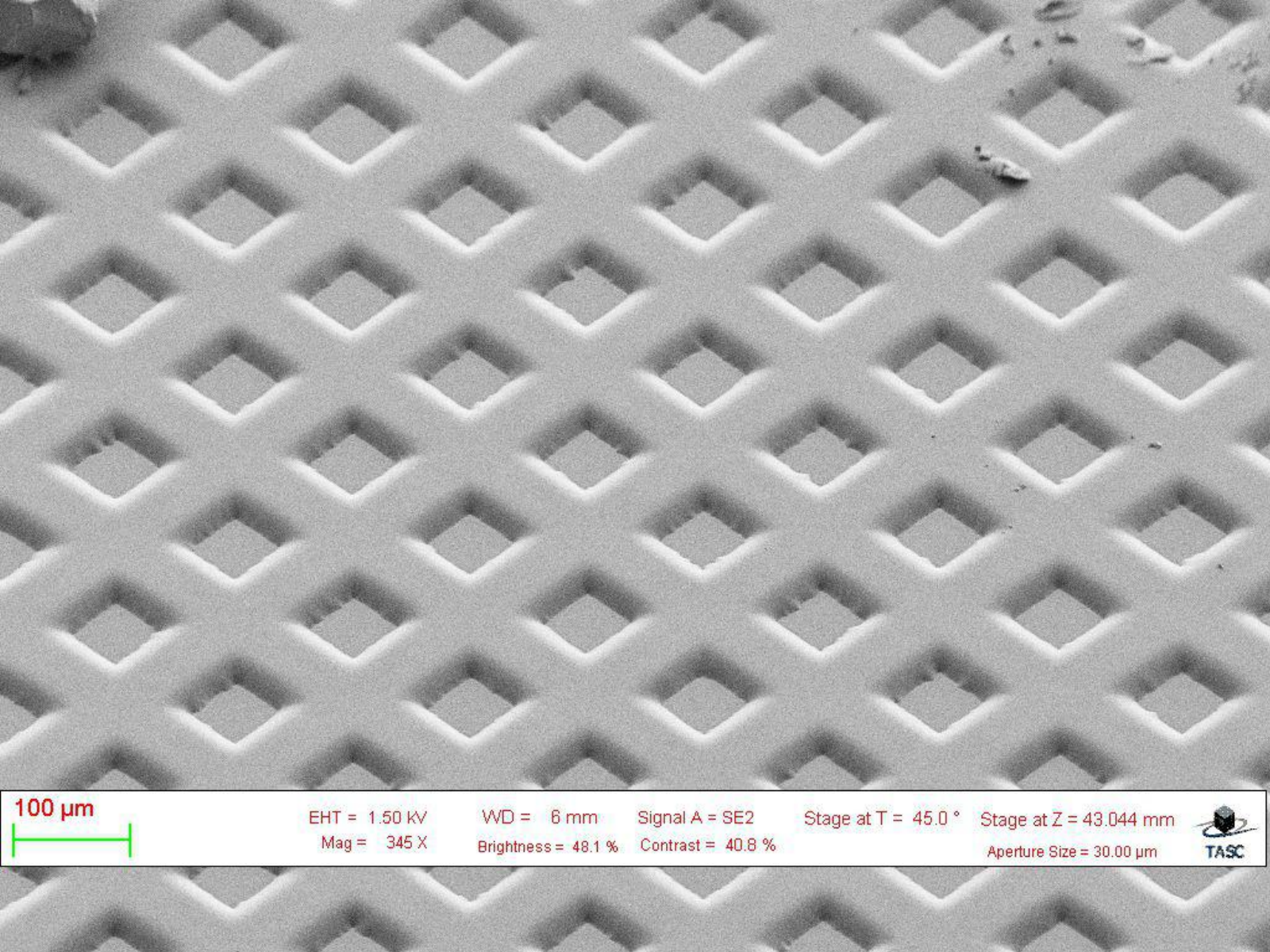} & This SEM image displays a highly ordered, diamond-shaped patterned surface, magnified 345 times, characteristic of nano-fabrication techniques. & This SEM image displays a highly ordered, diamond-shaped patterned surface, magnified 345 times, typical of nano-fabrication techniques & \begin{tabular}[c]{@{}c@{}}0.847\\ 0.947\\ 0.881\end{tabular}  \\ \hline
        \includegraphics[width=2cm,height=1.5cm,keepaspectratio]{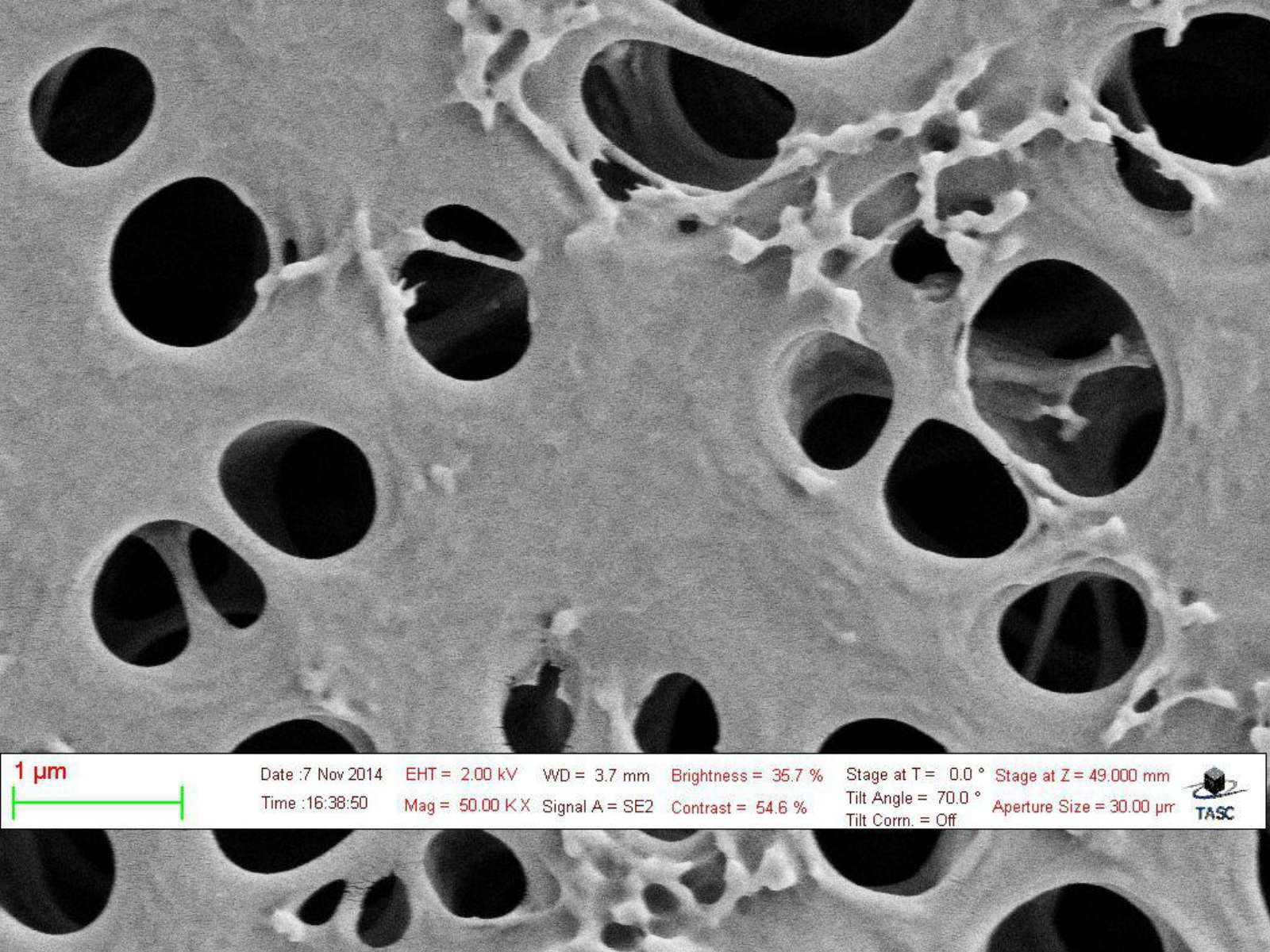} & This SEM image shows a porous sponge-like material with variously sized and shaped voids, magnified 50,000 times to reveal the texture and porosity. & This SEM image displays a porous sponge-like substance with varied sized and shaped voids, enlarged 50000 times to show the texture and porosity. & \begin{tabular}[c]{@{}c@{}}0.608\\ 0.735\\ 0.760\end{tabular}  \\ \hline
        \includegraphics[width=2cm,height=1.5cm,keepaspectratio]{Images/SEM-ImageCaptioning/Powder.pdf} & This SEM image reveals a dense aggregation of nanoscale particles forming a powder, with a magnification of 13,570 times. & This SEM image shows a dense cluster of nanoscale particles composing a powder, with a magnification of 13,570 times. & \begin{tabular}[c]{@{}c@{}}0.749\\ 0.735\\ 0.836\end{tabular}  \\ \hline
        \includegraphics[width=2cm,height=1.5cm,keepaspectratio]{Images/SEM-ImageCaptioning/Tips.pdf} & This SEM image shows a sharply pointed nanomaterial tip, highlighted against a stark background at a magnification of 100,000 times. & This SEM image displays a sharply pointed nanomaterial tip, highlighted against a stark background with a magnification of 100,000 times. & \begin{tabular}[c]{@{}c@{}}0.843\\ 0.735\\ 0.949\end{tabular} \\ \hline
        \end{tabular}    
        \label{VQA1}
\end{table*}

%%%%%%%%%%%%%%%%%%%%%%%%%%%%%%%%%%%%%%%%%%%%%%%%%%%%%%%%%%%%%%%%%%%%%%%%%%%%%%%
%%%%%%%%%%%%%%%%%%%%%%%%%%%%%%%%%%%%%%%%%%%%%%%%%%%%%%%%%%%%%%%%%%%%%%%%%%%%%%%
% APPENDIX
%%%%%%%%%%%%%%%%%%%%%%%%%%%%%%%%%%%%%%%%%%%%%%%%%%%%%%%%%%%%%%%%%%%%%%%%%%%%%%%
%%%%%%%%%%%%%%%%%%%%%%%%%%%%%%%%%%%%%%%%%%%%%%%%%%%%%%%%%%%%%%%%%%%%%%%%%%%%%%%

\twocolumn
\section{Technical Appendix}

\vspace{0mm} 
\subsection{Dynamic Adaptation of Mixture of Quantized Parameter-Efficient Experts (DyA-MoQPEs)}
\vspace{0mm}
Low-Rank Adaptation (LoRA, \cite{hu2021lora})  is a parameter-efficient technique that enables the efficient fine-tuning of large foundational models on consumer hardware (low-cost GPUs). LoRA injects and adapts these additional parameters while keeping the original pre-trained weights frozen, allowing for task-specific customization without full-parameter fine-tuning. LoRA dramatically reduces memory and computational requirements for fine-tuning foundational models with task-specific corpus without increasing inference latency. LoRA serves as a plug-and-play solution for tailoring general-purpose large-scale foundational models to specialized tasks, retaining parametric knowledge acquired from the vast training corpus and mitigating catastrophic forgetting of pre-training knowledge while effectively learning new information. LoRA incorporates a lightweight, trainable pair of low-rank matrices (adapter modules) into each pre-trained model layer. LoRA updates these ancillary parameters while keeping the original pre-trained weights fixed, achieving performance comparable to that of traditional full-parameter fine-tuning but with enhanced resource efficiency. Large-scale pretrained  models \cite{vaswani2017attention} benefit from LoRA's ability to incorporate low-rank adapter modules into their linear layers, enhancing performance on specialized tasks. These ubiquitous layers hold a significant portion of the parameters and directly influence learning, making them ideal targets for efficient fine-tuning. In LoRA, updates to the linear layer are achieved by introducing new trainable parameters, denoted as $\Delta \mathbf{W}$, that capture task-specific information without altering the original pre-trained weight matrix represented as $\mathbf{W}_0$. $\Delta \mathbf{W}$ is linearly added to $\mathbf{W}_0$ to achieve task-specific adaptation while keeping the original weights frozen. The low-rank adaptation of the linear layer, with input $\mathbf{X}$ and output $\mathbf{Y}$, can be mathematically described as follows:

\vspace{0mm}
\resizebox{0.95\linewidth}{!}{
\begin{minipage}{\linewidth}
\begin{equation}
   \mathbf{Y} = (\mathbf{W}_0 + \Delta \mathbf{W}) \mathbf{X} = \mathbf{W}_0\mathbf{X} + (\alpha \mathbf{A}\mathbf{B}) \mathbf{X}  
\end{equation}
\label{eqn:1}
\end{minipage}
}

\vspace{-4mm}
Here, $\mathbf{Y} \in \mathbb{R}^{b \times d_{out}}$ and $\mathbf{X} \in \mathbb{R}^{b \times d_{in}}$. The dimensions of the input and output are denoted by $d_{in}$ and $d_{out}$, respectively, with $b$ representing the batch size. The original weight matrix $\mathbf{W}_{0} \in \mathbb{R}^{d_{in} \times d_{out}}$ holds the pretraining knowledge, preserving the foundational model's general capabilities, while the low-rank addition $\Delta \mathbf{W}$ to $\mathbf{W}_0$ captures task-specific information during fine-tuning. $\mathbf{A} \in \mathbb{R}^{d_{in} \times r}$ is a projection-down weight matrix, and $\mathbf{B} \in \mathbb{R}^{r \times d_{out}}$ is a projection-up weight matrix. The rank of the decomposition, denoted as $r$, is a hyperparameter notably smaller than both $d_{in}$ and $d_{out}$, expressed as $r \ll d_{in}, d_{out}$. The value of $r$ is a critical parameter that optimizes the trade-off between model adaptability, efficiency, and generalization. The scaling factor $\alpha$ is typically set to $\frac{1}{r}$. During the fine-tuning, the trainable weight matrices $\mathbf{A}$ and $\mathbf{B}$ are updated, while $\mathbf{W}_0$ remains constant. Fine-tuning foundational models involves computing parameter gradients through a task-specific loss function, updating trainable parameters using Adam \cite{kingma2014adam} or SGD \cite{robbins1951stochastic} optimizers, and storing additional meta-data such as momentum and adaptive learning rates. Fine-tuning foundational models demands significant memory for model parameters, gradients, and optimizer states. LoRA reduces this memory overhead by decreasing the number of trainable parameters through low-rank adaptation. Consequently, LoRA requires fewer computational resources compared to full-parameter fine-tuning, offering a more efficient method for adapting foundational models to specialized tasks. While LoRA reduces memory usage due to fewer trainable parameters, it still requires significant memory to hold large intermediate input activations (\(\mathbf{X} \in \mathbb{R}^{b \times d_{in}}\); refer to Equation \ref{eqn:1}) during the computation of gradients for low-rank weights, $\mathbf{A}$ and $\mathbf{B}$, during backpropagation. This high activation memory demand limits scalability, especially under resource constraints. Methods like selective LoRA \cite{hu2021lora} or activation recomputation \cite{chen2016training} could help mitigate this issue, but they may impact efficiency. Thus, the high demand for activation memory remains a challenge in efficiently adapting large-scale foundational models with LoRA, posing a significant limitation. To overcome the aforementioned limitations, LoRA with Frozen-A (LoRA-FA) \cite{zhang2023lora}—a variant of LoRA—reduces activation memory footprint by avoiding the storage of full-rank input activations, enabling efficient fine-tuning of foundational models on limited resources without compromising performance. LoRA-FA accomplishes this through freezing both the original pre-trained weights, $\mathbf{W}_{0}$, and the projection-down weight, $\mathbf{A}$, while only updating the projection-up weight, $\mathbf{B}$, which is typically initialized to zero. The frozen projection-down weight matrix $\mathbf{A}$, sampled from a normal distribution, maps the high-dimensional input $\mathbf{X}$ into a reduced $r$-dimensional space ($\mathbf{A}\mathbf{X} \in \mathbb{R}^{b \times r}$, where $r \ll d_{in}$). This low-dimensional mapping further reduces the activation memory requirements for gradient computation of $\mathbf{B}$ during back-propagation. In essence, LoRA-FA effectively decreases the number of trainable parameters and also reduces the activation memory usage, making it an efficient technique for fine-tuning large-scale foundational models without increasing inference latency. We propose a novel approach that combines the advantages of the Mixture of Experts (MoEs) framework with Parameter-Efficient Fine-Tuning (PEFT) techniques, such as LoRA-FA. We refer to this innovative method as Mixture of Parameter-Efficient Experts (MoPEs) \cite{zadouri2023pushing}. This method adapts the MoE approach to be more parameter-efficient by integrating LoRA-FA adapters. We employ MoPEs technique to instruction-tune pretrained foundational models, thereby improving their performance on niche, domain-specific tasks while minimizing resource usage. In the MoPE architecture, a set of specialized experts, known as LoRA-FA adapters, are trained to address different aspects of the fine-tuning data. This targeted approach allows each expert to focus on specific data aspects, significantly enhancing the performance of pretrained decoder-only foundational models on complex downstream tasks. These multiple experts are activated based on a gating mechanism denoted as router $R$, designed for conditional computation. We represent the set of $K$ experts as ${\textbf{B}_0 = E(\textbf{X}; \theta_0), \ldots, \textbf{B}_K = E(\textbf{X}; \theta_K)}$, where each $\textbf{B}_k$ corresponds to the weight matrix of the $k$-th expert, which is learned during the fine-tuning based on the downstream task. Here, $E$ represents a parameterized function, and $\theta_k$ denotes the trainable parameters specific to expert $k$. The router $R$ typically takes the form of another feed-forward network, producing a $k$-dimensional vector that indicates the routing probabilities for each expert.

\vspace{-4mm}
\resizebox{0.90\linewidth}{!}{
\hspace{0mm}\begin{minipage}{\linewidth}
\begin{equation}
\mathbf{Y} = (\mathbf{W}_0 + \Delta \mathbf{W}) \mathbf{X} = \mathbf{W}_{0} \mathbf{X} + \mathbf{A} (\overline{\mathbf{B}}\mathbf{X}), \hspace{1mm}\overline{\mathbf{B}}=\sum_{k=1}^{K} R(\mathbf{X})_{k} \mathbf{B}_{k} \nonumber
\end{equation}
\end{minipage}
}

\vspace{-2mm}
Here, $\overline{\mathbf{B}}$ represents a composite weight matrix obtained by combining the contributions of multiple expert weight matrices, with each matrix weighted by its respective routing probability. We implement a top-$k$ routing strategy for soft merging, where only the $k$ experts with the highest routing probabilities contribute to the composite matrix. This effectively reduces computational complexity. While MoPEs slightly increases trainable parameters compared to LoRA-FA due to conditional computation, the reduced activation memory usage makes it an economical choice for fine-tuning on consumer-grade hardware with improved performance than LoRA-FA. While MoPEs effectively reduce memory usage without compromising their fine-tuning performance, they are not without limitations. Carefully tuning rank $r$ is crucial, as it balances model complexity and learning complex data patterns. A static rank, however, could limit adaptability to data distribution shifts. To address these limitations, which stem from a fixed rank size and require exhaustive searches for the optimal rank, we introduce `Dynamic low-rank adaptation with MoPEs' (denoted as DyA-MoPEs). Specifically, DyA-MoPEs can adapt across various ranks within the range from $r_{\text{min}}$ to $r_{\text{max}}$, where $r_{\text{min}}$ and $r_{\text{max}}$ are introduced as hyperparameters during training. This approach eliminates the need for multiple training iterations to determine the optimal singular rank. Dynamic low-rank adaptation offers significant advantages by allowing dynamic rank adjustments during training for effective performance across a broad range of ranks. Moreover, DyA-MoPEs can adapt their rank based on the task, making them suitable for continuous learning scenarios or contexts with frequent data distribution shifts. During fine tuning, we dynamically sample a rank \(b\) from a pre-defined categorical distribution, $b\sim p_B(\text{Range}[r_{\text{min}}, r_{\text{max}}])\) and the pair of low-rank matrices are truncated as follows:

\vspace{-2mm}
\resizebox{0.925\linewidth}{!}{
\begin{minipage}{\linewidth}
\begin{align*}
\overline{\mathbf{B}}^{\downarrow b} &= \overline{\mathbf{B}}[1 : b, :]\\
\mathbf{A}^{\downarrow b} &= \mathbf{A}[:, 1 : b]  \\
\mathbf{Y} &= W_0\mathbf{X} + \alpha \mathbf{A}^{\downarrow b}(\overline{\mathbf{B}}^{\downarrow b}\mathbf{X}) 
\end{align*}
\end{minipage}
}

This truncation keeps the first b rows of $\overline{\mathbf{B}}$ and the first b columns of $\mathbf{A}$, resulting in matrices with a lower rank. Consequently, the output $\mathbf{Y}$ is computed using these lower-rank matrices, allowing for dynamic adjustment of model complexity during training. We compute gradients for these truncated matrices and apply updates accordingly. To manage the increased computational complexity, we utilize custom gradient accumulation. This technique enables more stable and efficient learning by accumulating gradients over multiple iterations or steps. Additionally, we implement rank normalization to equalize the influence of different ranks on the model's learning process. By scaling gradients or updates according to the rank size, this method helps stabilize training and ensures fair contributions from all ranks. To reduce their memory footprint, we quantize the pre-trained weights or base weights ($\mathbf{W}_0$) of the Llama 2-7B model from a 16-bit format into a lower precision format (e.g., 8-bit quantization \cite{dettmers2023qlora, xu2023qa}). During inference, the product of low-rank adapter parameters, $\mathbf{A}^{\downarrow b}$ and $\overline{\mathbf{B}}^{\downarrow b}$, is combined with these quantized weights to approximate the original full-precision model.

\vspace{-2mm}
\subsection{Fine-Tuning, Pretrained Large Language Models(LLMs)}
\vspace{-1mm}
The Llama-2 \cite{touvron2023llama}, a sophisticated auto-regressive, language-optimized transformer architecture tailored specifically for various natural language processing tasks, leverages supervised fine-tuning (SFT) and reinforcement learning with human feedback (RLHF) optimized for chat applications and natural language generation tasks. The core strength of Llama-2 lies in its ability to process and generate text for end-user questions that closely resembles human language, making it highly suitable for complex language processing tasks. The Llama-2's architecture, an auto-regressive decoder, excels at open-ended conditional text generation, particularly suited for interpreting natural language questions.
Its advanced architectural features include RMSNorm pre-normalization\cite{zhang2019root}, SwiGLU activation functions inspired by PaLM\cite{chowdhery2022palm}, and rotary positional embeddings\cite{shaw2018self}. To extend its context comprehension, Llama-2 leverages a grouped-query attention mechanism\cite{ainslie2023gqa}, allowing it to process a significant number of 2048 tokens. The architecture, consisting of 32 layers, 32 attention heads, and a hidden size of 4096, efficiently handles batch sizes of up to 32 for sequences of up to 2048 tokens. We fine-tune Llama-2 using Parameter-Efficient Fine-Tuning (PEFT) methods, specifically employing the Dynamic Adaptation of Mixture of Quantized Parameter-Efficient Experts (DyA-MoQPEs) technique. This approach enhances Llama-2's performance on visual question answering (VQA) tasks related to electron micrograph analysis through Vision-Language Instruction Tuning. As a result, Llama-2 efficiently adapts its extensive language understanding capabilities to the specific context and nuances of niche domain topics, such as interpreting complex natural language queries related to electron micrographs. The fine-tuned Llama-2 model demonstrates a deeper understanding of end-user questions, effectively handling ambiguity and complex language for accurate image-text correspondence in VQA tasks, connecting textual concepts and entities with their visual counterparts in microscopic images. Our work integrates QDyA-MoPEs adapter modules into each linear layer of the grouped-query attention mechanism layers in the Llama 2-7B model architecture. These layers analyze different aspects of language understanding, with earlier layers focusing on fundamental syntactic elements and subsequent layers exploring more complex semantic connections. This integration allows for task-specific customization to effectively interpret natural language questions related to microscopic image analysis.

\vspace{-2mm}
\subsection{Generation of MultiModal Instruction-Tuning Data}
\vspace{-1mm}
We leverage GPT-4 Turbo with Vision, a state-of-the-art multimodal large language model (MLLM), to create a customized and comprehensive dataset of image-question answer pairs specifically designed for fine-tuning small multimodal models (SMMs) for visual question answering (VQA) tasks on electron micrographs. GPT-4 first generates challenging and contextually relevant questions that interpret and analyze these micrographs. Simultaneously, it utilizes knowledge distillation to produce corresponding answers using its internal knowledge representation grounded in the visual content of the microscopic images. These generated answers are enriched with domain-specific insights, ensuring accurate responses to open-ended user questions about electron micrographs. Our approach addresses the scarcity of high-quality vision-language datasets for analyzing microscopic images. By training SMMs using the generated vision-language instruction-following dataset, we enable them to acquire domain-specific adaptation abilities through transfer learning. This allows them to perform comparably to proprietary large multimodal models (LMMs) on VQA tasks without incurring excessive computational costs. Our approach offers a methodology for developing a highly efficient, accurate, and domain-specific framework to interpret complex microscopic images. It leverages multimodal intelligence, encompassing vision, language, and reasoning, to address these challenges. The compact multimodal models facilitate interaction between multiple modalities through joint representation learning. This process implicitly aligns semantic concepts across vision and language, enabling the smaller models to contextually understand and reason about these multimodal inputs in order to answer visual questions. This establishes a clear, concise, and relevant foundation for SMMs, allowing them to grasp the visual representation of concept-based instructions and their corresponding answers.
GPT-4 crafts questions to guide a comprehensive and thorough investigation of diverse facets, including fundamental characteristics like the size, distribution, and morphology of nanomaterials depicted in microscopic images, such as:

\vspace{-3mm}
\begin{tcolorbox}[colback=white!5!white,colframe=black!75!black]% [width=10cm]
\vspace{-2mm}
\textbf{Prompt 1:} **Basics** - This image depicts a nanomaterial. What specific type of nanomaterial is it? Additionally, what is the scale or resolution - that is, what real-world length does one unit of measurement in the image correspond to?. \textbf{Prompt 2:} **Morphology and Structure** - Can you describe the overall shape and morphology of the nanomaterials depicted in the image? - Are there any visible layers, phases, or distinct domains within the nanomaterials? - Do the nanomaterials exhibit a consistent size and shape throughout, or do they display variability in these aspects?.  \textbf{Prompt 3:} **Size and Distribution** - Can you estimate the approximate size or size range of the individual nanostructures depicted in the image? - Additionally, how are the nanomaterials distributed - are they evenly spaced, clustered, or randomly placed? - Finally, is there any visible evidence of aggregation or bundling among the nanostructures?". \textbf{Prompt 4:} **Surface Characteristics** - When examining the nanomaterials in the image, what are their surface textures like - are they predominantly smooth, rough, or do they possess distinct textures? - Additionally, are there any noticeable imperfections, such as defects, pores, or impurities, visible on the surfaces of these nanomaterials?.  \textbf{Prompt 5:} **Composition and Elements** - In the provided image, can we identify any evidence of compositional variations, such as changes in color, brightness, or contrast that might indicate different components? - Additionally, are there any discernible labels or markers within the image that specifically point to the presence of certain elements or compounds?.  \textbf{Prompt 6:} **Interactions and Boundaries** - Describe how the individual nanostructures visually interact with one another. For example, do they appear to be touching, fused together, or fully separate? - Examine the boundaries between nanostructures. Can you clearly distinguish boundaries between different structures or phases? - Or do they blend together without defined borders?.  
\vspace{-2mm}
\end{tcolorbox}

\vspace{-3mm}
\begin{tcolorbox}[colback=white!5!white,colframe=black!75!black]% [width=10cm]
\vspace{-2mm}
\textbf{Prompt 7:} **External Environment** - In the provided image, can you identify any signs of interaction between the nanomaterials and their surrounding environment or matrix, which might include solvents, polymers, or other materials? - Additionally, are there any discernible structures or objects present in the image that are not nanomaterials? If so, please describe these elements?. \textbf{Prompt 8:} **Image Technique and Modifications** - Can you identify the specific imaging technique, such as Scanning Electron Microscopy (SEM) or Transmission Electron Microscopy (TEM), used to capture this image of nanomaterials? - Additionally, were there any post-processing techniques or modifications applied, including but not limited to false coloring or 3D rendering?. \textbf{Prompt 9:} **Functional Features** - Can you identify any specific functional elements in the image, like active sites or regions with distinct properties? - Additionally, does the image depict any dynamic processes taking place within the subject, or is it primarily a static representation?.  \textbf{Prompt 10:}  **Context and Application** - What is the primary intended use or application of the nanomaterials as depicted in the image, and is the representation of these nanomaterials based on actual experimental samples, or are they theoretical or simulation-based representations?
\vspace{-2mm}
\end{tcolorbox}

\vspace{-3mm}
\subsection{Sampling Strategies for Instruction Tuning Dataset Generation}
\vspace{-1mm}
To generate instruction-tuning data using GPT-4 Turbo with vision for few-shot image classification tasks (refer to Figure \ref{fig:figure4}), and to discover key insights into high intra-class dissimilarity, high inter-class similarity, and spatial heterogeneity in electron micrographs (refer to Figures \ref{fig:figure5} - \ref{fig:figure7}), we follow the strategies outlined below. Given an input image $\mathbf{I}$ as a 3D tensor with dimensions $H \times W \times C$ (height, width, and number of channels, respectively), we divide it into non-overlapping patches of size $P \times P \times C$. This results in $n = \left(\frac{HW}{P^2}\right)$ patches. Each patch of size $P^2C$ is then encoded into a 1D vector, resulting in an encoded patch matrix $\mathbf{I'} \in \mathbb{R}^{n \times d}$, where $d$ represents the embedding dimension. To incorporate spatial information, we add positional embeddings to these patch embeddings. Additionally, we introduce a classification token, $<\hspace{-1mm}\textit{cls}\hspace{-1mm}>$, to represent the global image characteristics. The augmented patch sequence, including the classification token, is processed by the Vision Transformer (ViT \cite{dosovitskiy2020image}), which refines the patch representations through multiple encoder layers. We update the trainable parameters through a supervised learning task, aiming to minimize cross-entropy loss and maximize multiclass classification accuracy. Consequently, the output embedding $h_{\textit{cls}}$ corresponding to the $<\hspace{-1mm}\textit{cls}\hspace{-1mm}>$ token encapsulates a comprehensive representation of the microscopic image. We propose a similarity-driven sampling approach for the few-shot image classification task, based on the hypothesis that training on demonstrations resembling the target image's data distribution promotes model adaptability and accuracy. This method utilizes cosine similarity of the $h_{\textit{cls}}$ token embeddings to select the top-K most similar images to the target image from the training set. We follow the same strategy for sampling highly similar images across different nanomaterial categories to generate question-answer pairs, aiming to gain insights into high inter-class similarity. Conversely, we employ the inverse strategy for generating question-answer pairs for each target image, extracting insights on high intra-class dissimilarity by sampling highly dissimilar images within the same nanomaterial category.

\vspace{-2mm}
\subsection{Loss Functions}

\vspace{-1mm}
\subsubsection{Image-Text Matching loss (ITM)}
The ITM loss is fundamental to multimodal learning. It utilizes binary cross-entropy loss to enhance the alignment of image and text representations within a shared embedding space. Minimizing the LM loss allows the image-grounded text encoder to effectively determine whether an image and text pair are a match, thereby improving the alignment between image and text representations. For each image-text pair, a ground truth label, ($y_i$), is assigned, where ($y_i$ = 1) indicates a match (the image and text are relevant to each other), and ($y_i$ = 0) signifies a non-match. The encoder predicts the probability, ($p_i$), that each pair is a match. This probability is computed from the output of the encoder's final linear layer through a sigmoid function. The ITM loss is calculated using the binary cross-entropy loss as follows:

\vspace{-5mm}
\resizebox{0.95\linewidth}{!}{
\begin{minipage}{\linewidth}
\begin{align*}
L_{\text{ITM}} = -\frac{1}{b} \sum_{i=1}^{b} \hspace{1mm} \left[ \hspace{1mm} y_i \log(p_i) + (1 - y_i) \log(1 - p_i) \hspace{1mm} \right ]
\end{align*}
 \end{minipage}
}

\vspace{-1mm}
where $b$ represents the batch size. This approach ensures a balanced consideration of both matching and non-matching pairs in the loss calculation. It penalizes the encoder for incorrect predictions, thereby guiding it towards more precise representations for image-text matching pairs.

\vspace{-1mm}
\subsubsection{Language modeling loss (LM)} 
In VQA and image captioning tasks, minimizing LM loss is crucial. It ensures the image-grounded text decoder generates accurate and descriptive textual descriptions of the visual content, tailored to the corresponding end-user questions. Optimizing for LM loss in VQA and image captioning tasks discourages the model from relying solely on the question's linguistic patterns. This prevents language bias and promotes the inclusion of relevant visual information in the generated descriptions. This ensures the decoder learns the correct grammatical structure and vocabulary for answer sentences, resulting in texts that are not only coherent but also contextually aligned with both the image and the question. 
The framework is trained to enhance word prediction accuracy in a text sequence. It considers both preceding words and the visual context to refine its ability to interpret and respond to image-based queries. This involves minimizing the negative log-likelihood of the actual words, based on the predicted probabilities from the decoder, ultimately leading to grammatically correct, semantically coherent, and contextually appropriate answers. The LM loss is defined as follows:

\vspace{-5mm}
\resizebox{0.925\linewidth}{!}{
\begin{minipage}{\linewidth}
\begin{align*}
L_{LM} = -\sum_{i}^{N} \hspace{1mm} \log P(w_i | w_{<i}, I, Q) 
\end{align*}
 \end{minipage}
}

\vspace{-1mm}
Where $L_{LM}$ represents the language modeling loss, and $N$ is the number of words in the text. The term $w_i$ denotes the $i$-th word in the text, while $w_{<i}$ represents the sequence of all words preceding the $i$-th word. $I$ is the target image. $Q$ refers to the natural language question that the generated text aims to answer, conditioned on both the image $I$ and the previous words $w_{<i}$ in the sequence. The expression $P(w_i | w_{<i}, I, Q)$ represents the probability of the $i$-th word, given the preceding words, the target image, and the end-user question, as predicted by the model. At inference time, the decoder uses the knowledge acquired during training, such as the relationships between words, images, and end-user questions, to generate accurate text descriptions for a given image.

\vspace{-2mm}
\subsection{Additional Information}
\vspace{-1mm}
We train small-scale vision-language models on the electron micrographs analysis using an instruction-following dataset generated by GPT-4 Turbo with Vision through a teacher-student strategy. The robustness and effectiveness of these small-scale models depend on the composition and design of the training dataset, particularly the comprehensiveness and detail of the image-question-answer pairs. In this work, we propose a novel approach that leverages a balanced combination of concise, summarized answers and more comprehensive, detailed responses in training datasets for the same end-user questions. This method optimizes the performance of small-scale vision-language models across a range of tasks, from image captioning to complex visual question answering (VQA) tasks. Utilizing training data of varied lengths in the small-scale model training offers numerous advantages. It enhances flexibility and adaptability by exposing the small-scale model to diverse sentence structures and visual complexities, thus improving its ability to handle real-world scenarios with varying levels of detail. This approach improves generalizability and prevents overfitting to specific data patterns. Moreover, it challenges the small-scale model's reasoning and attention mechanisms, promoting a deeper understanding of the relationships between visual features and textual descriptions. These benefits lead to improved performance in tasks such as image captioning and VQA tasks, making the small-scale model more robust and versatile for practical applications. 

\vspace{-1mm}
\subsection{Experimental Setup} % 
\vspace{-1mm}
In this work, we propose a novel framework utilizing a teacher-student paradigm. A large multimodal model like GPT-4 Turbo with vision acts as a teacher to generate instruction-following data to train a smaller, specialized student model, called \texttt{sLAVA}, specifically designed for zero/few-shot image classification, image captioning, and VQA tasks in electron microscopy image analysis. It leverages the vision-language instruction-tuning approach to efficiently transfer knowledge from the larger to smaller model, enabling the student model to perform comparably to larger models in terms of generating accurate and contextually relevant responses to end-user questions based on input images. Additionally, \texttt{sLAVA} is better suited for on-premises enterprises adoption, ensuring data privacy and security. The \texttt{sLAVA} framework is a small-scale, visually conditioned autoregressive language generation model designed for micrograph analysis. It consists of a vision encoder that analyzes microscopic images, while a text encoder interprets end-user questions. The cross-attention mechanism in the image-grounded text encoder enables the small-scale model to effectively align multimodal information, facilitating accurate answer generation. The small-scale model then leverages this integrated multimodal understanding to generate accurate and contextually relevant answers or image captions.  The generated text is not only factually accurate but also contextually aligned with the specifics of the electron microscopy images. The small-scale model focuses on both zero/few-shot settings, using multi-modal prompts as inputs consisting of a microscopic image, supplementary image information, and the natural language question for precise analysis and response.  The framework adopts a bi-objective approach, optimizing both understanding-based and generation-based goals to improve performance in microscopic image-based analysis on the image captioning and VQA tasks. We trained the \texttt{sLAVA} framework using the tailored image-question-answer pairs dataset generated by GPT-4 based on the SEM dataset \cite{aversa2018first}, a collection of high-resolution images ($1024 \times 768 \times 3$) showcasing diverse nanomaterials. For preprocessing, we resized the images to $224 \times 224 \times 3$ and applied data standardization to normalize the mean and variance across channels to 0.5, constraining values between -1 and 1. To effectively capture local features, we divided the resized images ($224 \times 224 \times 3$) into the 32-pixel patches, representing each micrograph as a sequence of patches with an embedding dimension of 64. This patch-wise approach enabled the model to learn local features while retaining contextual information through patch sequences. This ultimately enhances the proficiency of the \texttt{sLAVA} framework to understand and analyze complex nanomaterial images. Parameter-efficient fine-tuning (PEFT) of the Llama-2-7b model utilizes the Dynamic Adaptation of Mixture of Quantized Parameter-Efficient Experts (DyA-MoQPEs) technique. A key hyperparameter, rank ($r$), controls the trade-off between the language model's capacity to learn complex data patterns and its overall complexity (number of trainable parameters) through low-rank approximation of weight matrices. During fine-tuning, we randomly sample $r$ from $[r_{min}=4, r_{max}=16]$, with higher values enabling more expressive fine-tuned langauge models with increased adaptable parameters, while lower ranks reduce complexity. (b) Alpha ($\alpha$) – a scaling factor applied to the low-rank weight matrix updates, typically set to a small value like $\frac{1}{r}$ based on the sampled rank. Alpha modulates the step size of the updates, with larger values allowing more aggressive adaptation, improving performance but potentially causing instability. (c) LoRA Dropout – applying dropout specifically to the low-rank adapter layers to prevent overfitting and improve generalization, usually set to 0.05. Additionally, we employ 8-bit quantization to enable efficient fine-tuning on consumer hardware while retaining comparable performance. We trained the \texttt{sLAVA} framework over 50 epochs using an initial learning rate of $10^{-3}$ and a batch size of 32 for controlled optimization. For the self-attention and cross-attention layers, we set the number of heads ($H$) to 4 and the key/query/value dimensionality ($d_h$) to 32. To optimize performance, we implemented two key strategies: (a) Early stopping based on the validation loss to prevent overfitting; and (b) A learning rate scheduler that reduces the rate by half if the validation loss plateaus for 5 consecutive epochs, assisting convergence. Additionally, we employed the Adam optimization algorithm \cite{kingma2014adam} to update parameters. Our instruction-following image-question-answer pairs dataset comprises three types: (a) zero-shot/few-shot multiclass classification tasks, (b) image captioning, and (c) visual question answering (VQA). During training, we minimize both the binary cross-entropy loss and the language modeling loss to update the trainable parameters of the framework. High-performance Nvidia V100 GPUs facilitated development and testing of the custom \texttt{sLAVA} framework. Rigorous optimization with early stopping and learning rate adjustments ensured a balance between expressiveness and overfitting, maximizing real-world performance for multimodal image analysis guided by natural language.

\vspace{-2mm}
\subsection{Evaluation Metrics}
\vspace{-1mm}
Image Captioning and VQA tasks combine computer vision and natural language processing to answer image-based questions. Evaluating the accuracy of these answers is challenging, but evaluation metrics assess linguistic similarity, grammatical correctness, and semantic relevance between the ground-truth and generated answers, driving the framework towards more human-like and accurate responses. Here's an overview of some key metrics:

\vspace{-2mm}
\begin{itemize}
\item BLEU Score (Bilingual Evaluation Understudy): The BLEU Score measures the quality of machine-generated text by comparing it with a reference translation (ground-truth). It analyzes the frequency of overlapping word sequences (n-grams) between the two texts. The focus of BLEU is on precision; it counts matching n-grams while preventing an overemphasis on repeated phrases. BLEU scores range from 0, indicating no overlap, to 1, indicating a perfect match. Higher scores signify a greater shared vocabulary and similarity in phrasing.
\item METEOR (Metric for Evaluation of Translation with Explicit Ordering): The METEOR metric evaluates machine-generated text by comparing it to the ground truth, focusing on word similarity. It considers synonyms, paraphrases, and variations of words. METEOR prioritizes exact matches, lemmas, stems, and semantic similarities, capturing both recall and precision on a 0-1 scale. Higher scores indicate greater similarity to the reference translation. While BLEU focuses on how often short phrases (n-grams) appear in the translation, METEOR provides a more comprehensive evaluation by including fluency, grammar, and semantic matching. This allows it to correlate better with human judgment of translation quality.
\item ROUGE Score (Recall-Oriented Understudy for Gisting Evaluation): ROUGE measures the quality of machine-generated text by comparing its lexical overlap with ground-truth. ROUGE-N, the basic metric, counts matching n-grams between the candidate and reference texts. Variants like ROUGE-L, ROUGE-W, and ROUGE-S focus on longest common subsequences, word sequences, and skip-bigrams, respectively. Scores range from 0, indicating no overlap, to 1 for complete lexical identity. Higher scores suggest better quality, indicating content similar to human references. While ROUGE primarily evaluates lexical similarity, variants such as ROUGE-L correlate well with human judgments of linguistic quality and coherence.
\end{itemize}

\vspace{-2mm}
\subsection{Empirical Insights into Nanomaterial Classification}
\vspace{-1mm}
Our research thoroughly evaluated the proposed framework \texttt{sLAVA} for classifying electron micrographs of diverse nanomaterials. These complex materials vary in composition, morphology, structure, and other properties, which is evident in their electron micrographs. The framework achieved high accuracy on the imbalanced SEM dataset\cite{aversa2018first} using metrics like precision, recall, and F1-score, demonstrating its effectiveness in categorizing nanomaterials with different patterns in a zero-/few-shot setting. Table \ref{tab:mccategory} reports the experimental results. The multi-metric approach provided a detailed analysis, highlighting \texttt{sLAVA}'s efficiency in handling various categories, especially those with fewer labeled instances. Overall, our findings confirm sLAVA's robustness in classifying nanomaterials, contributing to advancements in materials characterization and research.

\subsection{Additional Results}
\begin{figure*}[ht!]
\centering
\resizebox{0.945\linewidth}{!}{ 
\includegraphics[keepaspectratio,height=4.5cm,trim=0.0cm 0.0cm 0cm 0.0cm,clip]{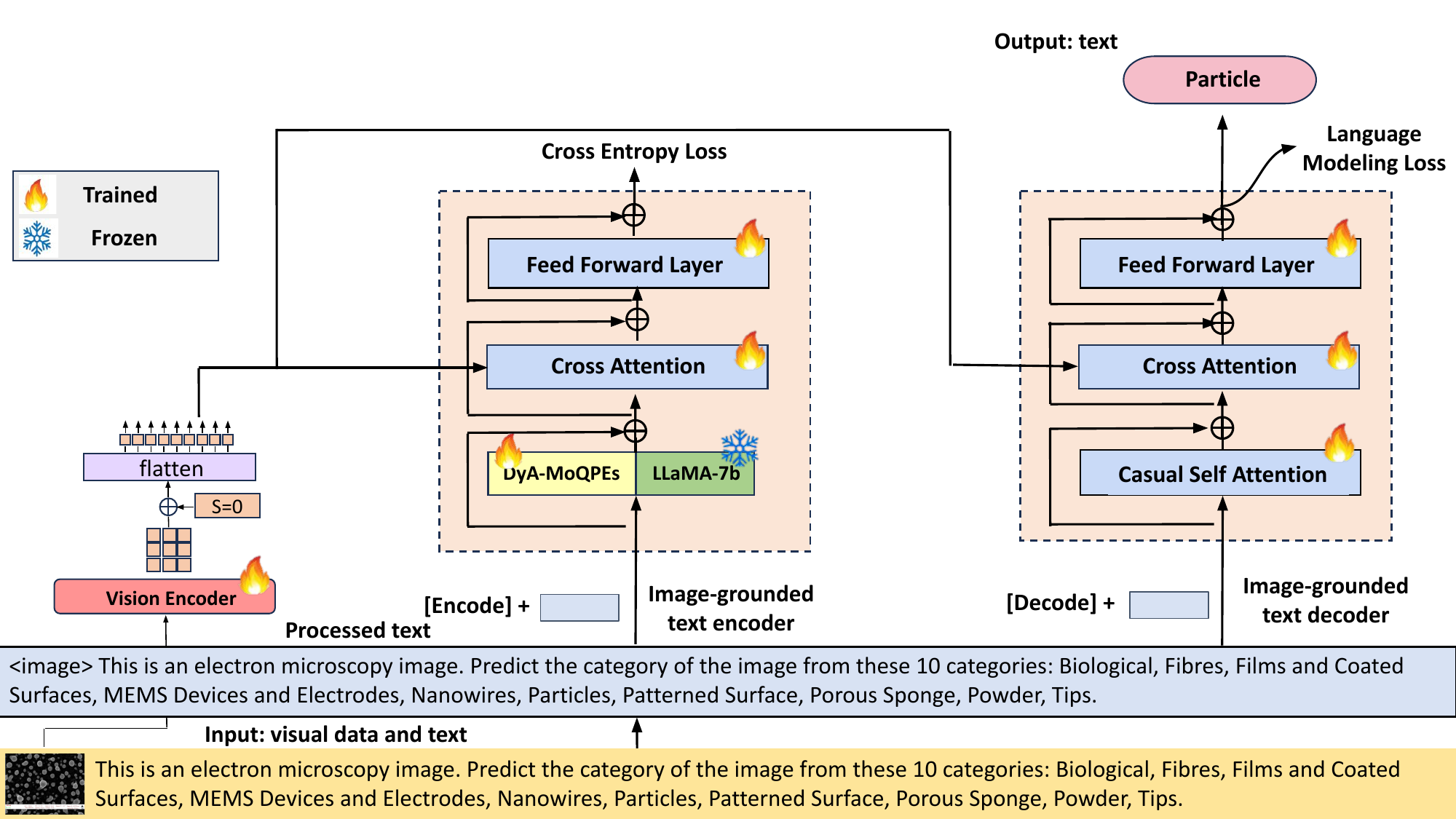} % left, bottom, right, top
}
\caption{The schematic depicts a variant of \texttt{sLAVA} (small-scale, language-and-vision assistant), a family of visually-conditioned, autoregressive text generation model. The small-scale vision-and-language model take as input a multimodal prompt consisting of the target electron micrographs and user-provided auxiliary text, along with the user question. The model then generates free-form text to answer end-user questions. The task is to categorize the image into one of ten categories, such as biological fibers and films, in a zero-shot setting.}
\label{fig:figure3}
\end{figure*}

\begin{figure*}[ht!] 
\centering
\resizebox{0.945\linewidth}{!}{
\includegraphics[keepaspectratio,height=4.5cm,trim=0.0cm 0.0cm 0cm 1.675cm,clip]{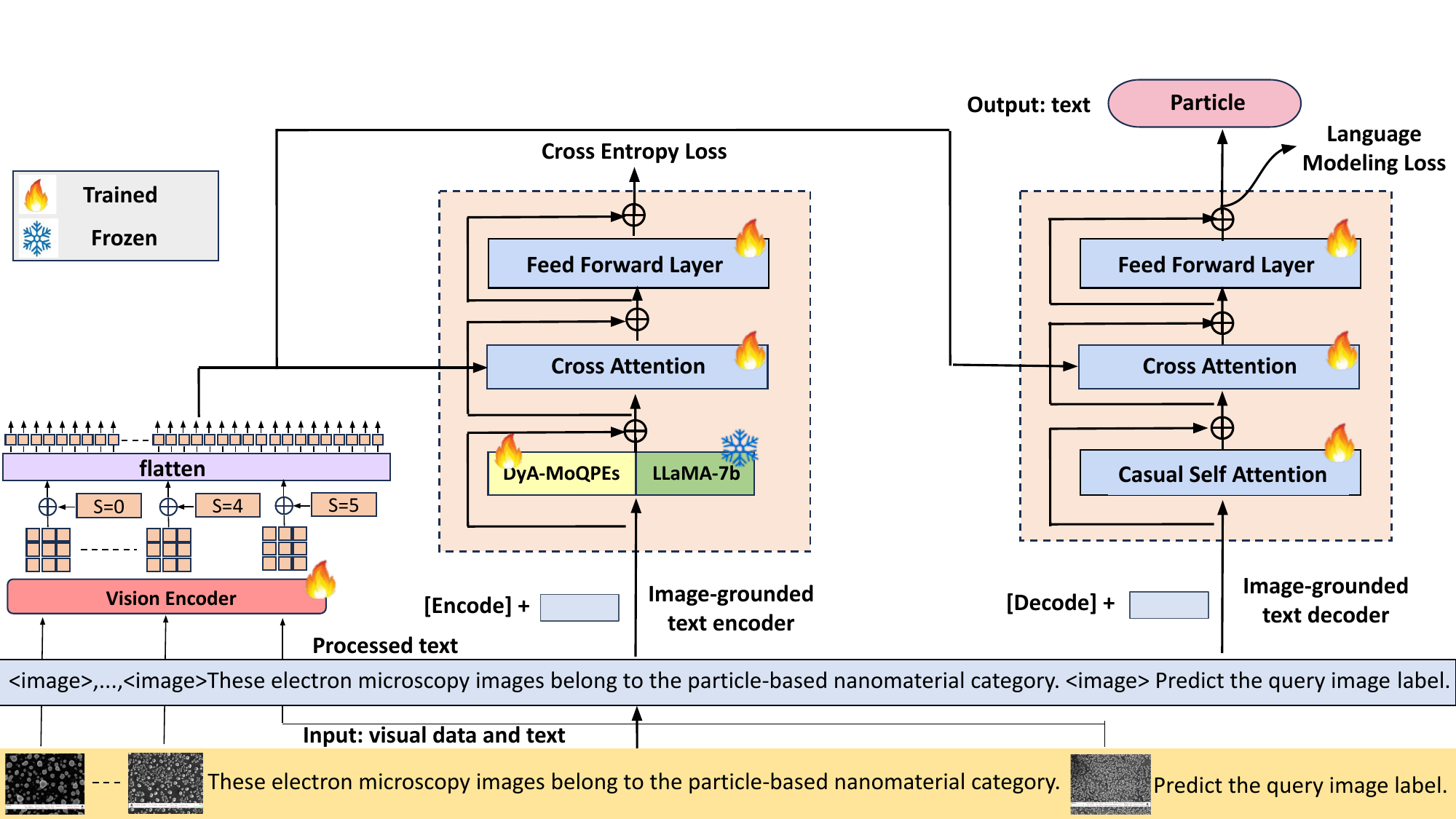} % left, bottom, right, top
}
\caption{The schematic depicts a variant of \texttt{sLAVA}, a small-scale language-and-vision assistant. It takes a multimodal prompt consisting of electron micrographs, interspersed arbitrarily with text, as input and generates free-form text as output. The input consists of a series of electron microscopy images, their corresponding ground-truth labels, and a task-specific instruction. In a few-shot setting, the objective is to predict the label for the target image.}
\label{fig:figure4}
\vspace{0mm}
\end{figure*}

\begin{figure*}[ht!]
\centering
\resizebox{0.875\linewidth}{!}{ 
\hspace*{0mm}\includegraphics[keepaspectratio,height=4.5cm,trim=0.0cm 0.0cm 0cm 0.1cm,clip]{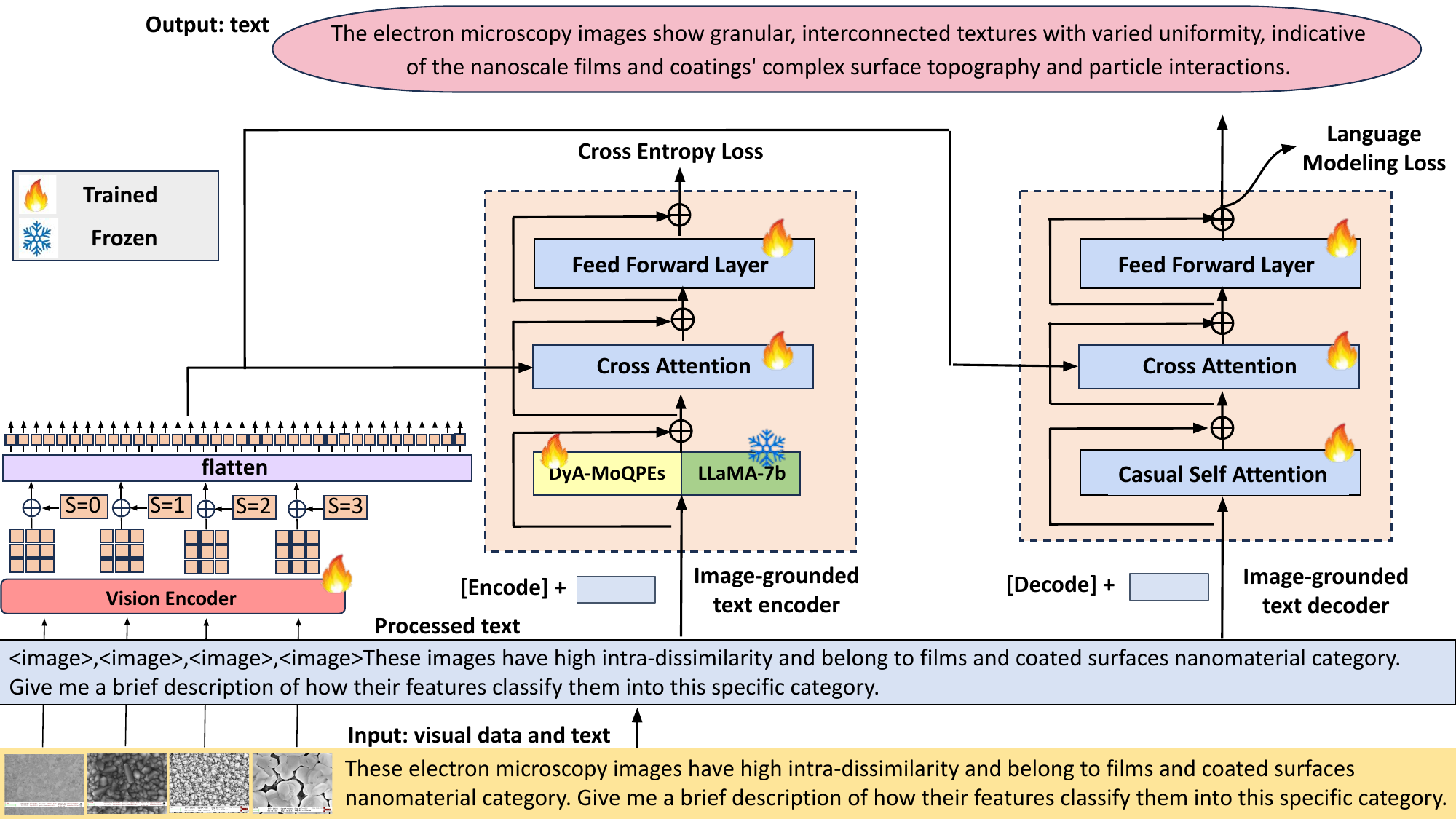} % left, bottom, right, top
}
\vspace{-2mm}
\caption{The schematic showcases a variant of \texttt{sLAVA}, a proposed small-scale language-and-vision assistant, which takes as input a multimodal prompt comprising the electron microscopy images and their corresponding supplementary text descriptions. The small-scale model's objective is to generate concise and accurate descriptions explaining how visual features in these high-contrast images determine their classification into specific nanomaterial categories. During the inference stage, \texttt{sLAVA} draws upon its pre-trained knowledge and domain-specific expertise to produce informative and accurate responses to the end-user's questions for unseen microscopic images within that category.}
\label{fig:figure5}
\vspace{-6mm}
\end{figure*}

\begin{figure*}[ht!]
\centering
\resizebox{0.875\linewidth}{!}{ 
\hspace*{0mm}\includegraphics[keepaspectratio,height=4.5cm,trim=0.0cm 0.0cm 0cm 0.0cm,clip]{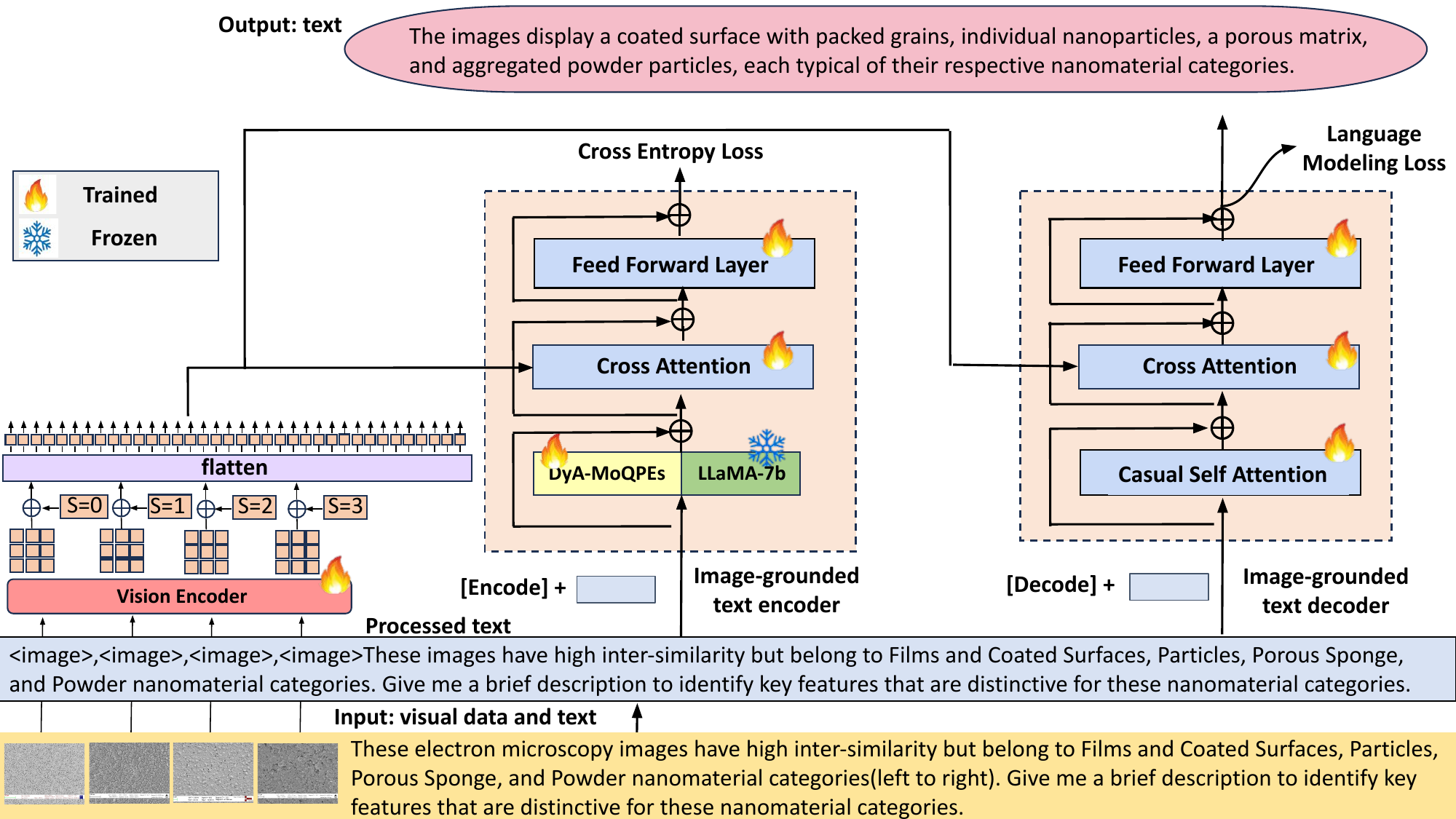} % left, bottom, right, top
}
\vspace{-2mm}
\caption{The schematic illustrates a variant of \texttt{sLAVA}, a proposed small-scale language-and-vision assistant specifically designed for electron microscopy image Visual Question Answering (VQA) tasks. The model takes in multimodal input: a sequence of similar-looking, high-resolution electron microscopy images showcasing diverse nanomaterial categories such as Films and Coated Surfaces, Particles, Porous Sponges, and Powders as well as the auxiliary text information. Additionally, \texttt{sLAVA} receives an end-user question that prompts it to analyze and describe the unique visual features distinguishing each category, thereby generating precise and concise responses.}
\label{fig:figure6}
\end{figure*}

\begin{figure*}[ht!]
\centering
\resizebox{0.875\linewidth}{!}{ 
\hspace*{0mm}\includegraphics[keepaspectratio,height=4.5cm,trim=0.0cm 0.0cm 0cm 0.0cm,clip]{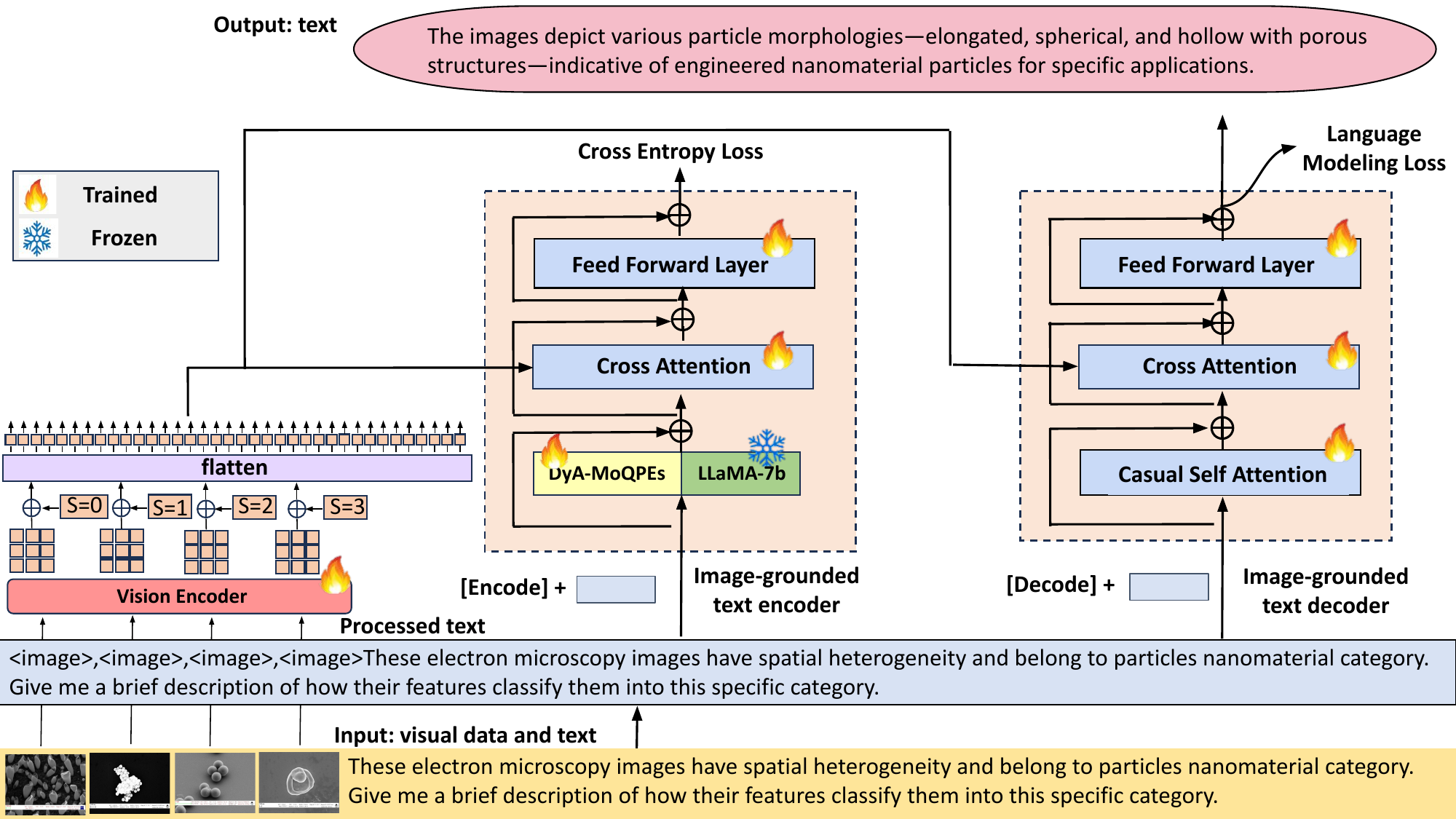} % left, bottom, right, top
}
\vspace{-3mm}
\caption{The schematic outlines the architecture of the small-scale language-and-vision assistant (\texttt{sLAVA}), which is tailored for the analysis of electron microscopy images of nanomaterials. Its multimodal input comprises a series of high-contrast electron microscopy images that showcase spatial heterogeneity and represent diverse particle morphologies. A task-specific directive instructs the multimodal model to generate accurate and concise descriptions, identifying and differentiating the visual characteristics that are distinctive to the nanomaterial category represented in each image.}
\label{fig:figure7}
\vspace{-5mm}
\end{figure*}

The Figures~\ref{fig:figure3},~\ref{fig:figure4},~\ref{fig:figure5},~\ref{fig:figure6}, and~\ref{fig:figure7} illustrate the small-scale, language-and-vision assistant (\texttt{sLAVA}). \texttt{sLAVA} belongs to a family of small multimodal models (SMMs) that take electron micrographs and supporting image information as input and produce free-form text output in response to end-user questions. Figures~\ref{fig:figure3} and~\ref{fig:figure4} show variants of the sLAVA framework on the zero/few-shot classification task. Tables~\ref{tab:tableclass1} and~\ref{tab:tableclass2} show the experimental results on the zero/few-shot multiclass classification task, comparing the accuracy of our proposed framework to several baseline algorithms.  Table~\ref{captioning_results2} shows the framework's performance on the open-ended VQA task. Unlike closed-ended VQA, which requires choosing the correct answer from a set of predefined options, open-ended VQA tasks require the small-scale model to generate its own free-form responses to end-user questions.
Table~\ref{VQA2} displays electron microscope images with their true captions and small-scale model generated captions. It additionally includes BLEU-2, ROUGE-L, and METEOR scores that evaluate the similarity of the small-scale model's generated captions to the correct captions. Tables~\ref{tab:tab0} to~\ref{tab:tab9} display samples from the instruction-tuning Q\&A pairs dataset, which was generated by GPT-4 Turbo with Vision for training the smaller multimodal model, \texttt{sLAVA}. Figure~\ref{fig:figure5},~\ref{fig:figure6}, and~\ref{fig:figure7} show variants of the sLAVA framework for the VQA task, addressing high intra-class dissimilarity, high inter-class similarity, and spatial heterogeneity in electron micrographs, respectively. Tables~\ref{captioning_results3},~\ref{captioning_results4}, and~\ref{captioning_results5} summarize the performance of various methods on the aforementioned VQA task.

\subsection{Related Work}
Large Language Models (LLMs) like Open AI ChatGPT\cite{gpt4v}, Google Gemini\cite{team2023gemini} have significantly advanced natural language processing by demonstrating remarkable abilities in understanding and generating human-like text. Building on this progress, Multi-modal Large Language Models (MLLMs) like MiniGPT-4\cite{zhu2023minigpt}, LLaVA\cite{liu2023visual}, and InstructBLIP\cite{dai2305instructblip} have emerged, integrating visual understanding with linguistic capabilities. These MLLMs, often based on open-source LLMs like LLaMA\cite{touvron2023llama} and Qwen\cite{bai2023qwen}, can process and interpret both text and images, leading to a more holistic comprehension of complex questions that require analysis of both modalities. InstructBLIP\cite{dai2305instructblip} is an advanced vision-language model that utilizes instruction tuning and components such as an image encoder, a large language model (LLM), and a Query Transformer (Q-Former) to improve its effectiveness across various multimodal tasks, including image captioning and visual question answering. Based on the BLIP-2\cite{li2023blip} framework, InstructBLIP emphasizes adaptability and efficiency, leveraging frozen components during training to optimize learning from diverse instructional datasets. This architecture supports a wide range of tasks, demonstrates strong zero-shot performance, and integrates with multiple datasets, positioning it as a robust and scalable option for research and application in multimodal machine learning. MiniGPT-4\cite{zhu2023minigpt} advance vision-language models by aligning a pretrained vision encoder with the Vicuna large language model\cite{chiang2023vicuna} using a single linear projection layer. This integration enables direct processing of visual data, enhancing the model's ability to handle complex tasks. The model's development includes a two-stage training process, starting with pretraining on 5 million image-text pairs to learn vision-language interactions, followed by a fine-tuning stage with a curated high-quality dataset to improve language outputs and usability. MiniGPT-4's capabilities, such as generating detailed image descriptions, creating websites from sketches, and composing stories from images, match or exceed those of GPT-4. The LLaVA model\cite{liu2023visual} represents a significant advancement in vision-language integration, leveraging a two-stage training process that initially adapt large language models (LLMs) to visual inputs through pre-training on extensive image-text pairs, followed by fine-tuning them on visual instructions. This approach, enhanced by the use of multi-layer perceptrons (MLP) instead of traditional linear projections, significantly improves the model's multimodal capabilities. Additionally, LLaVA model incorporates a Mixture of Experts (MoE) strategy, which optimizes processing by assigning specialized experts to handle different types of data, thus reducing redundancy and boosting efficiency in task-specific contexts. These architectural and methodological enhancements enable LLaVA model to excel across a variety of benchmarks, demonstrating superior performance in complex visual and language tasks.

\vspace{2mm} 

\vspace{0mm}

\vspace{-6mm}

\vspace{0mm}

\vspace{-2mm}
\begin{table*}[!ht]
\centering
\caption{Table shows the performance of \texttt{sLAVA} compared to baselines on open-ended VQA task.}
\vspace{0mm}
\scalebox{0.80}{
\begin{tabular}{l|c|c|c|c|c|c}
\toprule
Method                                                                  & BLEU-2 ($\uparrow$)                     & BLEU-4 ($\uparrow$)                     & ROUGE-1 ($\uparrow$)                    & ROUGE-2 ($\uparrow$)                    & ROUGE-L ($\uparrow$)                   & METEOR ($\uparrow$)                     \\ \midrule
\begin{tabular}[c]{@{}l@{}} InstructBLIP\cite{dai2305instructblip} \end{tabular}            & 0.715$\pm$0.063 & 0.580$\pm$0.078 & 0.820$\pm$0.032 & 0.721$\pm$0.011 & 0.777$\pm$0.042 & 0.835$\pm$0.048          \\ \midrule
\begin{tabular}[c]{@{}l@{}}LLaVA\cite{liu2023visual} \end{tabular}         & 0.722$\pm$0.070 & 0.588$\pm$0.085 & 0.821$\pm$0.032 & 0.724$\pm$0.011 & 0.779$\pm$0.042 & 0.836$\pm$0.046          \\ \midrule
\begin{tabular}[c]{@{}l@{}}MiniGPT-4\cite{zhu2023minigpt} \end{tabular}     & 0.746$\pm$0.075 & 0.607$\pm$0.090 & 0.836$\pm$0.033 & 0.737$\pm$0.012 & 0.792$\pm$0.043 & 0.855$\pm$0.047          \\ \midrule
\begin{tabular}[c]{@{}l@{}}\textbf{sLAVA} \end{tabular} &  \textbf{0.830 $\pm$0.085} & \textbf{0.757 $\pm$0.105} & \textbf{0.936 $\pm$0.036} & \textbf{0.813 $\pm$0.014} & \textbf{0.864 $\pm$0.050} & \textbf{0.914 $\pm$0.055}      \\  \bottomrule
\end{tabular}
}
\label{captioning_results2}
\vspace{-2mm}
\end{table*}

\vspace{-2mm}
\begin{table*}[!ht]
\centering
\caption{The table shows \texttt{sLAVA} excels on VQA task on high intra-dissimilarity of nanomaterials.}
\vspace{0mm}
\scalebox{0.80}{
\begin{tabular}{l|c|c|c|c|c|c}
\toprule
Method                                                                  & BLEU-2 ($\uparrow$)                     & BLEU-4 ($\uparrow$)                     & ROUGE-1 ($\uparrow$)                    & ROUGE-2 ($\uparrow$)                    & ROUGE-L ($\uparrow$)                   & METEOR ($\uparrow$)                     \\ \midrule
\begin{tabular}[c]{@{}l@{}} InstructBLIP\cite{dai2305instructblip} \end{tabular}            & 0.677$\pm$0.063 & 0.549$\pm$0.078 & 0.776$\pm$0.032 & 0.682$\pm$0.011 & 0.735$\pm$0.042 & 0.790$\pm$0.048          \\ \midrule
\begin{tabular}[c]{@{}l@{}}LLaVA\cite{liu2023visual} \end{tabular}         & 0.661$\pm$0.070 & 0.538$\pm$0.085 & 0.751$\pm$0.032 & 0.662$\pm$0.011 & 0.713$\pm$0.042 & 0.765$\pm$0.046          \\ \midrule
\begin{tabular}[c]{@{}l@{}}MiniGPT-4\cite{zhu2023minigpt} \end{tabular}     & 0.683$\pm$0.075 & 0.556$\pm$0.090 & 0.765$\pm$0.033 & 0.674$\pm$0.012 & 0.725$\pm$0.043 & 0.782$\pm$0.047          \\ \midrule
\begin{tabular}[c]{@{}l@{}}\textbf{sLAVA} \end{tabular} &  \textbf{0.759 $\pm$0.085} & \textbf{0.692 $\pm$0.105} & \textbf{0.856 $\pm$0.036} & \textbf{0.743 $\pm$0.014} & \textbf{0.790 $\pm$0.050} & \textbf{0.836 $\pm$0.055}      \\  \bottomrule
\end{tabular}
}
\label{captioning_results3}
\vspace{-4mm}
\end{table*}

\vspace{-2mm}
\begin{table*}[!ht]
\centering
\caption{The table shows \texttt{sLAVA} excels on VQA task on high inter-similarity of nanomaterials.}
\vspace{0mm}
\scalebox{0.80}{
\begin{tabular}{l|c|c|c|c|c|c}
\toprule
Method & BLEU-2 ($\uparrow$) & BLEU-4 ($\uparrow$) & ROUGE-1 ($\uparrow$) & ROUGE-2 ($\uparrow$) & ROUGE-L ($\uparrow$) & METEOR ($\uparrow$) \\ 
\midrule
\begin{tabular}[c]{@{}l@{}} InstructBLIP\cite{dai2305instructblip} \end{tabular} & 0.686$\pm$0.063 & 0.556$\pm$0.078 & 0.787$\pm$0.032 & 0.692$\pm$0.011 & 0.745$\pm$0.042 & 0.801$\pm$0.048 \\ 
\midrule
\begin{tabular}[c]{@{}l@{}}LLaVA\cite{liu2023visual} \end{tabular} & 0.685$\pm$0.070 & 0.558$\pm$0.085 & 0.779$\pm$0.032 & 0.687$\pm$0.011 & 0.741$\pm$0.042 & 0.794$\pm$0.046 \\ 
\midrule
\begin{tabular}[c]{@{}l@{}}MiniGPT-4\cite{zhu2023minigpt} \end{tabular} & 0.701$\pm$0.075 & 0.570$\pm$0.090 & 0.785$\pm$0.033 & 0.692$\pm$0.012 & 0.744$\pm$0.043 & 0.803$\pm$0.047 \\ 
\midrule
\begin{tabular}[c]{@{}l@{}}\textbf{sLAVA} \end{tabular} & \textbf{0.771$\pm$0.085} & \textbf{0.704$\pm$0.105} & \textbf{0.871$\pm$0.036} & \textbf{0.756$\pm$0.014} & \textbf{0.803$\pm$0.050} & \textbf{0.85002$\pm$0.055} \\  
\bottomrule
\end{tabular}
}
\label{captioning_results4}
\vspace{-4mm}
\end{table*}

\vspace{-2mm}
\begin{table*}[!ht]
\centering
\caption{The table shows \texttt{sLAVA} excels on VQA task related to nanomaterials' spatial heterogeneity.}
\vspace{0mm}
\scalebox{0.80}{
\begin{tabular}{l|c|c|c|c|c|c}
\toprule
Method                                                                  & BLEU-2 ($\uparrow$)                     & BLEU-4 ($\uparrow$)                     & ROUGE-1 ($\uparrow$)                    & ROUGE-2 ($\uparrow$)                    & ROUGE-L ($\uparrow$)                   & METEOR ($\uparrow$)                     \\ \midrule
\begin{tabular}[c]{@{}l@{}} InstructBLIP\cite{dai2305instructblip} \end{tabular}            & 0.623$\pm$0.055 & 0.504$\pm$0.068 & 0.714$\pm$0.028 & 0.628$\pm$0.010 & 0.677$\pm$0.037 & 0.727$\pm$0.042          \\ \midrule
\begin{tabular}[c]{@{}l@{}}LLaVA\cite{liu2023visual} \end{tabular}         & 0.629$\pm$0.061 & 0.511$\pm$0.074 & 0.715$\pm$0.028 & 0.631$\pm$0.010 & 0.679$\pm$0.037 & 0.728$\pm$0.040          \\ \midrule
\begin{tabular}[c]{@{}l@{}}MiniGPT-4\cite{zhu2023minigpt} \end{tabular}     & 0.650$\pm$0.066 & 0.529$\pm$0.079 & 0.728$\pm$0.029 & 0.642$\pm$0.010 & 0.691$\pm$0.037 & 0.745$\pm$0.041          \\ \midrule
\begin{tabular}[c]{@{}l@{}}\textbf{sLAVA} \end{tabular} &  \textbf{0.723$\pm$0.074} & \textbf{0.660$\pm$0.092} & \textbf{0.816$\pm$0.031} & \textbf{0.709$\pm$0.012} & \textbf{0.754$\pm$0.044} & \textbf{0.797$\pm$0.048}      \\  \bottomrule
\end{tabular}
}
\label{captioning_results5}
\vspace{-6mm}
\end{table*}

\vspace{-3mm}
\begin{table}[ht!]
\footnotesize
\centering
\setlength{\tabcolsep}{5pt}
\caption{Table shows the performance comparisons: our method vs. Convolutional Neural Networks(ConvNets), Vision Transformers (ViTs), \& Vision self-supervised learning(VSL) algorithms for multi-class classification task.} 
\vspace{2mm}
\label{tab:tableclass1}
\begin{tabular}{c|c|c|c}
\toprule
\multicolumn{2}{c|}{\textbf{Algorithms}} & \textbf{Top-1} & \textbf{Top-5} \\ 
\midrule
\multirow{6}{*}{\rotatebox[origin=c]{90}{\textbf{ConvNets}}} 
& AlexNet(\cite{krizhevsky2017imagenet}) & 0.528 & 0.827 \\
& DenseNet(\cite{huang2017densely}) & 0.569 & 0.929 \\
& ResNet(\cite{he2016deep}) & 0.485 & 0.897 \\
& VGG(\cite{simonyan2014very}) & 0.538 & 0.808 \\
& GoogleNet(\cite{szegedy2015going}) & 0.609 & 0.969 \\
& SqueezeNet(\cite{iandola2016squeezenet}) & 0.404 & 0.698 \\
\midrule
\multirow{6}{*}{\rotatebox[origin=c]{90}{\textbf{VSL}}} 
& Barlowtwins\cite{zbontar2021barlow} & 0.148 & 0.410 \\
& SimCLR\cite{chen2020simple} & 0.130 & 0.379 \\
& byol\cite{grill2020bootstrap} & 0.143 & 0.453 \\
& moco\cite{he2020momentum} & 0.169 & 0.472 \\
& simsiam\cite{chen2021exploring} & 0.188 & 0.535 \\
\midrule
\multirow{24}{*}{\rotatebox[origin=c]{90}{\textbf{Vision Transformers (ViTs)}}} 
& CCT\cite{hassani2021escaping} & 0.570 & 0.981 \\
& CVT\cite{CVT} & 0.577 & 0.930 \\
& ConViT\cite{ConViT} & 0.609 & 0.957 \\
& ConvVT\cite{CVT} & 0.319 & 0.921 \\
& CrossViT\cite{Crossvit} & 0.442 & 0.915 \\
& SwinT\cite{SwinT} & 0.707 & 0.993 \\
& VanillaViT\cite{dosovitskiy2020image} & 0.655 & 0.970 \\
& Visformer\cite{visformer} & 0.398 & 0.856 \\
& ATS\cite{fayyaz2021ats} & 0.540 & 0.973 \\
& CaiT\cite{CaiT} & 0.657 & 0.989 \\
& DeepViT\cite{Deepvit} & 0.546 & 0.988 \\
& Dino\cite{Dino} & 0.049 & 0.437 \\
& Distillation\cite{Distillation} & 0.533 & 0.955 \\
& LeViT\cite{Levit} & 0.624 & 0.970 \\
& NesT\cite{Nest} & 0.660 & 0.985 \\
& PatchMerger\cite{PatchMerger} & 0.578 & 0.975 \\
& PiT\cite{PiT} & 0.555 & 0.979 \\
& RegionViT\cite{Regionvit} & 0.606 & 0.948 \\
& SMIM\cite{SMIM} & 0.171 & 0.646 \\
& T2TViT\cite{T2TViT} & 0.749 & 0.992 \\
& ViT-SD\cite{ViT-SD} & 0.597 & 0.973 \\
\midrule
\multicolumn{1}{c|}{} & Zero-Shot-Image Captioning / \textbf{sLAVA} & \textbf{0.839} & \textbf{0.878} \\  \bottomrule
\multicolumn{1}{c|}{} & Few-Shot-Image Captioning / \textbf{sLAVA} & \textbf{0.987} & \textbf{0.994} \\  \bottomrule
\end{tabular}
\vspace{-2mm}
\end{table}

\vspace{-2mm}
\begin{table}[ht!]
\footnotesize
\centering
\setlength{\tabcolsep}{4pt}
\caption{The table shows the comparison of supervised-Learning Graph Neural Networks(GNNs), self-supervised Graph Contrasting Learning(GCL) Algorithms on the classification task.}
\label{tab:tableclass2}
\vspace{2mm}
\begin{tabular}{cc|c|c}
\hline
\multicolumn{2}{c|}{\textbf{Algorithms}} & \textbf{Top-1} & \textbf{Top-5}  \\ \hline
\multicolumn{1}{c|}{\multirow{4}{*}{\rotatebox[origin=c]{90}{\textbf{GCL}}}} 
& GBT\cite{bielak2021graph} & 0.547 & 0.706 \\
\multicolumn{1}{c|}{} & GRACE\cite{zhu2020deep} & 0.598 & 0.750 \\
\multicolumn{1}{c|}{} & BGRL\cite{thakoor2021bootstrapped} & 0.556 & 0.696 \\
\multicolumn{1}{c|}{} & InfoGraph\cite{sun2019infograph} & 0.526 & 0.702 \\
\hline
\multicolumn{1}{c|}{\multirow{15}{*}{\rotatebox[origin=c]{90}{\textbf{Graph Neural Networks}}}} 
& APPNP\cite{klicpera2018predict} & 0.632 & 0.786 \\
\multicolumn{1}{c|}{} & AGNN\cite{thekumparampil2018attention} & 0.538 & 0.894 \\
\multicolumn{1}{c|}{} & ARMA\cite{bianchi2021graph} & 0.582 & 0.987 \\
\multicolumn{1}{c|}{} & DNA\cite{fey2019just} & 0.622 & 0.916 \\
\multicolumn{1}{c|}{} & GAT\cite{velivckovic2017graph} & 0.491 & 0.985 \\
\multicolumn{1}{c|}{} & GGConv\cite{li2015gated} & 0.563 & 0.992 \\
\multicolumn{1}{c|}{} & GraphConv\cite{morris2019weisfeiler} & 0.658 & 0.996 \\
\multicolumn{1}{c|}{} & GCN2Conv\cite{chen} & 0.732 & 0.998 \\
\multicolumn{1}{c|}{} & ChebConv\cite{defferrard2016convolutional} & 0.504 & 0.951 \\ 
\multicolumn{1}{c|}{} & GraphConv\cite{morris2019weisfeiler} & 0.509 & 0.993 \\
\multicolumn{1}{c|}{} & GraphUNet\cite{gao2019graph} & 0.657 & 0.978 \\
\multicolumn{1}{c|}{} & MPNN\cite{gilmer2017neural} & 0.603 & 0.999 \\
\multicolumn{1}{c|}{} & RGGConv\cite{bresson2017residual} & 0.618 & 0.961 \\
\multicolumn{1}{c|}{} & SuperGAT\cite{kim2022find} & 0.598 & 0.985 \\
\multicolumn{1}{c|}{} & TAGConv\cite{du2017topology} & 0.598 & 0.999 \\
\hline
\multicolumn{1}{c|}{} & Zero-Shot-Image Captioning / \textbf{sLAVA} & \textbf{0.839} & \textbf{0.878} \\  \bottomrule
\multicolumn{1}{c|}{} & Few-Shot-Image Captioning / \textbf{sLAVA} & \textbf{0.987} & \textbf{0.994} \\  \bottomrule
\end{tabular}
\vspace{-2mm}
\end{table}

\vspace{-5mm}
\begin{table}[htbp]
\footnotesize
\centering
\resizebox{0.50\textwidth}{!}{%
\begin{tabular}{@{}c|ccc|c@{}}
\toprule
\multirow{2}{*}{\textbf{Category}} & \multicolumn{3}{c|}{\textbf{Multi-class metrics}} \\ \cmidrule(lr){2-4}
                                  & \multicolumn{1}{c|}{\textbf{Precision}} & \multicolumn{1}{c|}{\textbf{Recall}} & \textbf{F1 Score} \\ \midrule
Biological                        & \multicolumn{1}{c|}{0.971$\pm$0.009}    & \multicolumn{1}{c|}{0.993$\pm$0.007} & 0.983$\pm$0.013 \\
Tips                              & \multicolumn{1}{c|}{0.954$\pm$0.005}    & \multicolumn{1}{c|}{0.967$\pm$0.008} & 0.964$\pm$0.011 \\
Fibres                            & \multicolumn{1}{c|}{0.995$\pm$0.007}    & \multicolumn{1}{c|}{1.000$\pm$0.000} & 1.000$\pm$0.000 \\
Porous Sponge                     & \multicolumn{1}{c|}{0.971$\pm$0.014}    & \multicolumn{1}{c|}{0.981$\pm$0.013} & 0.965$\pm$0.010 \\
Films Coated Surface              & \multicolumn{1}{c|}{0.979$\pm$0.005}    & \multicolumn{1}{c|}{0.979$\pm$0.009} & 0.988$\pm$0.008 \\
Patterned Surface                 & \multicolumn{1}{c|}{0.988$\pm$0.016}    & \multicolumn{1}{c|}{0.983$\pm$0.006} & 0.982$\pm$0.014 \\
Nanowires                         & \multicolumn{1}{c|}{0.979$\pm$0.012}    & \multicolumn{1}{c|}{0.989$\pm$0.007} & 0.995$\pm$0.011 \\
Particles                         & \multicolumn{1}{c|}{0.982$\pm$0.006}    & \multicolumn{1}{c|}{0.978$\pm$0.011} & 0.968$\pm$0.023 \\
MEMS Devices                      & \multicolumn{1}{c|}{0.983$\pm$0.011}    & \multicolumn{1}{c|}{0.970$\pm$0.008} & 0.966$\pm$0.009 \\
Powder                            & \multicolumn{1}{c|}{0.985$\pm$0.014}    & \multicolumn{1}{c|}{0.971$\pm$0.009} & 0.955$\pm$0.011 \\
\bottomrule
\end{tabular}}
\vspace{-1mm}
\caption{The table shows the effectiveness of our proposed framework, compared to existing methods, in terms of precision, recall, and F1-score for accurately classifying nanomaterials of different categories.}
\label{tab:mccategory}
\vspace{-4mm}
\end{table}

\vspace{-2mm}
\begin{table*}[!htb]
    \caption{The table showcases sample electron microscope images alongside their corresponding ground truth captions and captions generated by the small-scale vision-language model on a VQA task for characterizing nanostructure size, distribution, and aggregation. To evaluate the quality of these machine-generated descriptions, BLEU-2, ROUGE-L, and METEOR metrics are included, assessing their similarity to the accurate labels.}
    \vspace{2mm}
      \centering
        \begin{tabular}{|>{\centering\arraybackslash}m{2cm}|m{5.85cm}|m{6.25cm}|m{1.75cm}|}
        \hline 
         Image & Ground Truth & Answers & BLEU-2/ \ ROGUE-L/ \ METEOR \\ \hline
        \includegraphics[width=2cm,height=2cm,keepaspectratio]{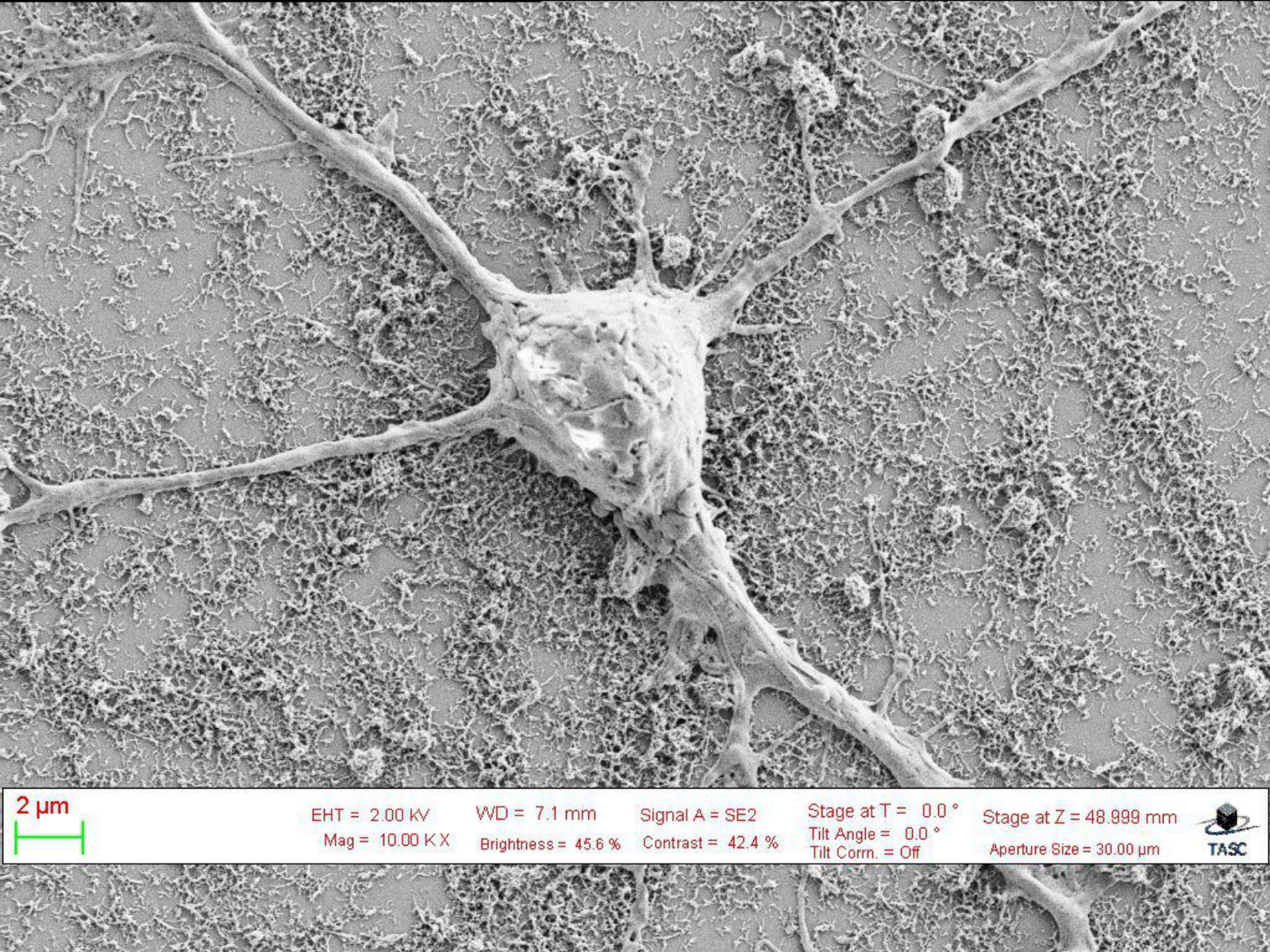} & The SEM image shows nanostructures that vary in size, clustered predominantly around a central node, with a non-uniform, network-like distribution and noticeable aggregation at interconnection points. & The SEM image displays nanostructures that vary in size, predominantly clustered around a central node, with a non-uniform, network-like pattern and noticeable aggregation at the interconnection points. & \begin{tabular}[c]{@{}c@{}}0.762\\ 0.877\\ 0.947\end{tabular} \\ \hline
        \includegraphics[width=2cm,height=2cm,keepaspectratio]{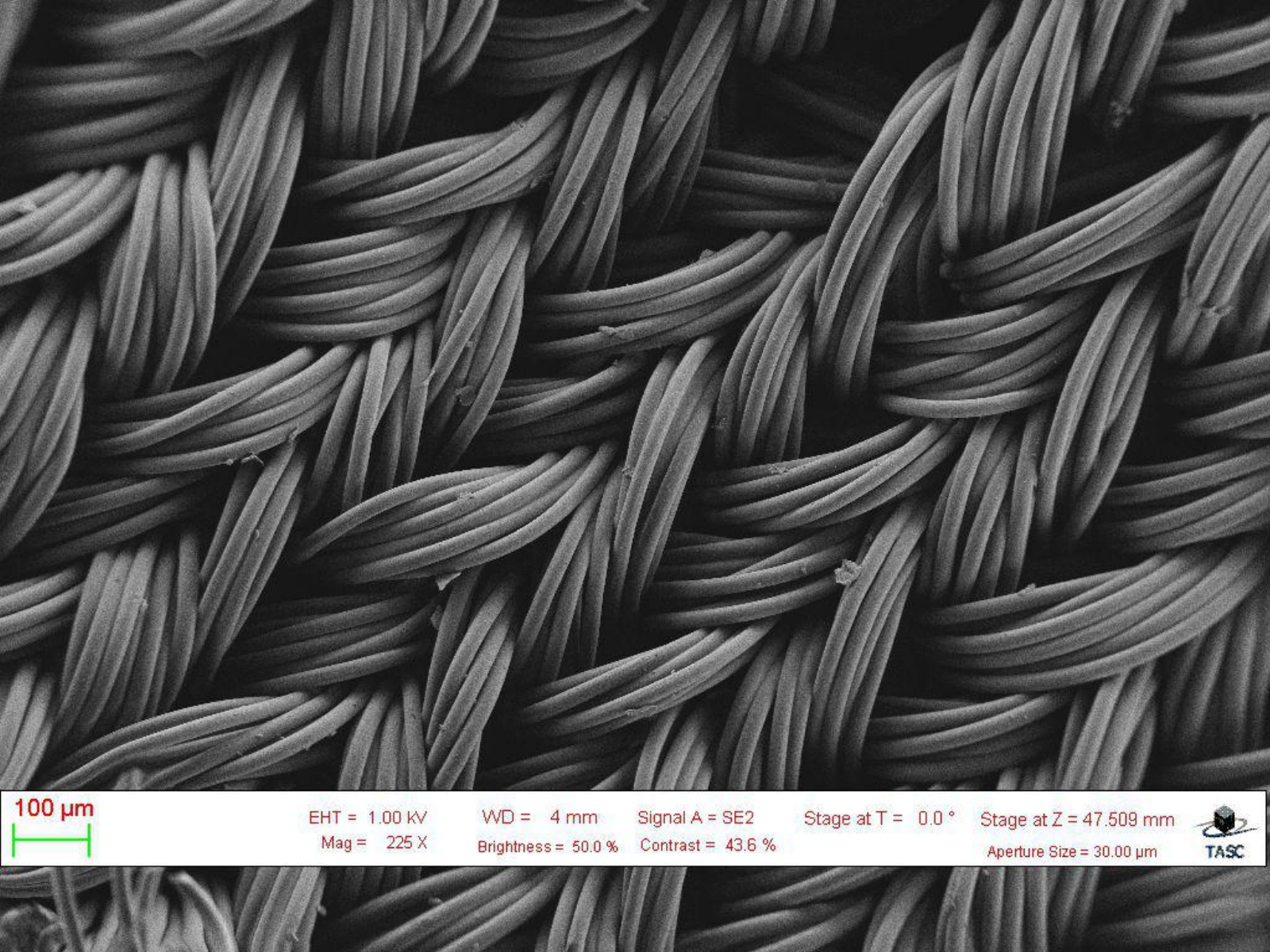} & The SEM image shows tightly aligned fibrous nanostructures with micrometer-scale width, forming a uniform, mat-like arrangement without random aggregation. & The SEM image illustrates tightly aligned fibrous nanostructures with micrometer-scale width, forming a uniform, mat-like configuration without random aggregation. & \begin{tabular}[c]{@{}c@{}}0.834\\ 0.905\\ 0.892\end{tabular}  \\ \hline
        \includegraphics[width=2cm,height=2cm,keepaspectratio]{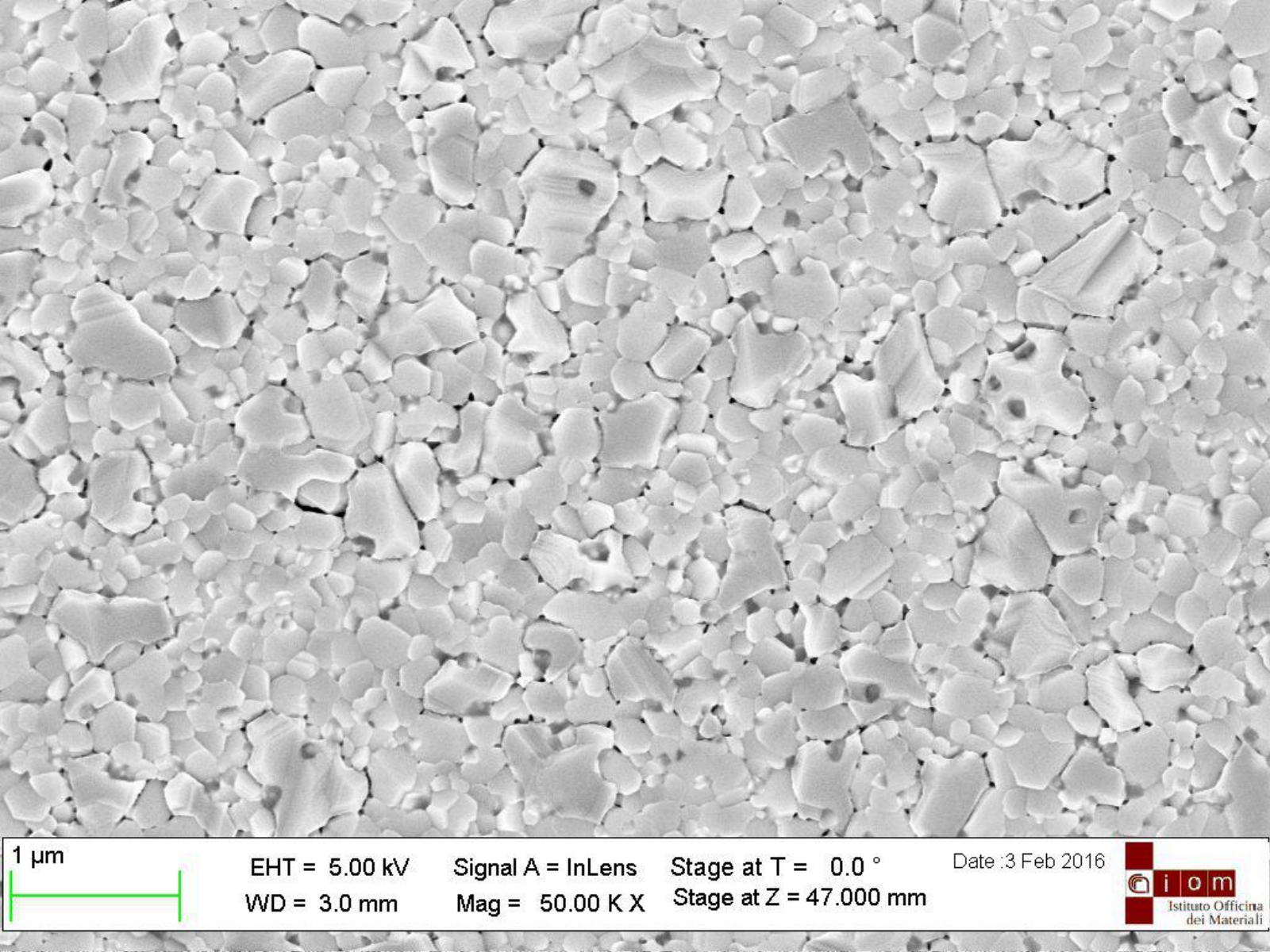} & The SEM image reveals hexagonal nanostructures tightly packed in a uniform pattern, with individual sizes in the sub-micrometer range and minimal interstitial spacing, indicating a structured assembly. & The SEM image reveals hexagonal nanostructures compactly arranged in a uniform layout, with individual sizes in the sub-micrometer range and limited space between them, denoting a structured formation. & \begin{tabular}[c]{@{}c@{}}0.614\\ 0.702\\ 0.698\end{tabular}  \\ \hline
        \includegraphics[width=2cm,height=2cm,keepaspectratio]{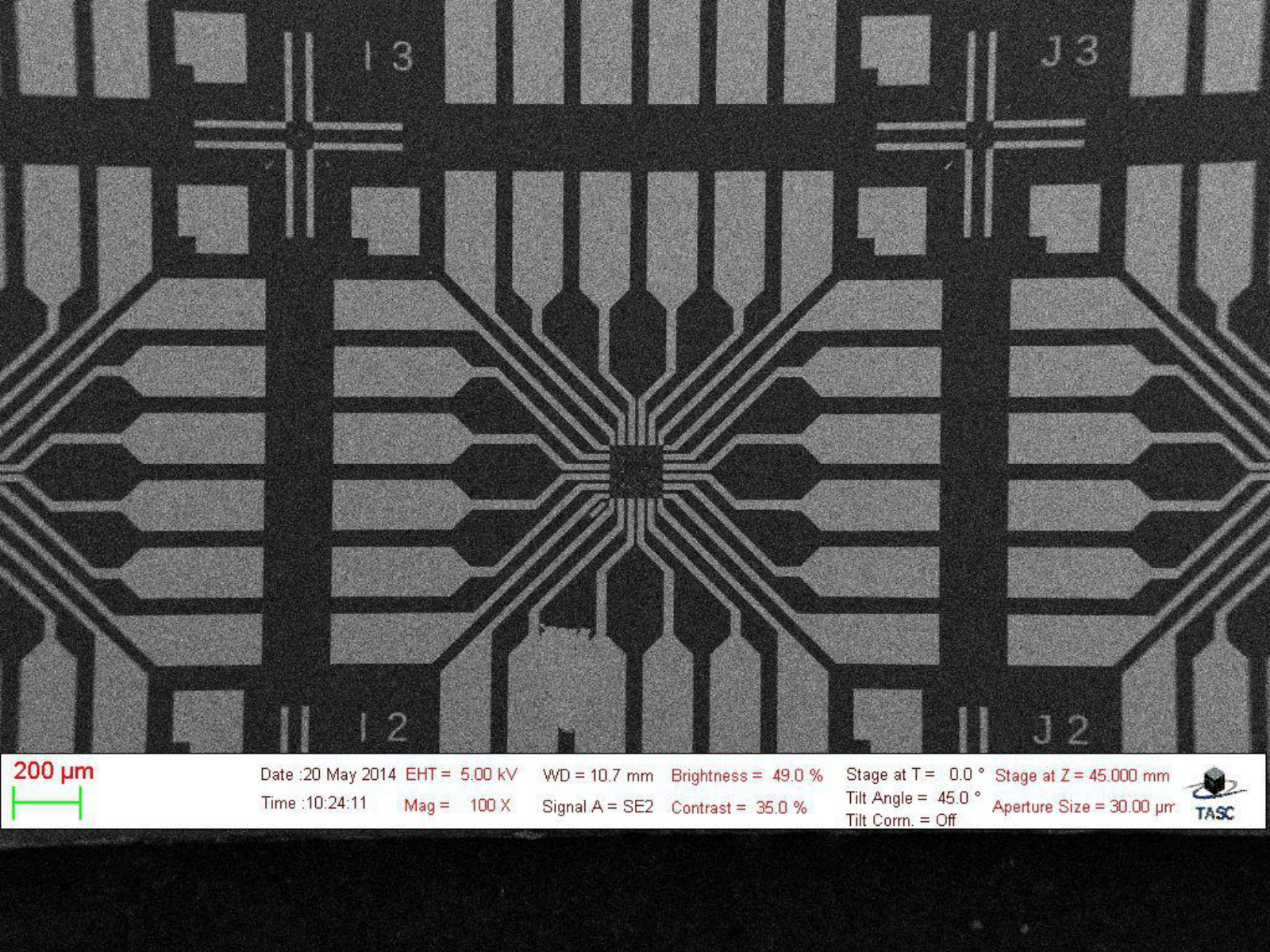} & The image shows microstructures with micrometer-scale, geometric precision, indicative of controlled lithography with no random aggregation, arranged in a regular, circuit-like pattern. & The image displays microstructures with micrometer-scale, geometrically precise, indicating controlled lithography with no random clusters, arranged in a regular, circuit-style pattern. & \begin{tabular}[c]{@{}c@{}}0.570\\ 0.723\\ 0.813\end{tabular}  \\ \hline
        \includegraphics[width=2cm,height=2cm,keepaspectratio]{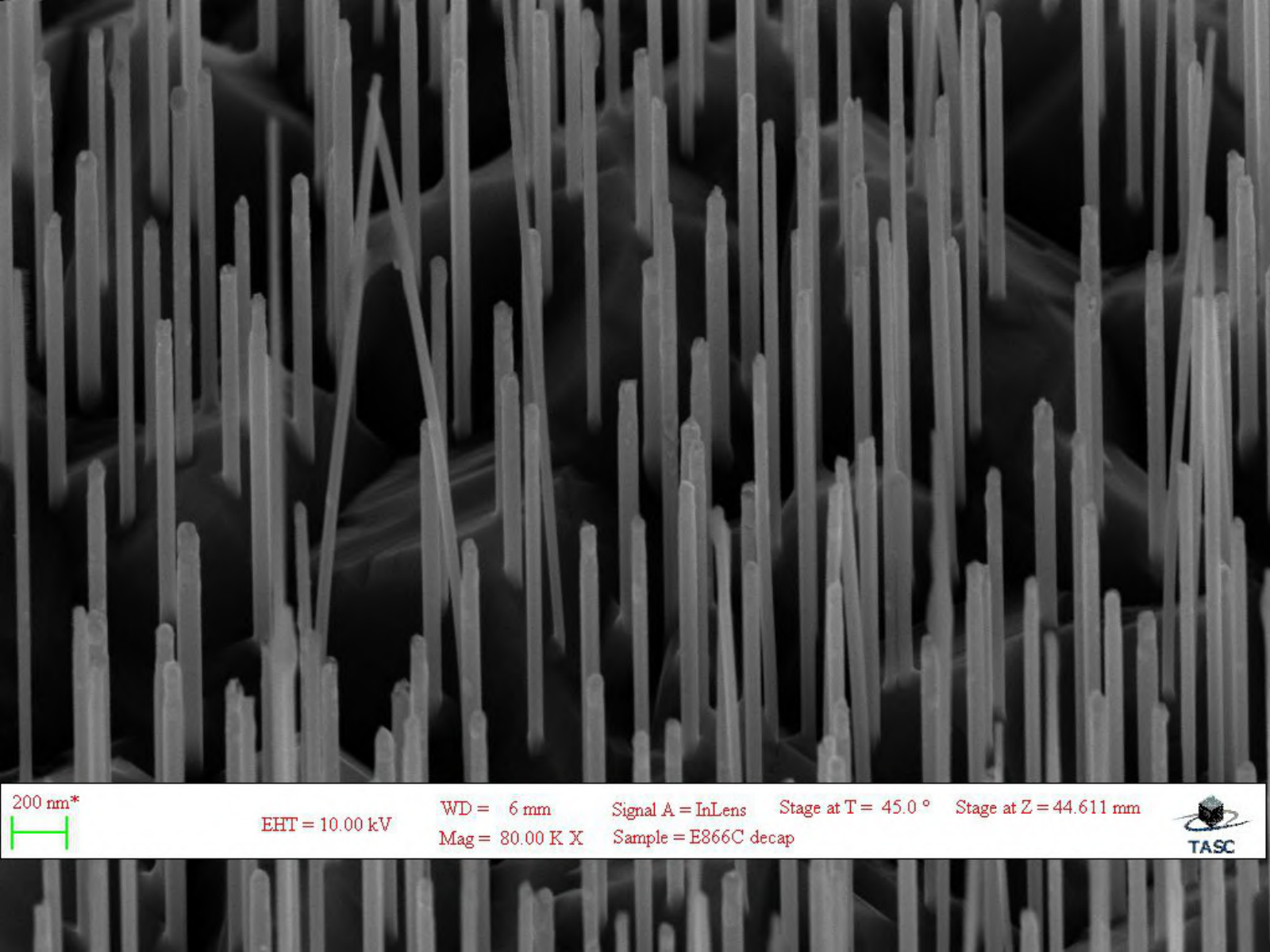} & The image features uniformly spaced, vertically aligned nanowires with diameters in the hundreds of nanometers, showing no significant aggregation. & The image depicts uniformly spaced, vertically oriented nanowires with diameters spanning hundreds of nanometers, without substantial aggregation. & \begin{tabular}[c]{@{}c@{}}0.494\\ 0.667\\ 0.638\end{tabular}  \\ \hline
        \includegraphics[width=2cm,height=2cm,keepaspectratio]{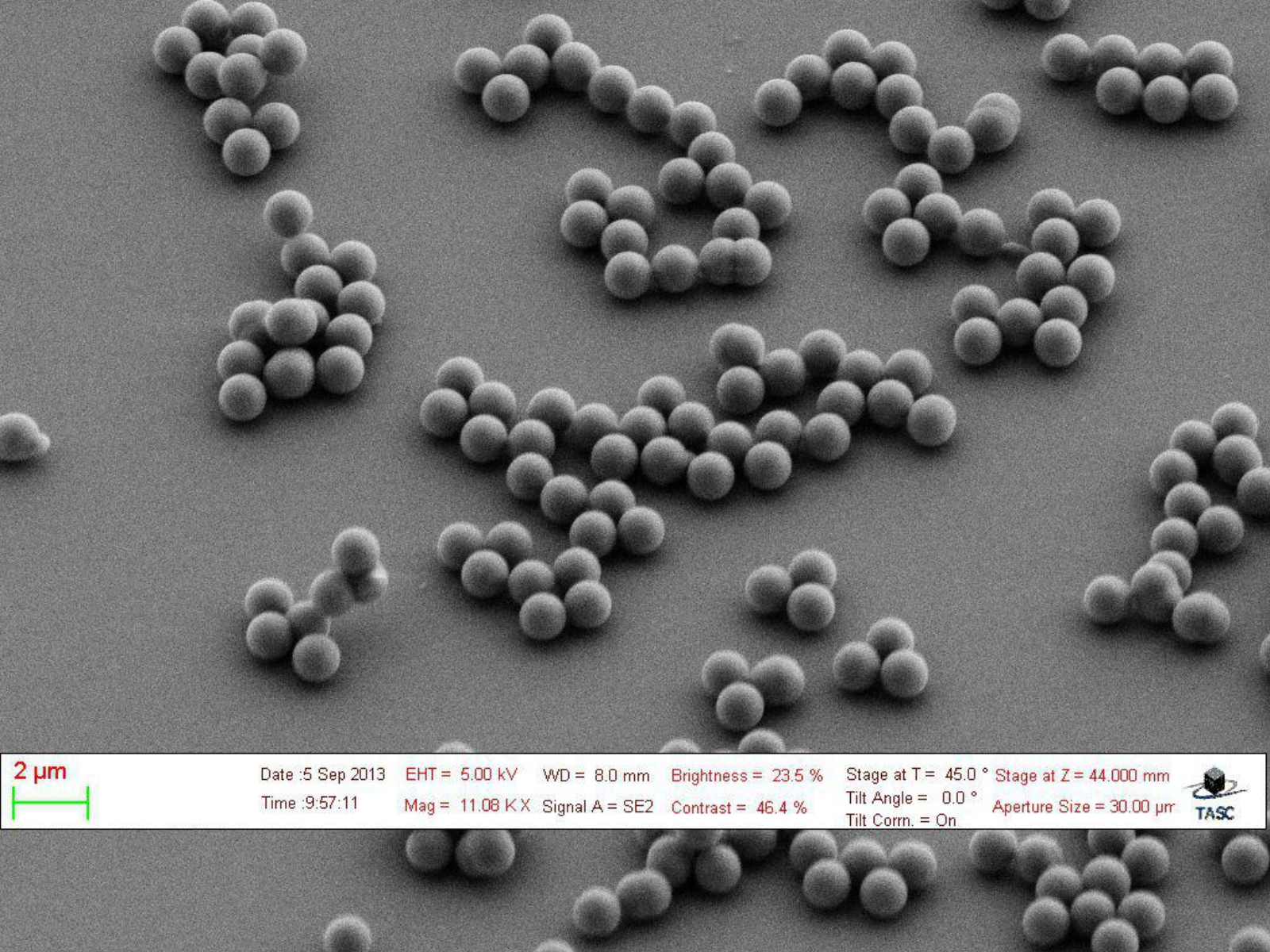} & The SEM image displays spherical nanostructures grouped in clusters, with sizes likely in the tens to hundreds of nanometers range, scattered randomly across the surface. & The SEM image shows spherical nanostructures formed in clusters, with their sizes likely varying from tens to hundreds of nanometers, randomly spread across the surface. & \begin{tabular}[c]{@{}c@{}}0.574\\ 0.760\\ 0.764\end{tabular} \\ \hline
        \includegraphics[width=2cm,height=2cm,keepaspectratio]{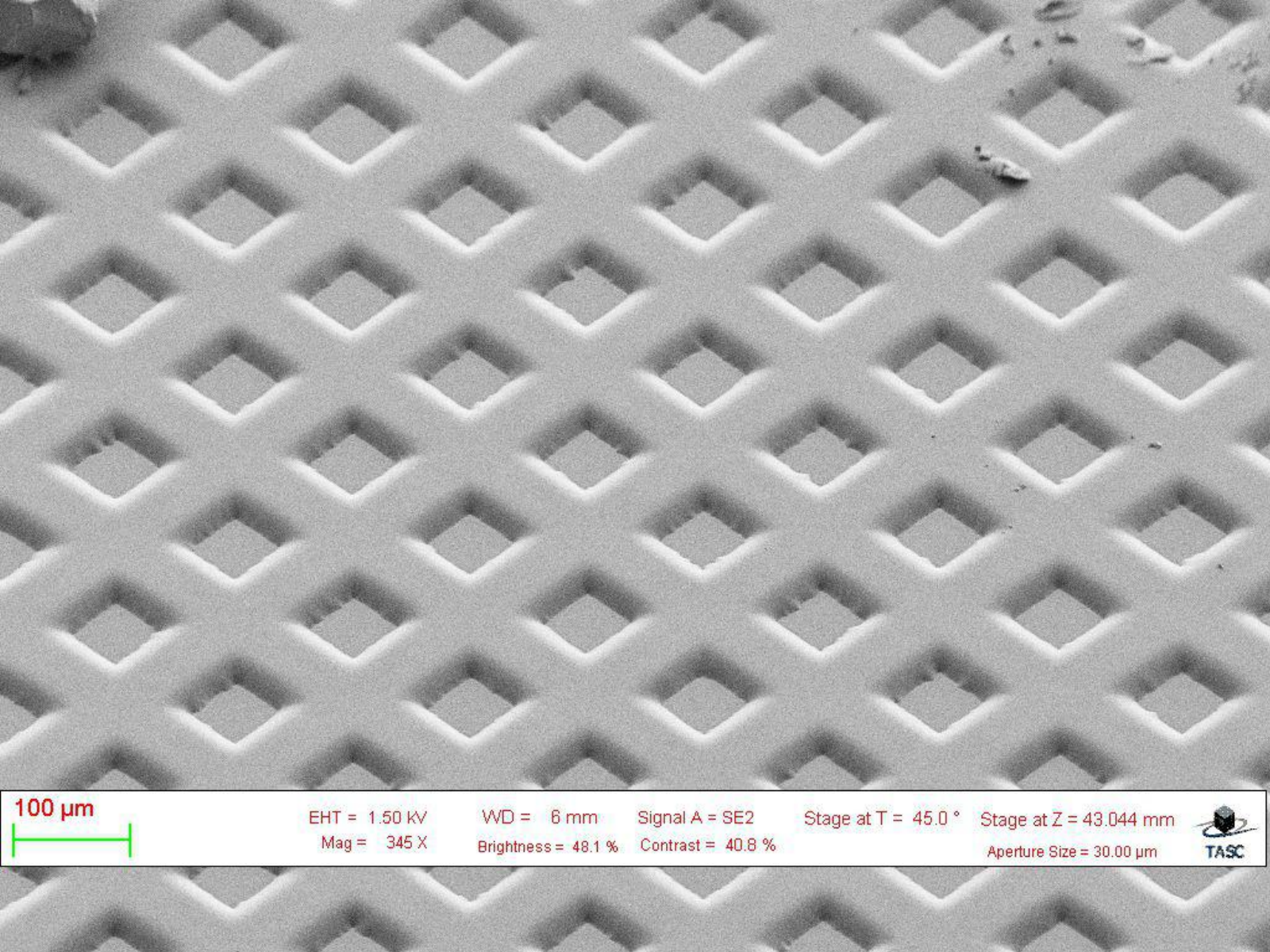} & The image shows a diamond-shaped nano-lattice with uniform units several micrometers in size, indicating precision fabrication without random aggregation. & The image presents a diamond-configured nano-lattice with uniform structures a few micrometers in size, demonstrating precision in fabrication without evident aggregation. & \begin{tabular}[c]{@{}c@{}}0.431\\ 0.682\\ 0.632\end{tabular}  \\ \hline
        \includegraphics[width=2cm,height=2cm,keepaspectratio]{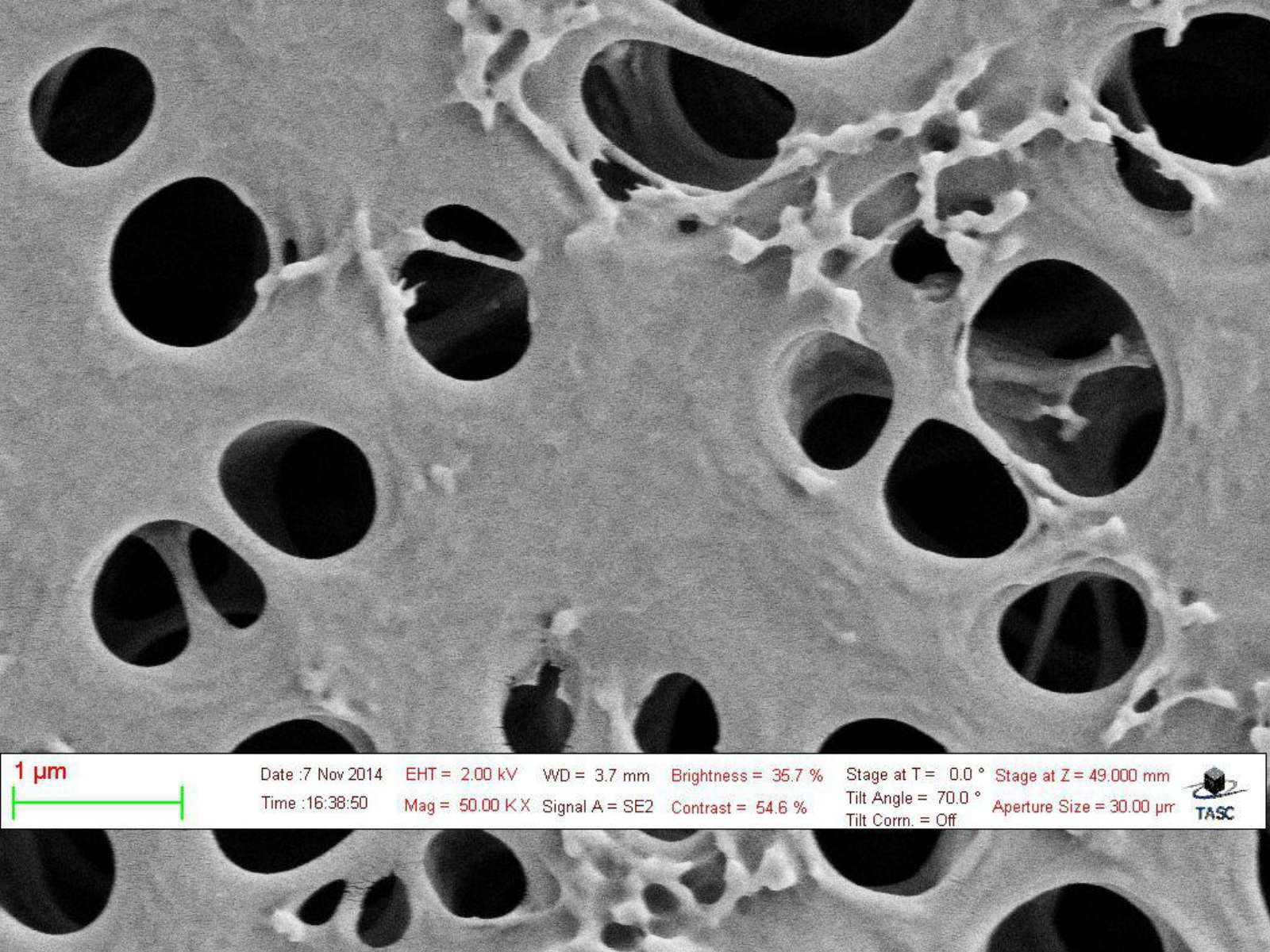} & The image shows randomly distributed, irregularly shaped voids up to a few micrometers in size within a solid matrix, with no aggregation. & The image depicts randomly arranged, unevenly shaped voids, each up to a few micrometers in size, within a solid framework, without any aggregation. & \begin{tabular}[c]{@{}c@{}}0.470\\ 0.711\\ 0.666\end{tabular}  \\ \hline
        \includegraphics[width=2cm,height=2cm,keepaspectratio]{Images/QA/SEM/Powder.pdf} & The image presents densely packed, aggregated nanostructures with a size range likely within the sub-micrometer to micrometer scale, as inferred from the 2 $\mu$m scale bar. & The image demonstrates densely aggregated nanostructures with dimensions between the sub-micrometer and micrometer scale, as inferred from the 2$\mu$m scale bar. & \begin{tabular}[c]{@{}c@{}}0.609\\ 0.711\\ 0.708\end{tabular}  \\ \hline
        \includegraphics[width=2cm,height=2cm,keepaspectratio]{Images/QA/SEM/Tips.pdf} & The image captures a single nanostructure with a tapered end, measuring less than 200 nm at its widest point, isolated without any visible aggregation or neighboring structures. & The micrograph shows an individual nanostructure with a tapering tip, spanning under 200 nm at its broadest region, separated without any discernible clustering or adjacent formations. & \begin{tabular}[c]{@{}c@{}}0.307\\ 0.711\\ 0.430\end{tabular} \\ \hline
        \end{tabular}    
        \label{VQA2}
        
\end{table*}

\onecolumn

\pagestyle{empty} 

\begin{adjustwidth}{-0.15cm}{-0.15cm} 
\begin{tcolorbox}[colback=white!5!white,colframe=black!75!black]% [width=10cm]
\vspace{-5mm}
\begin{tabularx}{1.0\textwidth}{bss}
\caption{The table shows question-answer pairs about a microscopic image of patterned surface nanomaterial, created with GPT-4 Turbo with vision. In summary, GPT-4 helps generate detailed instruction-tuning dataset about these nanomaterials images for training smaller multimodal models(SMMs), to serve as a domain-specific vision-language assistant for enterprise adoption.}\label{tab:tab0}\\ [-0.75em]
\toprule 
 \\ [-0.75em]
\begin{overpic}[width=0.15\columnwidth]{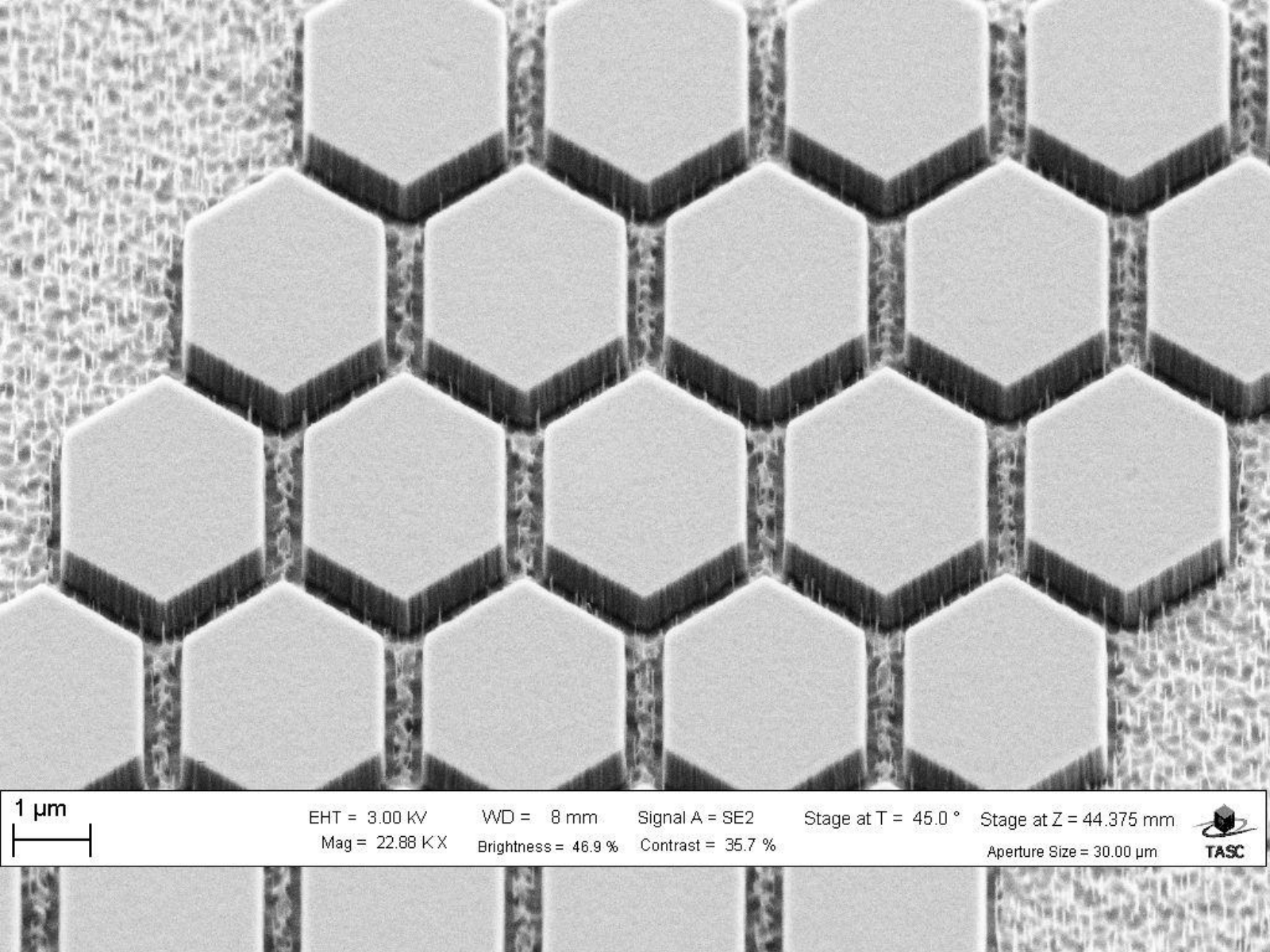}
  \put(425,40){\makebox[0pt][c]{%
    \parbox{0.65\columnwidth}{%
      \centering
      \textbf{Examine the provided input image and answer the following questions about the patterned surface nanomaterial it depicts.} 
    }%
  }}  
\end{overpic}  \\ [-0.25em]

\midrule
\textbf{Query prompt}  & \textbf{GPT-4 Turbo with vision output} \\
\midrule
\endhead
\textcolor{red}{**Basics**} &  The image is a scanning electron microscope (SEM) view of a patterned nanomaterial with hexagonal structures. The scale bar indicates that 1 micrometer ($\mu$m) is the unit of measurement, which means each unit on the image corresponds to 1 $\mu$m in the real world.
\\

\midrule
\textcolor{black}{**Morphology and Structure**} &  The nanomaterials in the image are hexagonal and uniformly sized, arranged in a honeycomb pattern without visible layers or phases within the individual structures. Their consistent shape suggests a precise and controlled fabrication process.
\\

\midrule
\textcolor{black}{**Size and Distribution**} &  The nanostructures appear to be roughly 300 to 500 nanometers in size, judging from the 1-micrometer scale bar. They are evenly spaced in a hexagonal pattern, without any sign of clustering or random placement. There's also no visible aggregation or bundling; each nanostructure is distinct and separate from its neighbors.
\\

\midrule
\textcolor{black}{**Surface Characteristics**} & The nanomaterials have smooth, flat surfaces with uniform hexagonal shapes and no visible imperfections, suggesting high purity and structural integrity in their creation.
\\

\midrule
\textcolor{black}{**Composition and Elements**} & The electron microscope image displays uniform brightness and contrast across the nanostructures, suggesting a consistent material composition without visible evidence of different components. There are no labels or markers indicating specific elements or compounds; such information would require additional analytical methods not provided in the image.
\\
 
\midrule
\textcolor{black}{**Interactions and Boundaries**} & The nanostructures are fully separate with clear, well-defined boundaries between them, suggesting precise fabrication.
\\
 
\midrule
\textcolor{black}{**External Environment**} & The nanomaterials are set against a textured substrate with no visible signs of interaction like dissolution or blending. The image only shows the nanostructures and their immediate environment, with no other distinct objects or structures present.
\\
 
\midrule
\textcolor{black}{**Image Technique and Modifications**} & The image is a raw SEM micrograph with potential minor adjustments to brightness and contrast, showcasing the three-dimensional detail of the nanostructures.
\\
 
\midrule
\textcolor{black}{**Functional Features**} & The image shows uniform nanostructures without identifiable functional elements or regions, providing a static structural representation rather than revealing behavioral or dynamic information.
\\

\midrule
\textcolor{black}{**Context and Application**} & The uniform hexagonal shape and close packing of the nanomaterials in the image suggest potential applications in photonics, electronics, or materials science, and the realistic details indicate it depicts actual experimental samples.
\\ [-0.75em]

\end{tabularx}
\end{tcolorbox}
\end{adjustwidth}

\begin{adjustwidth}{-0.15cm}{-0.15cm} 
\begin{tcolorbox}[colback=white!5!white,colframe=black!75!black]% [width=10cm]
\vspace{-5mm}
\begin{tabularx}{1.0\textwidth}{bss}
\caption{The table shows question-answer pairs about a microscopic image of biological nanomaterial, created with GPT-4 Turbo with vision. In summary, GPT-4 helps generate detailed instruction-tuning dataset about these nanomaterials images for training smaller multimodal models(SMMs), to serve as a domain-specific vision-language assistant for enterprise adoption.}\label{tab:tab1}\\ [-0.75em]
\toprule 
 \\ [-0.95em]
\begin{overpic}[width=0.155\columnwidth]{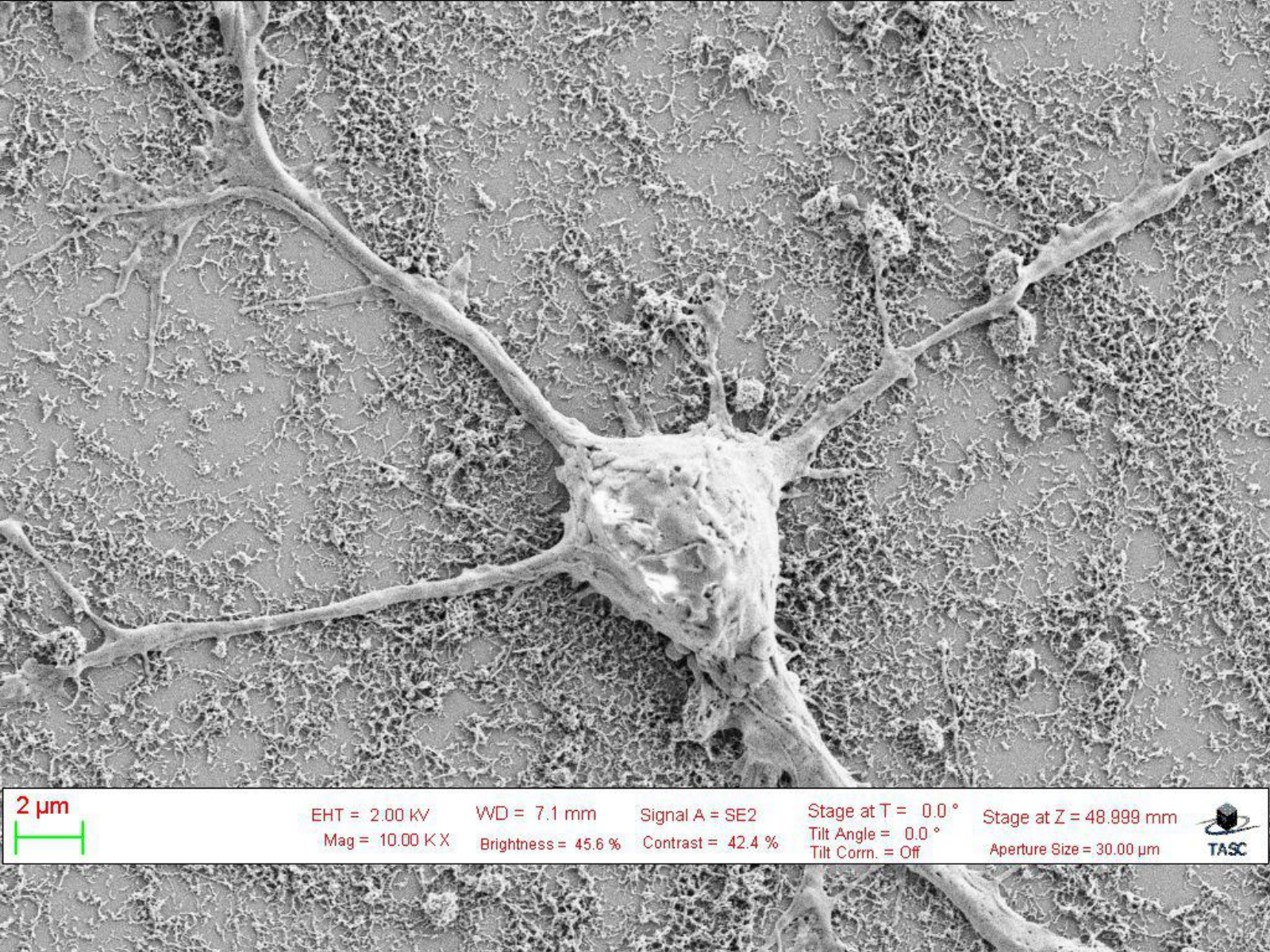}
  \put(425,35){\makebox[0pt][c]{%
    \parbox{0.65\columnwidth}{%
      \centering
      \textbf{Examine the provided input image and answer the following questions about the biological nanomaterial it depicts.} 
    }%
  }}  
\end{overpic}  \\ [-0.25em]

\midrule
\textbf{Query prompt}  & \textbf{GPT-4 Turbo with vision output} \\
\midrule
\endhead
\textcolor{black}{**Basics**} & The image is a Scanning Electron Microscope (SEM) image of a neuron with dendrites. The scale bar represents 2 micrometers, indicating the high magnification level of the image.
\\

\midrule
\textcolor{black}{**Morphology and Structure**} & The depicted neuron features a roughly spherical soma with a nucleus and organelles. Branch-like dendrites extend from the soma, varying in thickness and length, crucial for receiving signals and neuronal networking. Unlike inorganic materials, the neuron has an irregular surface and shows variability in size and shape, typical of biological structures.
\\

\midrule
\textcolor{black}{**Size and Distribution**} & The image shows a neuron's cell body and finer branching dendrites. The scale bar indicates a cell body diameter of several micrometers. The dendrites vary in width, likely from hundreds of nanometers to a micrometer. The non-uniform dendrites extend outward in a complex, branching pattern characteristic of neuronal networking. Unlike aggregated inorganic nanostructures, the dendrites appear individually extended with clear spaces between to establish neuronal synapses.
\\

\midrule
\textcolor{black}{**Surface Characteristics**} & The neuron's surface is rough and irregular, typical for biological structures under high magnification. There are no uniform textures, and while there are no imperfections like in synthetic materials, the surface complexity reflects the presence of biological molecules.
\\

\midrule
\textcolor{black}{**Composition and Elements**} & The SEM image is grayscale, showing variations in brightness and contrast that may suggest different materials, but specific compositional details cannot be determined. There are no labels indicating the presence of specific elements or compounds in the image.
\\
 
\midrule
\textcolor{black}{**Interactions and Boundaries**} &  The dendrites in the image are separate and not touching or fused, with each dendrite clearly distinguishable from others. The boundaries between these structures are well-defined, showing no blending or undefined borders.
\\
 
\midrule
\textcolor{black}{**External Environment**} & In the SEM image, there are no visible interactions between the neuron and an external environment like solvents or polymers. The background is uniform, with no discernible structures or objects other than the neuron and its dendrites.
\\
 
\midrule
\textcolor{black}{**Image Technique and Modifications**} &  The image is captured using Scanning Electron Microscopy (SEM), indicated by its high resolution and grayscale appearance. There is no evidence of post-processing techniques like false coloring or 3D rendering; it appears to be a standard SEM micrograph.
\\
 
\midrule
\textcolor{black}{**Functional Features**} & The SEM image of the neuron does not reveal specific functional elements like active sites. It's a static representation, as SEM captures a frozen moment and cannot depict dynamic processes within the neuron.
\\

\midrule
\textcolor{black}{**Context and Application**} & The image depicts actual neurons studied to understand the nervous system and brain function. The detailed SEM imaging indicates these are experimentally observed, not simulated, neuron structures.
\\ [-0.75em]

\end{tabularx}
\end{tcolorbox}
\end{adjustwidth}

\begin{adjustwidth}{-0.15cm}{-0.15cm}
\begin{tcolorbox}[colback=white!5!white,colframe=black!75!black]% [width=10cm]
\vspace{-5mm}
\begin{tabularx}{1.0\textwidth}{bss}
\caption{The table shows question-answer pairs about a microscopic image of fibers nanomaterial, created with GPT-4 Turbo with vision. In summary, GPT-4 helps generate detailed instruction-tuning dataset about these nanomaterials images for training smaller multimodal models(SMMs), to serve as a domain-specific vision-language assistant for enterprise adoption.}\label{tab:tab2}\\ [-0.75em]
\toprule 
 \\ [-0.75em]
\begin{overpic}[width=0.135\columnwidth]{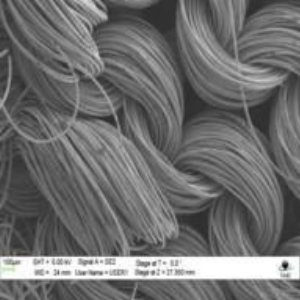}
  \put(475,50){\makebox[0pt][c]{%
    \parbox{0.65\columnwidth}{%
      \centering
      \textbf{Examine the provided input image and answer the following questions about the fibres nanomaterial it depicts.} 
    }%
  }}  
\end{overpic}  \\ [-0.25em]

\midrule
\textbf{Query prompt}  & \textbf{GPT-4 Turbo with vision output} \\
\midrule
\endhead
\textcolor{black}{**Basics**} & The image depicts nanofibers, which are a type of nanomaterial. The scale or resolution of the image is not clear, so I cannot determine the real-world length each unit of measurement corresponds to.
\\

\midrule
\textcolor{black}{**Morphology and Structure**} &  The image of the nanomaterials exhibits a fibrous morphology with a consistently sized and shaped appearance, indicating a uniform manufacturing process. They are arranged in bundles and aggregations of parallel fibers, resembling twisted yarn or rope. There are no clearly visible layers, phases, or distinct domains within the individual nanofibers from the image alone. The fibers appear to have a relatively consistent diameter and form throughout the visible area, suggesting a homogeneous material without significant variability in size or shape.
\\

\midrule
\textcolor{black}{**Size and Distribution**} & The size of the individual nanostructures cannot be estimated without a clear scale reference. The nanofibers are clustered, resembling bundles, indicating some degree of aggregation. They do not appear to be evenly spaced but rather randomly oriented and aggregated.
\\

\midrule
\textcolor{black}{**Surface Characteristics**} & The fibers appear to have a relatively smooth surface, without distinct textures such as roughness or granular patterns. From the image provided, there are no noticeable imperfections, defects, or pores visible on the surfaces of the nanomaterials. However, the resolution and clarity of the image limit the ability to make detailed observations regarding surface characteristics.
\\

\midrule
\textcolor{black}{**Composition and Elements**} & The grayscale image lacks visual cues to determine compositional variations. No brightness, contrast, labels, or markers indicate distinct components or elemental compounds. Additional spectroscopic data and analysis would be required beyond what is provided.
\\
 
\midrule
\textcolor{black}{**Interactions and Boundaries**} & The nanostructures appear to be touching and fused together, suggesting strong interfiber interactions. The boundaries between individual nanostructures are not clearly distinguishable, indicating they are closely packed or bundled. They blend together without defined borders, which is typical for aggregated nanofibers.
\\
 
\midrule
\textcolor{black}{**External Environment**} & There are no visible signs of interaction with the environment or matrix, and no other structures or objects besides the nanomaterials are evident.
\\
 
\midrule
\textcolor{black}{**Image Technique and Modifications**} & The high-resolution, depth of field, and surface detail indicate this is likely an unprocessed grayscale SEM image, without false coloring or 3D rendering. The characteristics are typical of scanning electron microscopy.
\\
 
\midrule
\textcolor{black}{**Functional Features**} & The image is a static representation of the physical structure of the nanomaterials, without clearly identifiable active sites or visible dynamic processes.
\\

\midrule
\textcolor{black}{**Context and Application**} & The image alone does not provide enough information to determine the intended application of the nanomaterials. Additional context is required to infer whether these are experimental or simulated samples.
\\ [-0.75em]

\end{tabularx}
\end{tcolorbox}
\end{adjustwidth}

\begin{adjustwidth}{-0.15cm}{-0.15cm}
\begin{tcolorbox}[colback=white!5!white,colframe=black!75!black]% [width=10cm]
\vspace{-5mm}
\begin{tabularx}{1.0\textwidth}{bss}
\caption{The table shows question-answer pairs about a microscopic image of films and coated surface nanomaterials, created with GPT-4 Turbo with vision. In summary, GPT-4 helps generate detailed instruction-tuning dataset about these nanomaterials images for training smaller multimodal models(SMMs), to serve as a domain-specific vision-language assistant for enterprise adoption.}\label{tab:tab3}\\ [-0.75em]
\toprule 
 \\ [-0.75em]
\begin{overpic}[width=0.15\columnwidth]{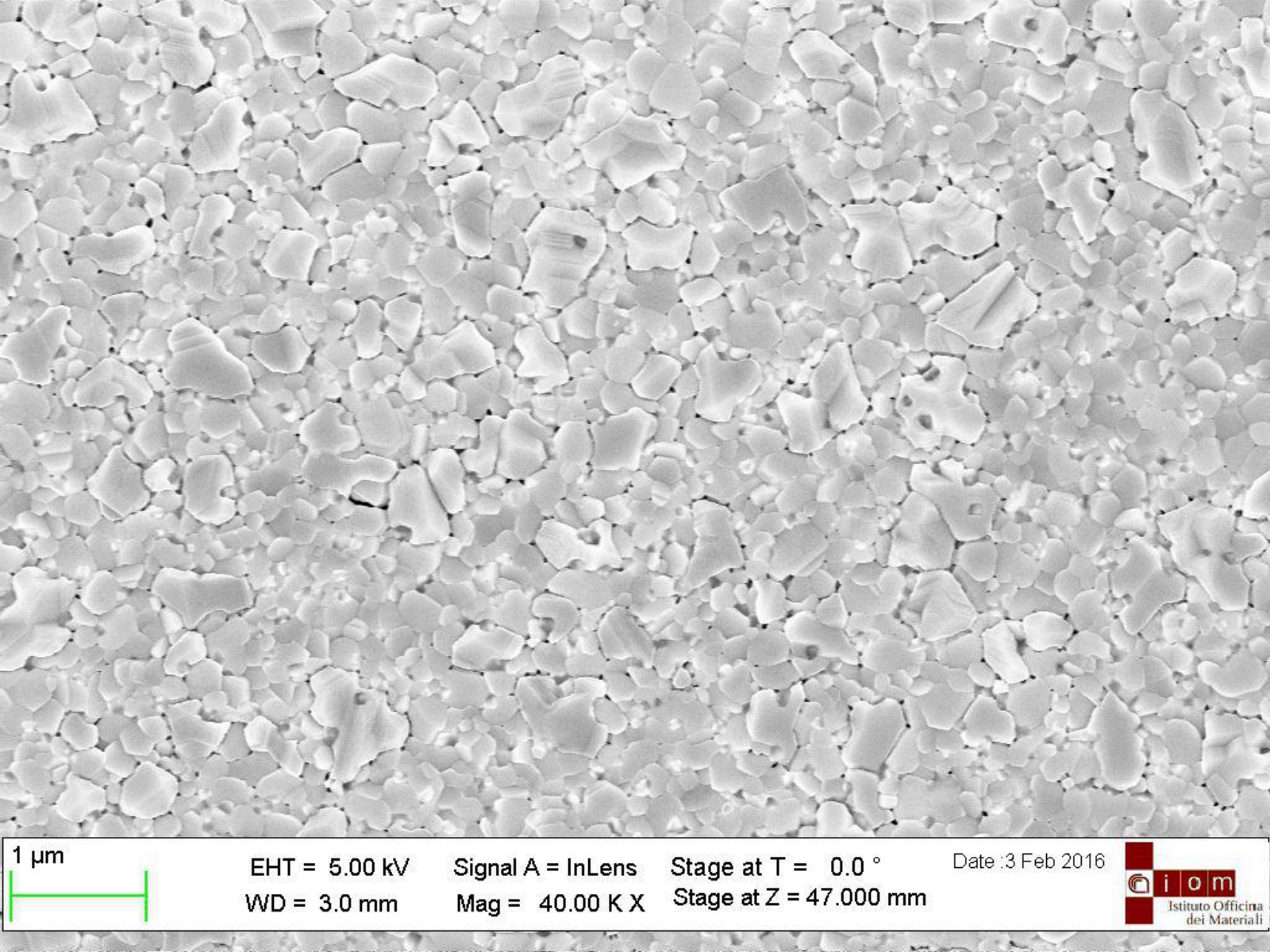}
  \put(450,40){\makebox[0pt][c]{%
    \parbox{0.65\columnwidth}{%
      \centering
      \textbf{Examine the provided input image and answer the following questions about the films and coated surface nanomaterial it depicts.} 
    }%
  }}  
\end{overpic}  \\ [-0.25em]

\midrule
\textbf{Query prompt}  & \textbf{GPT-4 Turbo with vision output} \\
\midrule
\endhead
\textcolor{black}{**Basics**} &  The image is a scanning electron microscope (SEM) image of a nanomaterial with polygonal shaped particles. The scale bar indicates that one unit of measurement in the image corresponds to one micrometer (1 µm). The magnification is 40,000 times (40.00 KX). The specific type of nanomaterial cannot be determined from the image alone without additional information.
\\

\midrule
\textcolor{black}{**Morphology and Structure**} & The nanomaterials in the image exhibit an overall irregular polygonal shape, which is characteristic of crystalline structures that have multiple facets. There don't appear to be visible layers, phases, or distinct domains within the individual particles; they seem relatively homogenous in texture and contrast. There is some variability in size and shape, but the particles generally maintain a consistent morphology throughout the image. This uniformity suggests a controlled synthesis process that promotes a specific crystal growth pattern.
\\

\midrule
\textcolor{black}{**Size and Distribution**} & The nanostructures are sub-micrometer, uniform in size and morphology, evenly spaced and randomly placed without aggregation. They show consistent features across the image.
\\

\midrule
\textcolor{black}{**Surface Characteristics**} & The surfaces of the nanomaterials appear predominantly smooth with some minor texture variations, which are typical for materials at this scale. There are no obvious imperfections such as defects, pores, or impurities that can be discerned from this image. The edges of the nanostructures are well-defined, which often indicates a high level of purity and structural integrity in the synthesis process.
\\

\midrule
\textcolor{black}{**Composition and Elements**} & The SEM image does not show variations in brightness or contrast that would indicate different components, suggesting a uniform composition. There are no labels or markers indicating specific elements or compounds.
\\
 
\midrule
\textcolor{black}{**Interactions and Boundaries**} & The nanostructures appear to be separate and not fused together, with clear boundaries between each particle. There is no blending of borders, indicating distinct and individual structures.
\\
 
\midrule
\textcolor{black}{**External Environment**} & The image does not show any signs of interaction between the nanomaterials and their surrounding environment; no solvents, polymers, or other materials are visible. The background is consistent and featureless, which suggests it is likely a vacuum or a solid substrate with no additional structures or objects present other than the nanomaterials themselves.
\\
 
\midrule
\textcolor{black}{**Image Technique and Modifications**} & The image is a raw, grayscale Scanning Electron Microscopy (SEM) micrograph without any apparent post-processing.
\\
 
\midrule
\textcolor{black}{**Functional Features**} & The SEM image does not reveal specific functional elements like active sites. It is a static representation and does not depict any dynamic processes within the nanomaterials.
\\

\midrule
\textcolor{black}{**Context and Application**} & The intended use or application of the nanomaterials and whether they are actual samples or theoretical representations cannot be determined from the image alone; additional context is needed.
\\ [-0.75em]

\end{tabularx}
\end{tcolorbox}
\end{adjustwidth}

\begin{adjustwidth}{-0.15cm}{-0.15cm}
\begin{tcolorbox}[colback=white!5!white,colframe=black!75!black]% [width=10cm]
\vspace{-5mm}
\begin{tabularx}{1.0\textwidth}{bss}
\caption{The table shows question-answer pairs about a microscopic image of nanowires nanomaterials, created with GPT-4 Turbo with vision. In summary, GPT-4 helps generate detailed instruction-tuning dataset about these nanomaterials images for training smaller multimodal models(SMMs), to serve as a domain-specific vision-language assistant for enterprise adoption.}\label{tab:tab4}\\ [-0.75em]
\toprule 
 \\ [-0.85em]
\begin{overpic}[width=0.15\columnwidth]{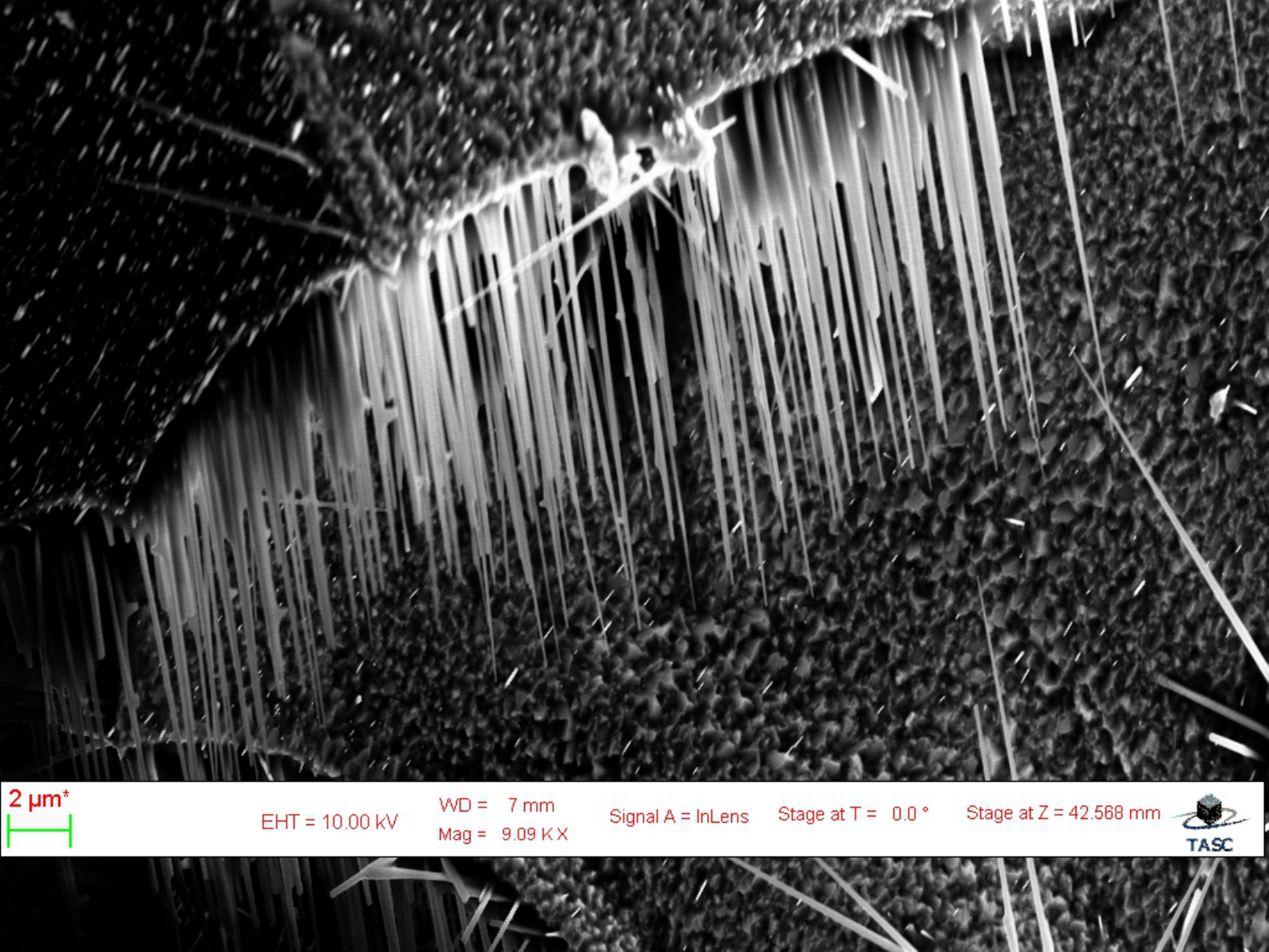}
  \put(450,40){\makebox[0pt][c]{%
    \parbox{0.65\columnwidth}{%
      \centering
      \textbf{Examine the provided input image and answer the following questions about the nanowires nanomaterial it depicts.} 
    }%
  }}  
\end{overpic}  \\ [-0.25em]

\midrule
\textbf{Query prompt}  & \textbf{GPT-4 Turbo with vision output} \\
\midrule
\endhead
\textcolor{black}{**Basics**} & The image depicts a forest of nanowires, a type of nanomaterial. The scale bar indicates 2 micrometers ($\mu$m), which means that one unit of measurement in the image corresponds to a real-world length of 2 micrometers..
\\

\midrule
\textcolor{black}{**Morphology and Structure**} &  The nanowires in the image exhibit a high aspect ratio, being much longer than they are wide, typical for this material. They are uniform in diameter and length, indicating a controlled synthesis process. The image resolution and contrast do not reveal distinct layers or phases. However, the nanowires are densely packed with a consistent orientation, characteristic of well-aligned nanowire arrays. They are consistent in size and shape across the visible field, suggesting a uniform growth mechanism. Variability is minimal, which is desirable for applications requiring predictable electronic, optical, and mechanical properties.
\\

\midrule
\textcolor{black}{**Size and Distribution**} & The nanowires appear to be evenly spaced and uniformly distributed. There's no visible evidence of aggregation or bundling, suggesting they are well-separated individual structures. Without precise measurements, estimating their diameter is challenging, but given the 2 $\mu$m scale bar, they likely range from tens to hundreds of nanometers in width.
\\

\midrule
\textcolor{black}{**Surface Characteristics**} & The nanowires have smooth surfaces without visible textures, pores or impurities, indicating high quality and clean synthesis.
\\

\midrule
\textcolor{black}{**Composition and Elements**} & No evidence of compositional variations or specific elements is discernible from the image provided. There are no labels indicating the presence of certain elements or compounds.
\\
 
\midrule
\textcolor{black}{**Interactions and Boundaries**} &  The nanostructures appear to be separate and not fused together, with clear boundaries between each nanowire. They are well-defined and do not blend into each other, indicating that they are individually distinct.
\\
 
\midrule
\textcolor{black}{**External Environment**} &  The image doesn't show any signs of interaction between the nanomaterials and an external environment or matrix such as solvents or polymers. The background appears relatively uniform and does not display any discernible structures or objects other than the nanomaterials themselves. Any such interactions or additional structures are not visible in this image.
\\
 
\midrule
\textcolor{black}{**Image Technique and Modifications**} &  The image is likely captured using Scanning Electron Microscopy (SEM). There are no evident post-processing modifications like false coloring or 3D rendering in the image.
\\
 
\midrule
\textcolor{black}{**Functional Features**} & The image appears to be a static representation and does not reveal specific functional elements like active sites or regions with distinct properties. No dynamic processes are depicted; it primarily shows the static structure of the nanowires.
\\

\midrule
\textcolor{black}{**Context and Application**} & The primary use of the nanowires cannot be determined from the image alone. They have various potential applications, but specific intent isn't clear. The image appears to depict actual experimental samples, not theoretical or simulation-based representations.
\\ [-0.75em]

\end{tabularx}
\end{tcolorbox}
\end{adjustwidth}

\begin{adjustwidth}{-0.15cm}{-0.15cm}
\begin{tcolorbox}[colback=white!5!white,colframe=black!75!black]% [width=10cm]
\vspace{-5mm}
\begin{tabularx}{1.0\textwidth}{bss}
\caption{The table shows question-answer pairs about a microscopic image of porous sponge nanomaterial, created with GPT-4 Turbo with vision. In summary, GPT-4 helps generate detailed instruction-tuning dataset about these nanomaterials images for training smaller multimodal models(SMMs), to serve as a domain-specific vision-language assistant for enterprise adoption.}\label{tab:tab5}\\ [-0.75em]
\toprule 
 \\ [-0.75em]
\begin{overpic}[width=0.15\columnwidth]{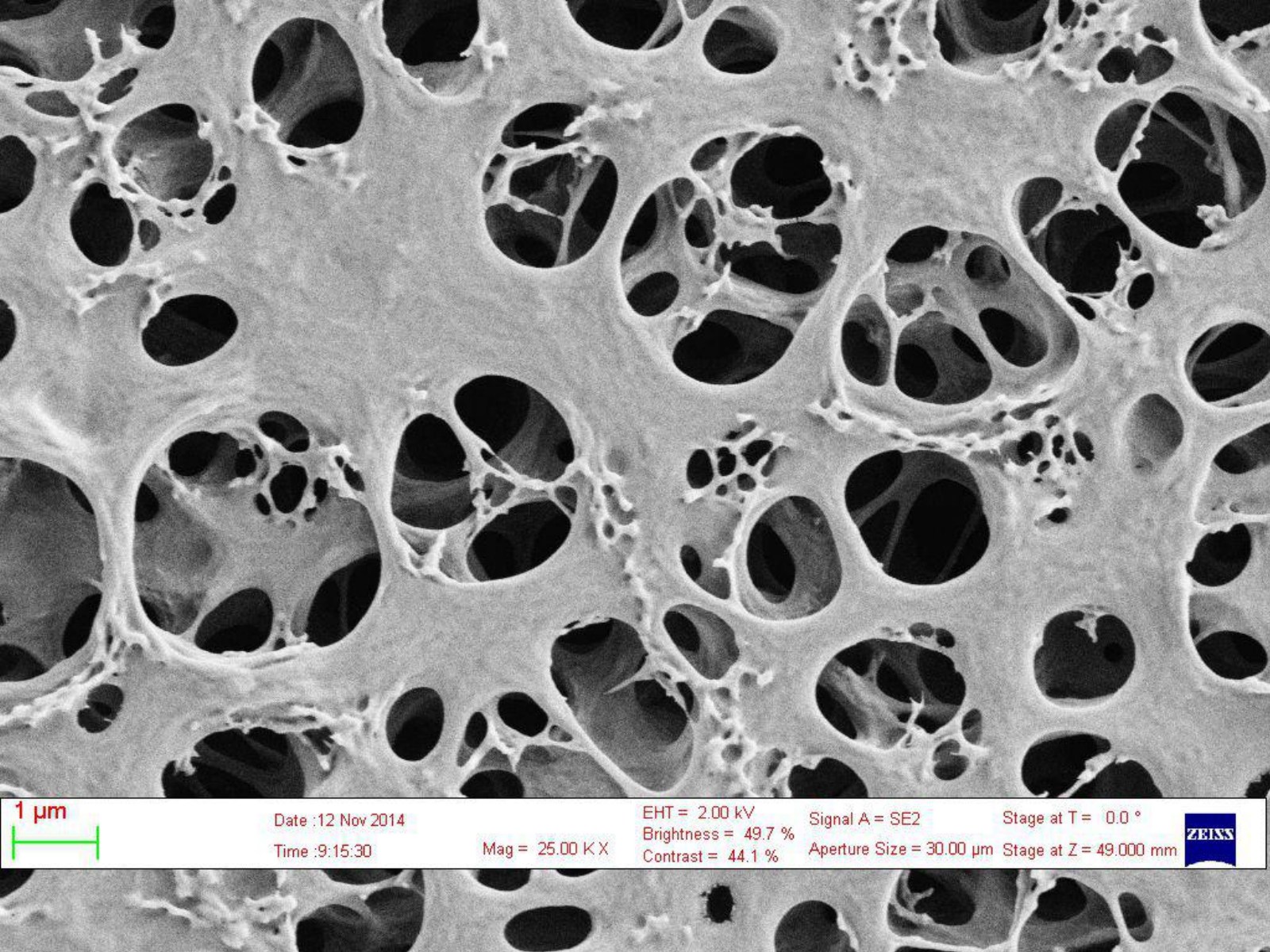}
  \put(425,40){\makebox[0pt][c]{%
    \parbox{0.65\columnwidth}{%
      \centering
      \textbf{Examine the provided input image and answer the following questions about the porous sponge nanomaterial it depicts.} 
    }%
  }}  
\end{overpic}  \\ [-0.25em]

\midrule
\textbf{Query prompt}  & \textbf{GPT-4 Turbo with vision output} \\
\midrule
\endhead
\textcolor{black}{**Basics**} & The image shows a nanoporous material. The scale bar indicates that 1 $\mu$m (micrometer) in the image corresponds to a real-world length of 1 $\mu$m.
\\

\midrule
\textcolor{black}{**Morphology and Structure**} & The nanomaterial depicted in the image exhibits a highly porous, sponge-like morphology with a network of interconnected voids and struts. There is a considerable variability in the pore sizes and shapes, with some being nearly circular while others are more irregular. There are no apparent layers or phases visible; it appears to be a homogenous single-phase material. The struts of the material also show variability in thickness. Overall, the structure is quite heterogeneous with a random, non-uniform distribution of pores and solid regions.
\\

\midrule
\textcolor{black}{**Size and Distribution**} & The nanoporous material has randomly distributed pores and struts that range in size from sub-micron to several microns across, based on the 1 $\mu$m scale bar. The nanoporous structure appears to be a single integrated network without evidence of nanostructure aggregation or bundling. This foam-like material with varying nanostructure sizes and random porosity is characteristic of aerogels and similar nanoporous materials
\\

\midrule
\textcolor{black}{**Surface Characteristics**} & The surfaces of the nanomaterials appear rough and irregular, with a high degree of texture likely due to the porous nature of the material. The pores themselves are the imperfections in the surface, and there could also be other defects not clearly distinguishable at this magnification and resolution.
\\

\midrule
\textcolor{black}{**Composition and Elements**} & The SEM image doesn't provide direct evidence of compositional variations or specific elements. Variations in brightness and contrast suggest differences in composition, but without additional data, we cannot identify specific elements or compounds. There are no markers indicating composition in the image.
\\
 
\midrule
\textcolor{black}{**Interactions and Boundaries**} & The nanostructures appear to be fused together, forming a continuous network with no clear boundaries between individual structures. The pores and struts blend into each other without defined borders, characteristic of a single-phase material with a homogenous composition.
\\
 
\midrule
\textcolor{black}{**External Environment**} & The SEM image only shows the nanomaterial itself, without any visible surrounding environment or matrix.
\\
 
\midrule
\textcolor{black}{**Image Technique and Modifications**} &  The image is captured using Scanning Electron Microscopy (SEM). There are no apparent post-processing modifications like false coloring or 3D rendering; it appears to be a standard grayscale SEM image.
\\
 
\midrule
\textcolor{black}{**Functional Features**} & The SEM image is a static representation that reveals surface morphology and structure, but does not show dynamic processes or functional properties.
\\

\midrule
\textcolor{black}{**Context and Application**} & The image likely shows an actual experimental sample of a nanoporous material. Its specific intended use or application is not clear from the image alone, but such materials are commonly used in catalysis, filtration, sensors, and energy storage.
\\ [-0.75em]

\end{tabularx}
\end{tcolorbox}
\end{adjustwidth}

\begin{adjustwidth}{-0.15cm}{-0.15cm}
\begin{tcolorbox}[colback=white!5!white,colframe=black!75!black]% [width=10cm]
\vspace{-5mm}
\begin{tabularx}{1.0\textwidth}{bss}
\caption{The table shows question-answer pairs about a microscopic image of tips nanomaterial, created with GPT-4 Turbo with vision. In summary, GPT-4 helps generate detailed instruction-tuning dataset about these nanomaterials images for training smaller multimodal models(SMMs), to serve as a domain-specific vision-language assistant for enterprise adoption.}\label{tab:tab6}\\ [-0.75em]
\toprule 
 \\ [-3.5em]
\begin{overpic}[width=0.15\columnwidth]{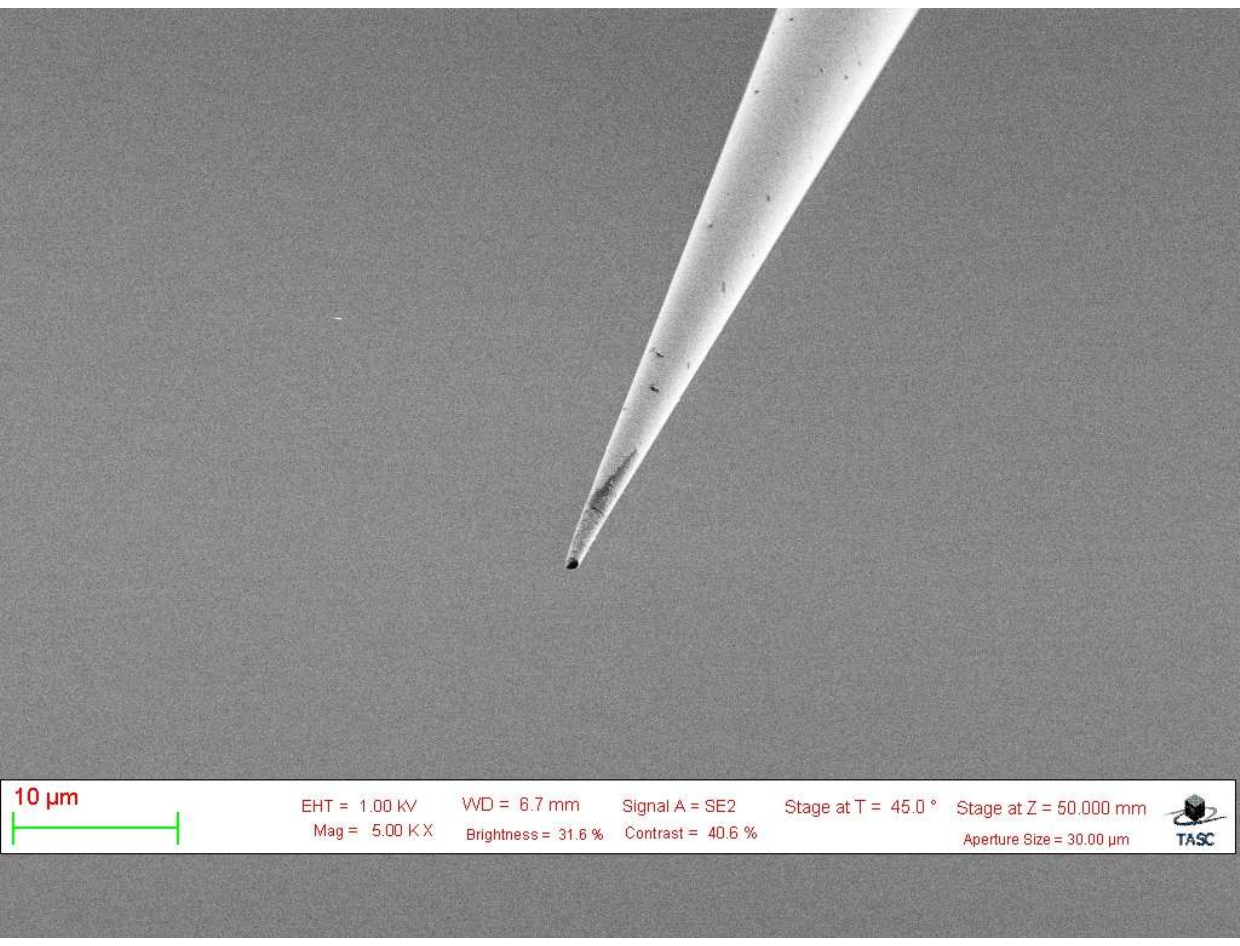}
  \put(300,50){\makebox[0pt][c]{%
    \parbox{0.65\columnwidth}{%
      \centering
      \textbf{Examine the provided input image and answer the following questions about the tip nanomaterial it depicts.} 
    }%
  }}  
\end{overpic}  \\ [-2.5em]

\midrule
\textbf{Query prompt}  & \textbf{GPT-4 Turbo with vision output} \\
\midrule
\endhead
\textcolor{black}{**Basics**} & The image shows a tip nanomaterial, likely a nanowire or nanorod, viewed under a scanning electron microscope (SEM). The scale bar represents 10 micrometers ($\mu$m), indicating the magnified view's resolution.
\\

\midrule
\textcolor{black}{**Morphology and Structure**} & The nanomaterial image shows elongated, needle-like shapes typical of nanowires or nanorods. The surface is relatively smooth with minor roughness, which is common at this scale. There are no visibly distinct layers, phases, or domains. The structure is uniform without clear segmentation. The material exhibits a consistent shape, tapering to a finer point at the tip. There does not appear to be significant variability in size along its length, indicating a uniform growth or fabrication process. However, without a wider view of these nanomaterials, it is difficult to comment on the consistency across multiple specimens.
\\

\midrule
\textcolor{black}{**Size and Distribution**} & The image shows a single nanomaterial less than 1 micrometer wide based on the 10 micrometer scale bar. The full length cannot be determined from this limited view. With only one nanostructure visible, no conclusions about distribution, spacing, clustering, or aggregation can be drawn. A wider view with multiple nanostructures is necessary to assess those characteristics. The key observation from this image is the estimated maximum width of this nanomaterial.
\\

\midrule
\textcolor{black}{**Surface Characteristics**} & The surface of the nanomaterial appears predominantly smooth with some minor roughness. There are no clearly visible defects, pores, or impurities on the surface at this magnification.
\\

\midrule
\textcolor{black}{**Composition and Elements**} &  The uniform grayscale contrast implies consistent nanomaterial composition but lacks chemical details. Additional analyses like EDX are needed with SEM to determine elemental composition, since SEM provides surface topology not chemistry.
\\
 
\midrule
\textcolor{black}{**Interactions and Boundaries**} &  The image only shows a single nanostructure, so interactions or boundaries with other structures cannot be assessed. There are no visible boundaries or distinct phases within this nanomaterial.
\\
 
\midrule
\textcolor{black}{**External Environment**} &  No signs of interaction with a surrounding environment or other materials are visible in the image, and there are no other structures or objects present besides the nanomaterial.
\\
 
\midrule
\textcolor{black}{**Image Technique and Modifications**} & The image is captured using Scanning Electron Microscopy (SEM). There are no apparent post-processing modifications like false coloring or 3D rendering; it appears to be a standard grayscale SEM image.
\\
 
\midrule
\textcolor{black}{**Functional Features**} & The SEM image does not reveal specific functional elements like active sites, and it represents a static view of the nanomaterial. SEM images typically don't capture dynamic processes.
\\

\midrule
\textcolor{black}{**Context and Application**} &  The primary use or application of the nanomaterials in the image cannot be determined from the SEM image alone. The image depicts an actual experimental sample, not a theoretical or simulation-based representation. Applications for such nanostructures vary widely, including electronics, sensors, and biomedical devices
\\ [-0.75em]

\end{tabularx}
\end{tcolorbox}
\end{adjustwidth}

\begin{adjustwidth}{-0.15cm}{-0.15cm}
\begin{tcolorbox}[colback=white!5!white,colframe=black!75!black]% [width=10cm]
\vspace{-5mm}
\begin{tabularx}{1.0\textwidth}{bss}
\caption{The table shows question-answer pairs about a microscopic image of particles nanomaterial, created with GPT-4 Turbo with vision. In summary, GPT-4 helps generate detailed instruction-tuning dataset about these nanomaterials images for training smaller multimodal models(SMMs), to serve as a domain-specific vision-language assistant for enterprise adoption.}\label{tab:tab7}\\ [-0.75em]
\toprule 
 \\ [-0.75em]
\begin{overpic}[width=0.15\columnwidth]{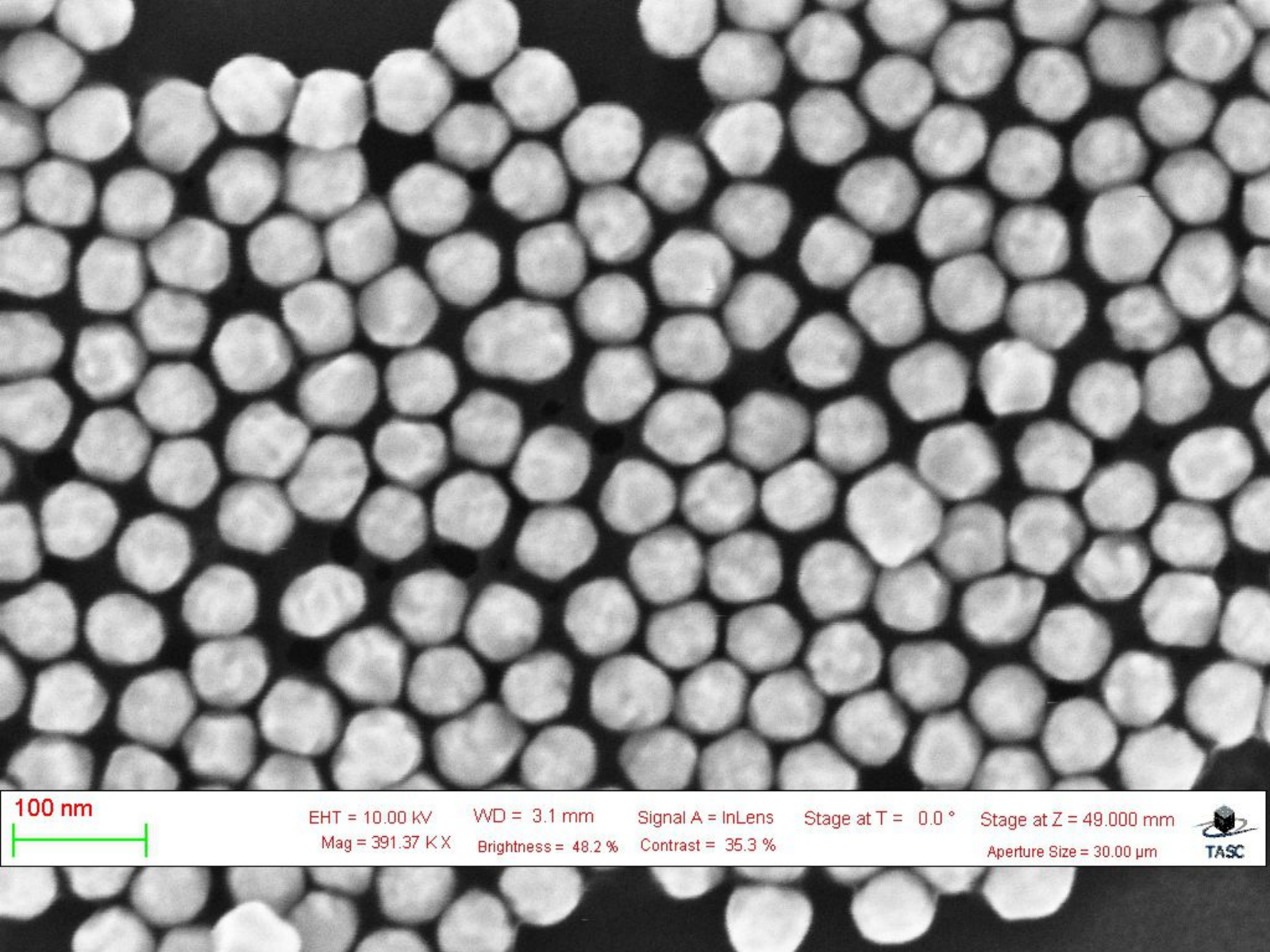}
  \put(425,40){\makebox[0pt][c]{%
    \parbox{0.65\columnwidth}{%
      \centering
      \textbf{Examine the provided input image and answer the following questions about the particles nanomaterial it depicts.} 
    }%
  }}  
\end{overpic}  \\ [-0.25em]

\midrule
\textbf{Query prompt}  & \textbf{GPT-4 Turbo with vision output} \\
\midrule
\endhead
\textcolor{black}{**Basics**} & The image depicts spherical nanoparticles, and the scale bar indicates that 100 nanometers (nm) corresponds to the real-world length in the image.
\\

\midrule
\textcolor{black}{**Morphology and Structure**} & The nanomaterials in the image have a spherical shape with a consistent size and shape throughout, indicating a uniform and monodisperse batch. There are no visible layers, phases, or distinct domains within the individual nanoparticles; they appear to be solid and single-phased under the resolution provided by the SEM. The uniformity suggests a well-controlled synthesis.
\\

\midrule
\textcolor{black}{**Size and Distribution**} & The individual nanostructures are uniform in size. Although exact measurements require specialized software, an estimate can be made using the scale bar, which represents 100 nanometers. The particles are slightly smaller than the scale bar, likely around 70-90 nanometers in diameter. They are distributed in an orderly, hexagonal close-packed arrangement, indicating even spacing without randomness. There is no evidence of aggregation or bundling; the nanoparticles maintain distinct boundaries and do not appear fused or clumped, suggesting good stability without a tendency to aggregate under current conditions.
\\

\midrule
\textcolor{black}{**Surface Characteristics**} & The surfaces of the nanomaterials appear predominantly smooth without distinct textures. There are no noticeable imperfections such as defects, pores, or impurities visible on the surfaces of these nanomaterials in the image.
\\

\midrule
\textcolor{black}{**Composition and Elements**} & There are no visible compositional variations or specific markers indicating different elements in the image; it all appears uniform in grayscale, suggesting a consistent material composition throughout the nanoparticles.
\\
 
\midrule
\textcolor{black}{**Interactions and Boundaries**} & The individual nanostructures appear to be touching but not fused, with clear boundaries between each particle, suggesting they are separate entities. There's no blending of borders, indicating distinct nanostructures.
\\
 
\midrule
\textcolor{black}{**External Environment**} & There are no signs of interaction between the nanomaterials and any surrounding matrix in the image. It shows just the nanoparticles without any visible solvents, polymers, or other materials. There are also no discernible structures or objects present that are not nanomaterials; the entire field of view is occupied by the nanoparticles.
\\
 
\midrule
\textcolor{black}{**Image Technique and Modifications**} & The image characteristics and footer information indicate this is a standard, unprocessed grayscale SEM image. There is no evidence of techniques like false coloring or 3D rendering applied; it appears to be raw SEM imaging data.
\\
 
\midrule
\textcolor{black}{**Functional Features**} & The image does not show any specific functional elements or active sites on the nanoparticles. It is a static representation, with no dynamic processes depicted.
\\

\midrule
\textcolor{black}{**Context and Application**} & The intended use or application of the nanomaterials cannot be determined from the image alone, and it's unclear whether they represent actual experimental samples or are theoretical/simulation-based.
\\ [-0.75em]

\end{tabularx}
\end{tcolorbox}
\end{adjustwidth}

\begin{adjustwidth}{-0.15cm}{-0.15cm}
\begin{tcolorbox}[colback=white!5!white,colframe=black!75!black]% [width=10cm]
\vspace{-5mm}
\begin{tabularx}{1.0\textwidth}{bss}
\caption{The table shows question-answer pairs about a microscopic image of powder nanomaterial, created with GPT-4 Turbo with vision. In summary, GPT-4 helps generate detailed instruction-tuning dataset about these nanomaterials images for training smaller multimodal models(SMMs), to serve as a domain-specific vision-language assistant for enterprise adoption.}\label{tab:tab8}\\ [-0.85em]
\toprule 
\begin{overpic}[width=0.15\columnwidth]{Images/GPT4V-Images/Powder/Image.pdf}
  \put(425,40){\makebox[0pt][c]{%
    \parbox{0.65\columnwidth}{%
      \centering
      \textbf{Examine the provided input image and answer the following questions about the powder nanomaterial it depicts.} 
    }%
  }}  
\end{overpic}  \\ [-0.25em]

\midrule
\textbf{Query prompt}  & \textbf{GPT-4 Turbo with vision output} \\
\midrule
\endhead
\textcolor{black}{**Basics**} &  The image is a scanning electron microscope (SEM) image of a powder nanomaterial. The type of nanomaterial cannot be determined from the image alone. The scale bar indicates a measurement of 2 micrometers ($\mu$m), meaning each unit on the scale corresponds to real-world lengths of 2 $\mu$m.
\\

\midrule
\textcolor{black}{**Morphology and Structure**} &  The nanomaterials in the SEM image appear to have an irregular, clustered morphology with varying sizes and shapes, which suggests they do not exhibit a consistent size and shape throughout. There are no clear layers, phases, or distinct domains visible; the material seems to be agglomerated, with particles fusing together in some areas. The variability in size and shape is common in powder nanomaterials, which can result from synthesis conditions or post-synthesis treatments.
\\

\midrule
\textcolor{black}{**Size and Distribution**} & The nanostructures vary in size from sub-micrometers to a few micrometers, with uneven spacing and visible clustering and aggregation, typical for nanopowder samples viewed at high magnification.
\\

\midrule
\textcolor{black}{**Surface Characteristics**} & The nanomaterials in the image exhibit a rough and irregular surface texture, with no distinct patterns. The surfaces are not smooth, which is typical for powdered nanomaterials due to their high surface area to volume ratio. There are noticeable imperfections such as non-uniformities in size and shape, which could indicate defects, pores, or impurities. These imperfections are common in nanostructures and can affect their physical and chemical properties.
\\

\midrule
\textcolor{black}{**Composition and Elements**} &  The SEM image does not provide evidence of compositional variations, as it is in grayscale and changes in brightness or contrast can be due to topography rather than composition. There are no labels or markers indicating the presence of specific elements or compounds within the image.
\\
 
\midrule
\textcolor{black}{**Interactions and Boundaries**} & The nanostructures appear to be predominantly clustered and touching, with many of them fused together, creating a cohesive mass. This suggests a high degree of interaction between the particles. It is challenging to distinguish clear boundaries between different structures or phases; they blend together without defined borders, which is indicative of aggregation common in powders at the nanoscale.
\\
 
\midrule
\textcolor{black}{**External Environment**} &  The SEM image shows the nanomaterials with no visible surrounding environment or external materials.
\\
 
\midrule
\textcolor{black}{**Image Technique and Modifications**} & The image is an unmodified scanning electron microscopy (SEM) micrograph showing nanomaterial morphology without post-processing.
\\
 
\midrule
\textcolor{black}{**Functional Features**} & The SEM image is a static snapshot showing morphology. No dynamic processes or functional elements are discernible without additional data.
\\

\midrule
\textcolor{black}{**Context and Application**} & The SEM image depicts experimental nanomaterials with unspecified applications, as nanomaterials have diverse uses and require additional context. The image itself does not indicate a primary intended application.
\\ [-0.75em]

\end{tabularx}
\end{tcolorbox}
\end{adjustwidth}

\begin{adjustwidth}{-0.15cm}{-0.15cm} 
\begin{tcolorbox}[colback=white!5!white,colframe=black!75!black]% [width=10cm]
\vspace{-5mm}
\begin{tabularx}{1.0\textwidth}{bss}
\caption{The table shows question-answer pairs about a microscopic image of MEMS devices and electrodes, created with GPT-4 Turbo with vision. In summary, GPT-4 helps generate detailed instruction-tuning dataset about these nanomaterials images for training smaller multimodal models(SMMs), to serve as a domain-specific vision-language assistant for enterprise adoption.}\label{tab:tab9}\\ [-0.85em]
\toprule 
\begin{overpic}[width=0.15\columnwidth]{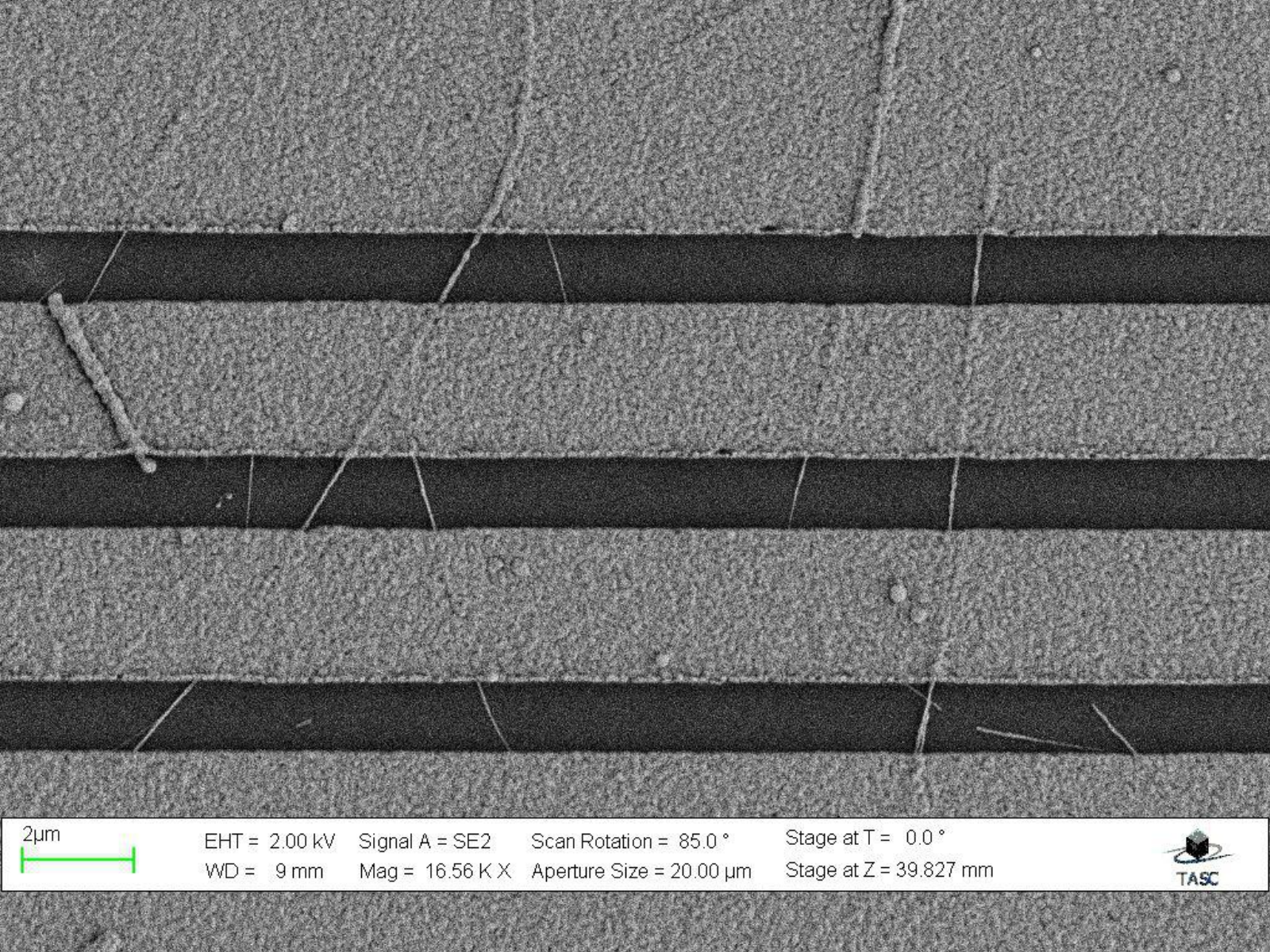}
  \put(425,40){\makebox[0pt][c]{%
    \parbox{0.65\columnwidth}{%
      \centering
      \textbf{Examine the provided input image and answer the following questions about the MEMS devices and electrodes nanomaterial it depicts.} 
    }%
  }}  
\end{overpic}  \\ [-0.25em]

\midrule
\textbf{Query prompt}  & \textbf{GPT-4 Turbo with vision output} \\
\midrule
\endhead
\textcolor{black}{**Basics**} & The image is a scanning electron microscope (SEM) picture of a MEMS device. The scale bar shows 2 micrometers, indicating the resolution of the image.
\\

\midrule
\textcolor{black}{**Morphology and Structure**} & The image shows parallel lines with uniform width and spacing, suggesting consistent size and shape, indicative of structured electrodes or components of a MEMS device. There are no visible layers or phases within the materials, and no variability in size and shape is apparent among the nanomaterials themselves.
\\

\midrule
\textcolor{black}{**Size and Distribution**} &  The parallel nanostructures have a consistent size and even spacing, suggesting precise fabrication techniques, with no visible aggregation or bundling, indicating high design and fabrication precision.
\\

\midrule
\textcolor{black}{**Surface Characteristics**} & The surfaces of the nanostructures in the image appear predominantly smooth, which is typical for fabricated MEMS components. There are some small particles and imperfections visible on the surface, which could be due to impurities, defects, or debris.
\\

\midrule
\textcolor{black}{**Composition and Elements**} & The image does not show color variations, and there are no labels indicating specific elements or compounds. The consistent contrast suggests uniform material composition for the structures displayed.
\\
 
\midrule
\textcolor{black}{**Interactions and Boundaries**} & The nanostructures appear fully separate and not fused together, with clear and distinct boundaries between them. The image shows no blending of structures; each one is individually distinguishable.
\\
 
\midrule
\textcolor{black}{**External Environment**} & The nanostructures are embedded in a matrix with no clear signs of interaction. Small particles and irregularities are present on the surface, likely impurities or debris, but not part of the nanomaterials themselves.
\\
 
\midrule
\textcolor{black}{**Image Technique and Modifications**} & The image is likely captured using Scanning Electron Microscopy (SEM), as indicated by the surface detail and depth of field. There is no evidence of post-processing techniques like false coloring or 3D rendering; it appears as a standard grayscale SEM image.
\\
 
\midrule
\textcolor{black}{**Functional Features**} & The image appears static and does not depict dynamic processes. It shows uniform structures, likely part of a MEMS device, but without additional context, specific functional elements like active sites cannot be conclusively identified. The image primarily provides a static representation of the nanostructures.
\\

\midrule
\textcolor{black}{**Context and Application**} & The nanomaterials are likely for microelectronics or sensors, typical of MEMS devices. The image, resembling real SEM samples, suggests it's based on actual experimental samples rather than being theoretical or simulation-based.
\\[-0.75em]

\end{tabularx}
\end{tcolorbox}
\end{adjustwidth}

\twocolumn

\vspace{-2mm}
\subsection{Additional datasets and Experimental results}
\vspace{-1mm}
To assess the robustness and applicability of our framework, we conducted a comprehensive evaluation using a diverse set of open-source benchmark datasets. We carefully selected datasets that were relevant to our research domain and encompassed a broad spectrum of applications, ensuring a generalizable evaluation process. This rigorous approach not only verified the effectiveness of our framework on these established datasets but also demonstrated its adaptability to a wide range of scenarios. This is particularly significant because our framework extends beyond the SEM dataset\cite{aversa2018first} for which it was initially developed, showcasing its potential for real-world use cases.

\vspace{-4mm}
\begin{figure}[ht!]
    \centering
    \includegraphics[width=8.7cm]{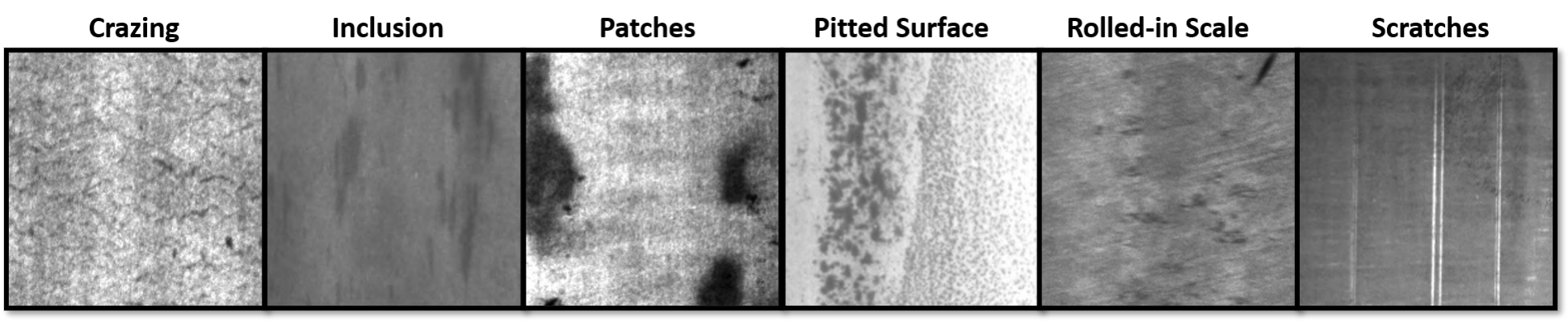}
    \vspace{-8mm}
    \caption{The figure showcases a selection of electron microscopy images from the NEU-SDD dataset\cite{deshpande2020one}, which clearly illustrate six common types of surface defects found on hot-rolled steel strips: \textit{pitted surfaces, scratches, rolled-in scale, crazing, patches, and inclusion defects}. These microscopic images provide a comprehensive visual representation of the various types of defects that can occur on steel surfaces, allowing for a better understanding of their characteristics and potential impact on the material's properties and performance.}
    \vspace{-1mm}
    \label{fig:addataset3}
    \vspace{-2mm}
\end{figure}

\vspace{-3mm}
\subsubsection{NEU-SDD(\cite{deshpande2020one})} 
\vspace{-1mm}
To rigorously evaluate our proposed framework's performance on zero/few-shot label prediction and VQA tasks for steel material surface defects, we leveraged the comprehensive NEU-SDD dataset\footnote{Datasource: \url{http://faculty.neu.edu.cn/yunhyan/NEU_surface_defect_database.html}\label{note1}}. The diverse dataset encompasses a variety of surface defect types, making it well-suited for assessing the generalizability of the proposed framework's performance. The dataset includes an extensive collection of 1,800 electron microscopy images depicting surface defects on hot-rolled steel plates, providing a comprehensive resource for evaluating our framework's ability to understand complex visual information and answer insightful questions about the surface defects. The NEU-SDD dataset comprises grayscale images, each having a dimension of 200 $\times$ 200 pixels, and is carefully classified into six distinct defect types, with 300 representative images for each category. These categories depict a diverse range of surface imperfections, including pitted surfaces, scratches, rolled-in scale, crazing, patches, and inclusion defects. Figure~\ref{fig:addataset3} provides illustrative images from each defect category. The NEU-SDD dataset is a valuable benchmark for developing and testing algorithms that can answer questions about images of surface defects. Its large size, diversity of defect types depicted, and high-quality images make it a demanding and representative dataset for evaluating VQA methods in various surface defect contexts.

\vspace{-3mm}
\subsubsection{Corrosion Monitoring Inspection(CMI)}
\begin{figure}[ht!]
    \centering
    \includegraphics[width=6.25cm]{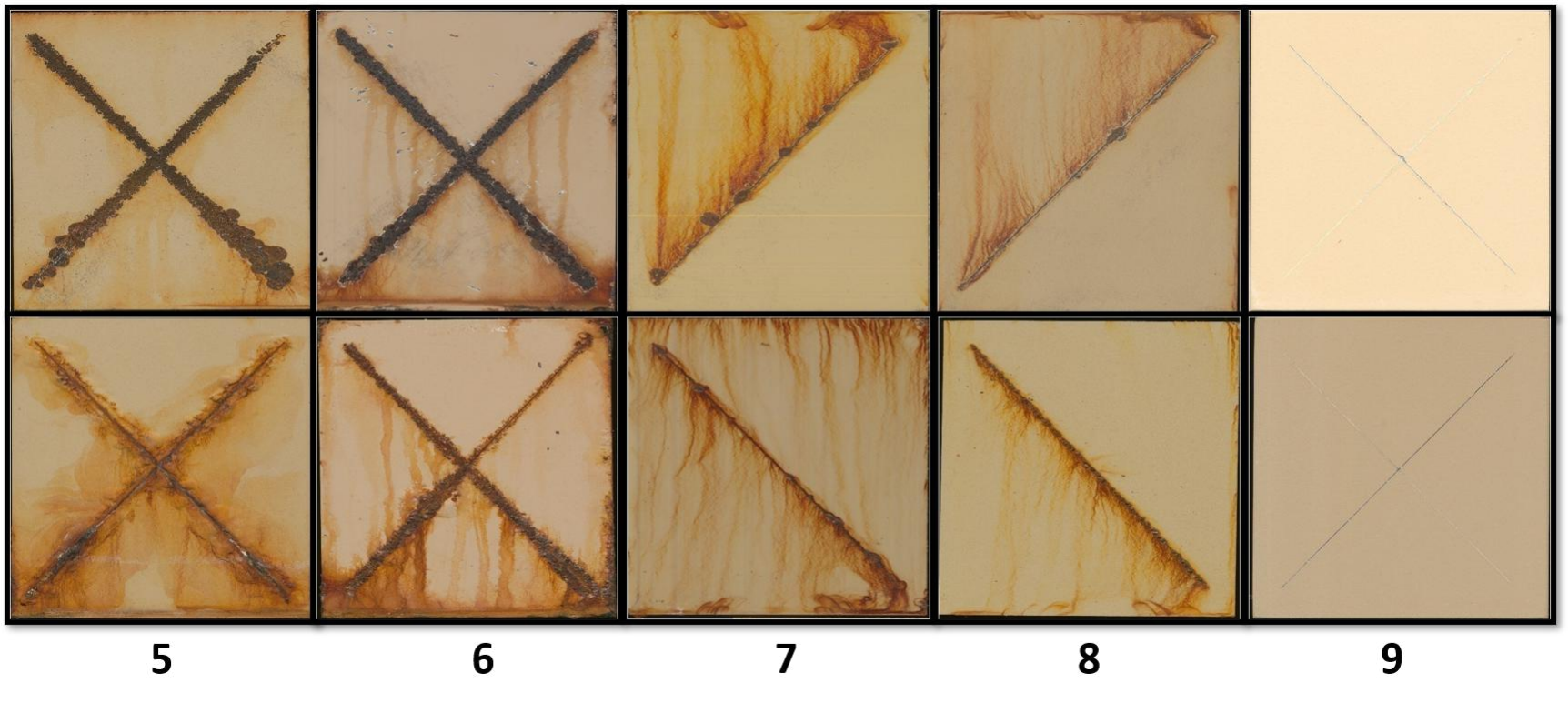}
    \vspace{-3mm}
    \caption{The figure displays a selection of meticulously classified electron micrographs from the CMI dataset. Each micrograph is assigned a score (ranging from 5 to 9, with higher scores indicating less severe corrosion) according to ASTM-D1654 standards. These micrographs illustrate a progression of increasing corrosion severity (due to pitting, thinning, cracking) as the score decreases, thus reflecting more extensive damage. This diverse collection of electron micrographs, encompassing the entire spectrum of corrosion severity levels, facilitates the development and evaluation of cutting-edge algorithms for precise corrosion assessment. Moreover, it provides a realistic and faithful representation of corrosion damage across various degrees of severity.}
    \label{fig:addataset1}
    \vspace{-3mm}
\end{figure}

\vspace{-2mm}
The CMI dataset\footnote{\url{https://arl.wpi.edu/corrosion_dataset}\label{note2}} contains 600 detailed electron micrographs of corroded panels, carefully curated by corrosion experts. This collection of images vividly captures deterioration across varying severity levels of corrosion damage. The images are classified according to the ASTM-D1654 standards, with individual scores ranging from 5 to 9 (with higher scores indicating less corrosion severity), with 120 unique micrographs per score. Each high-resolution micrograph, measuring 512 $\times$ 512 pixels, provides a granular view of the corrosion damage. We used the CMI dataset (as shown in Figure~\ref{fig:addataset1} with representative images from each scoring category) to conduct experimental studies evaluating the effectiveness of our proposed framework for zero/few-shot prediction and VQA tasks.

\vspace{-2mm}

\vspace{-2mm}
\subsubsection{KTH-Tips} 
\begin{figure*}[ht!]
    \centering
    \includegraphics[width=6.15cm]{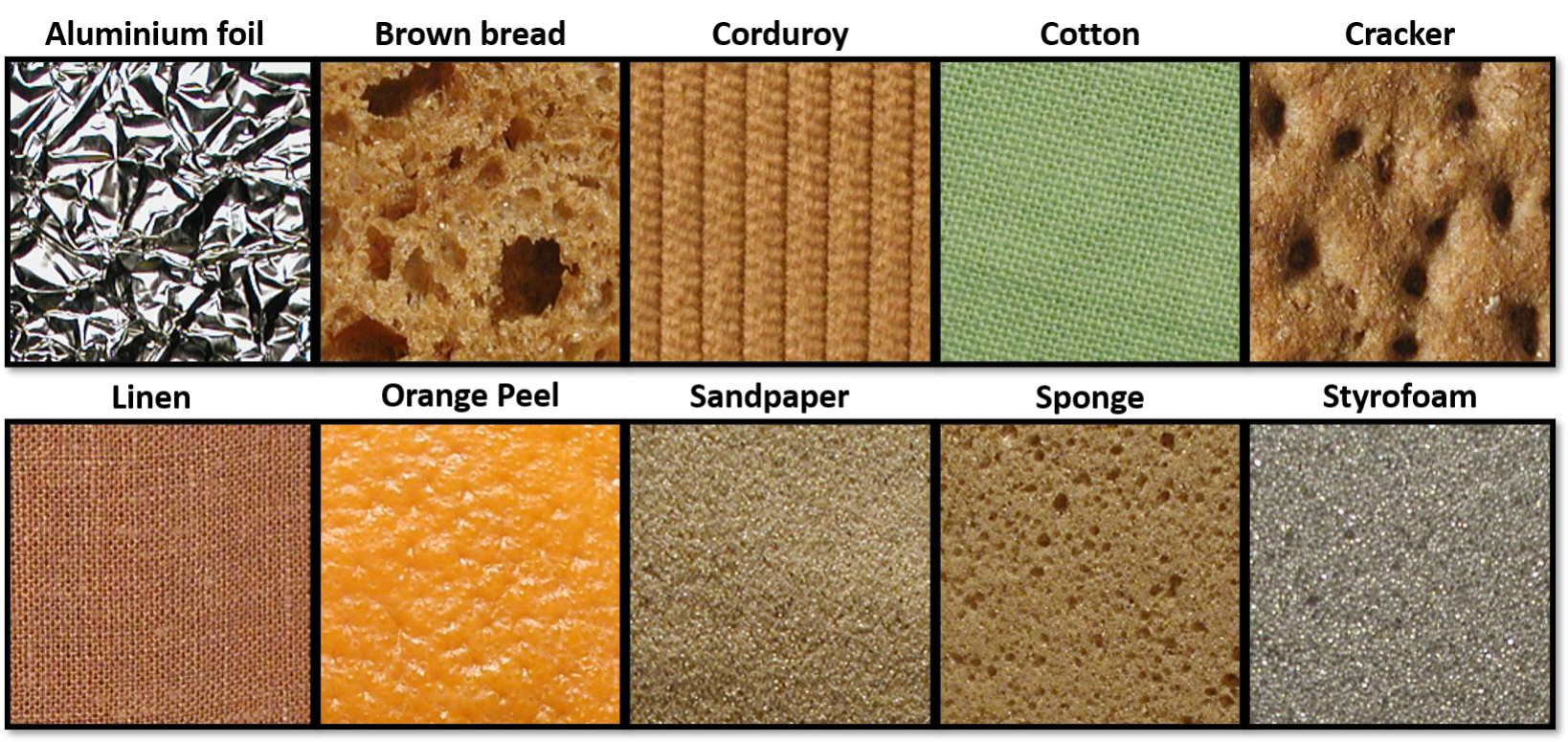}
    \vspace{-2mm}
    \caption{The figure presents a sample collection of electron micrographs from the KTH-TIPS texture dataset. These micrographs showcase the ten distinct material classes, including \textit{sponge, orange peel, styrofoam, cotton, cracker, linen, crust, sandpaper, aluminum foil, and corduroy}.    
    }
    \label{fig:addataset2}
    \vspace{0mm}
\end{figure*}

\vspace{-1mm}
The KTH-TIPS dataset\footnote{\url{https://www.csc.kth.se/cvap/databases/kth-tips/index.html}\label{note3}}, a seminal benchmark in texture analysis, comprises an extensive collection of 810 high-resolution electron micrographs. Each image, having a dimension of 200 $\times$ 200 pixels, has been meticulously categorized into one of ten distinct material classes, showcasing a rich diversity of textures. Included are materials such as \textit{sponge, orange peel, styrofoam, cotton, cracker, linen, crust, sandpaper, aluminum foil, and corduroy}. The microscopic images capture each texture under varying real-world conditions, such as differences in lighting, orientation, and scale. This versatility makes the KTH-TIPS dataset challenging and comprehensive for evaluating texture recognition and analysis methods. Figure \ref{fig:addataset2} presents illustrative samples from each of the ten material categories.

\vspace{0mm}
\begin{table*}[!ht]
\centering
\caption{The table compares the performance of our proposed framework on open-ended VQA tasks across benchmark datasets to several well-known baselines.}
\vspace{2mm}
\scalebox{0.80}{
\hspace*{-5mm}\begin{tabular}{l|c|c|c|c|c|c}
\toprule
Method & BLEU-2 ($\uparrow$) & BLEU-4 ($\uparrow$) & ROUGE-1 ($\uparrow$) & ROUGE-2 ($\uparrow$) & ROUGE-L ($\uparrow$) & METEOR ($\uparrow$) \\ 
\midrule
\begin{tabular}[c]{@{}l@{}} InstructBLIP\cite{dai2305instructblip} \end{tabular} & 0.808 & 0.656 & 0.920 & 0.817 & 0.884 & 0.939 \\ 
\midrule
\begin{tabular}[c]{@{}l@{}}LLaVA\cite{liu2023visual} \end{tabular} & 0.806 & 0.657 & 0.935 & 0.818 & 0.890 & 0.937 \\ 
\midrule
\begin{tabular}[c]{@{}l@{}}MiniGPT-4\cite{zhu2023minigpt} \end{tabular} & 0.839 & 0.672 & 0.939 & 0.813 & 0.879 & 0.957 \\ 
\midrule
\begin{tabular}[c]{@{}l@{}}\textbf{sLAVA} \end{tabular} & \textbf{0.929} & \textbf{0.839} & \textbf{0.971} & \textbf{0.914} & \textbf{0.965} & \textbf{0.988} \\  
\bottomrule
\end{tabular}
}
\label{captioning_results6}
\vspace{0mm}
\end{table*}

\vspace{0mm}
\begin{table*}[ht!]
\footnotesize
\centering
\resizebox{0.40\textwidth}{!}{%
\subfloat{%
\setlength{\tabcolsep}{3pt}
\begin{tabular}{cc|cccc}
\hline
\multicolumn{2}{c|}{\textbf{Algorithms}}                                       & \textbf{NEU-SDD} & \textbf{CMI} & \textbf{KTH-TIPS}  \\ \hline
\multicolumn{1}{c|}{\multirow{4}{*}{\rotatebox[origin=c]{90}{\textbf{Baselines}}}} & ResNet                   & 0.906	& 0.928	& 0.941 &             \\
\multicolumn{1}{c|}{}                                          & GoogleNet                & 0.936	& 0.928	& 0.929
              \\
\multicolumn{1}{c|}{}                                          & SqueezeNet                & 0.955	& 0.943	& 0.963
              \\ 
\multicolumn{1}{c|}{}                                          & VanillaViT               & 0.962	& 0.968	& 0.972
 \\ 
\hline
\multicolumn{1}{c|}{}                                          &  \textbf{sLAVA}                  &     \textbf{0.992}              &      \textbf{0.987}             &    \textbf{0.989}               &                     \\ \hline
\end{tabular}}}
\vspace{-1mm}
\caption{The table compares the performance of the proposed framework to well-established baselines on benchmark datasets for multi-class classification.}
\label{tab:auxexp}
\vspace{0mm}
\end{table*}

To evaluate the multi-category texture recognition capabilities of our proposed method, particularly for zero/few-shot classification and VQA tasks, we conducted thorough experiments on this comprehensive dataset.

\vspace{0mm}
\subsubsection{Additional Information} 
\vspace{0mm}
The belief that a single set of prompts can effectively enable advanced AI models, like GPT-4 Turbo with Vision, to handle various image datasets is flawed. Instead, specialized prompts tailored to specific tasks are necessary for the accurate generation of instruction-following datasets to customize small-scale vision-language models. This tailored approach maximizes the capabilities of sophisticated small-scale multimodal models and ensures their most effective use. For instance, in our supplementary experiments, we demonstrate this by employing GPT-4 Turbo with Vision to generate high-quality question-answer pairs for microscopy images, using customized prompts for each dataset. Our approach, which trains small-scale multimodal models with expert-generated instruction-answer pairs, allows these smaller models to excel in complex tasks like nanomaterial image analysis. This highlights the importance of specialized prompting and instruction tuning in creating small-scale multimodal  models. The performance of the proposed framework, \texttt{sLAVA}, is compared with baseline models in multi-class classification tasks (accuracy is shown in Table~\ref{tab:auxexp}) and in open-ended Visual Question Answering (VQA) tasks (detailed in Table~\ref{captioning_results6}). To illustrate open-ended VQA performance, Tables \ref{tab:onvqa1}, \ref{tab:onvqa2}, and \ref{tab:onvqa3} present examples including images, questions, and the generated answers/descriptions. These tables offer more than just text comparisons, including performance evaluation metrics like BLEU-2, ROUGE-L, and METEOR for the generated text.  Additionally, Tables \ref{tab:tab11} - \ref{tab:tab17} show samples from the instruction-tuning Q\&A pairs dataset, generated by GPT-4 Turbo with Vision for training smaller multimodal models.

\onecolumn

\section{CMI}
\begin{adjustwidth}{-0.15cm}{-0.15cm} 
\begin{tcolorbox}[colback=white!5!white,colframe=black!75!black]% [width=10cm]
\vspace{-5mm}
\begin{tabularx}{1.0\textwidth}{bss}
\caption{The table shows question-answer pairs about a microscopic image of the corrosion 5 category from the CMI dataset, created with GPT-4 Turbo with vision. In summary, GPT-4 helps generate detailed instruction-tuning dataset about these microscopic images for training smaller multimodal models(SMMs), to serve as a domain-specific vision-language assistant for enterprise adoption.}\label{tab:tab11}\\ [-0.85em]
\toprule
\begin{overpic}[width=0.15\columnwidth]{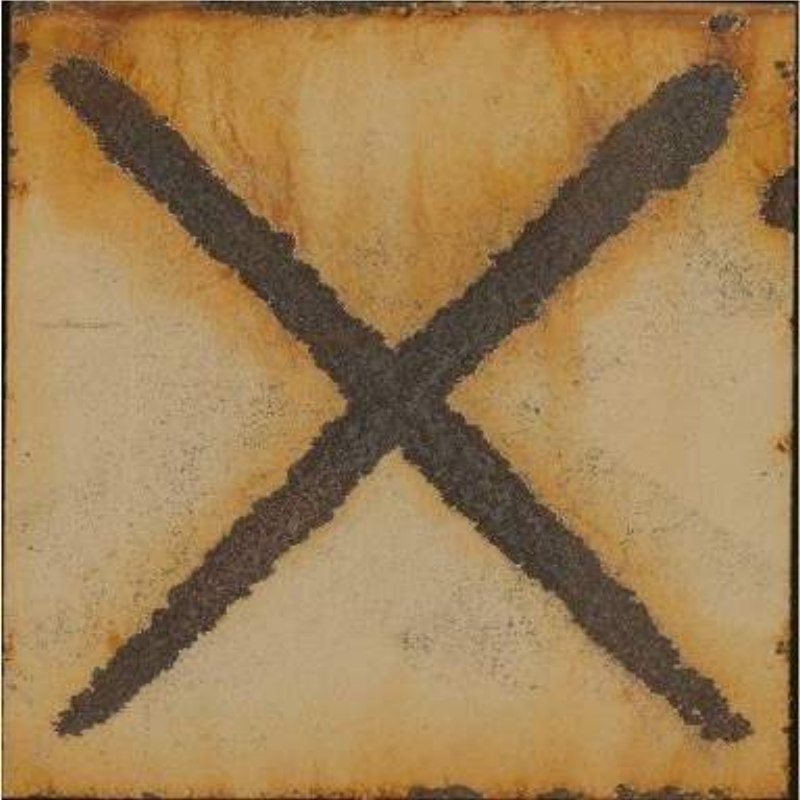}
\vspace{-5mm}
  \put(450,50){\makebox[0pt][c]{%
    \parbox{0.65\columnwidth}{%
      \centering
      \textbf{Examine the provided input image and answer the following questions about the corrosion 5 category it depicts.} 
    }%
  }}  
\end{overpic}  \\

\midrule
\textbf{Query prompt}  & \textbf{GPT-4 Turbo with vision output} \\
\midrule
\endhead
\textcolor{black}{What is the primary purpose of the image?} & The primary purpose of this image is to serve as a detailed visual record of corrosion, categorized under the corrosion 5 category, according to specific standardized criteria for assessing and studying corrosion damage.
\\

\midrule
\textcolor{black}{How is the severity of corrosion visually represented in this rating?} & In the corrosion 5 category, the surface would show moderate levels of rust and deterioration, with noticeable but not extreme material damage, as the underlying metal remains largely intact, reflecting a midpoint in the corrosion scale.
\\

\midrule
\textcolor{black}{What is the dominant feature in the image?} & The dominant feature in the image is the crossed pattern of corrosion, characterized by two dark, corroded streaks forming an "X" shape across a lighter, perhaps more oxidized or corroded background. This pattern indicates areas of more severe corrosion within the panel, providing a clear contrast to the surrounding material and highlighting the progression of deterioration.
\\

\midrule
\textcolor{black}{Describe the color and texture of the X-shape?} & The ``X" shape in the image appears to be a darker shade, suggesting a concentration of corrosion or an area where the material has been more significantly affected. The texture looks rough and uneven, indicative of material degradation typically seen with rust and corrosion. This contrasts with the lighter, possibly more uniformly corroded background, which might appear smoother or less pitted in comparison.
\\

\midrule
\textcolor{black}{What's notable about the background surrounding the X-shape?} & The lighter-colored background surrounding the ``X" shape suggests less severe corrosion compared to the darker, more corroded ``X" itself. The mottled texture and areas of discoloration indicate uneven surface degradation and the early stages of rust and corrosion.
\\
 
\midrule
\textcolor{black}{Are there visible boundaries or divisions between the X-shape and the background?} & The boundaries between the ``X" shape and the background are distinct, with the darker ``X" sharply contrasting against the lighter, less corroded background, indicating a clear division in the corrosion's progression.
\\
 
\midrule
\textcolor{black}{Is there a pattern to the corrosion emanating from the X-shape?} & There is no discernible pattern to the corrosion emanating from the ``X" shape; it appears contained within its boundaries.
\\
 
\midrule
\textcolor{black}{How do the dark speckles on the background contribute to the overall appearance of corrosion?} & The dark speckles on the background add to the overall appearance of corrosion by indicating early stages of deterioration and providing a contrast to the more severely corroded ``X" shape.
\\
 
\midrule
\textcolor{black}{Given the context of corrosion ratings, what does a rating of 5 signify in terms of material integrity and longevity?} & A rating of 5 indicates moderate corrosion, where the material has undergone noticeable deterioration but still maintains significant integrity, though its longevity may be compromised if left unaddressed.
\\[-0.75em]

\end{tabularx}
\end{tcolorbox}
\end{adjustwidth}

\begin{adjustwidth}{-0.15cm}{-0.15cm} 
\begin{tcolorbox}[colback=white!5!white,colframe=black!75!black]% [width=10cm]
\vspace{-5mm}
\begin{tabularx}{1.0\textwidth}{bss}
\caption{The table shows question-answer pairs about a microscopic image of the corrosion 9 category from the CMI dataset, created with GPT-4 Turbo with vision. In summary, GPT-4 helps generate detailed instruction-tuning dataset about these microscopic images for training smaller multimodal models(SMMs), to serve as a domain-specific vision-language assistant for enterprise adoption.}\label{tab:tab12}\\ [-0.85em]
\toprule
\begin{overpic}[width=0.15\columnwidth]{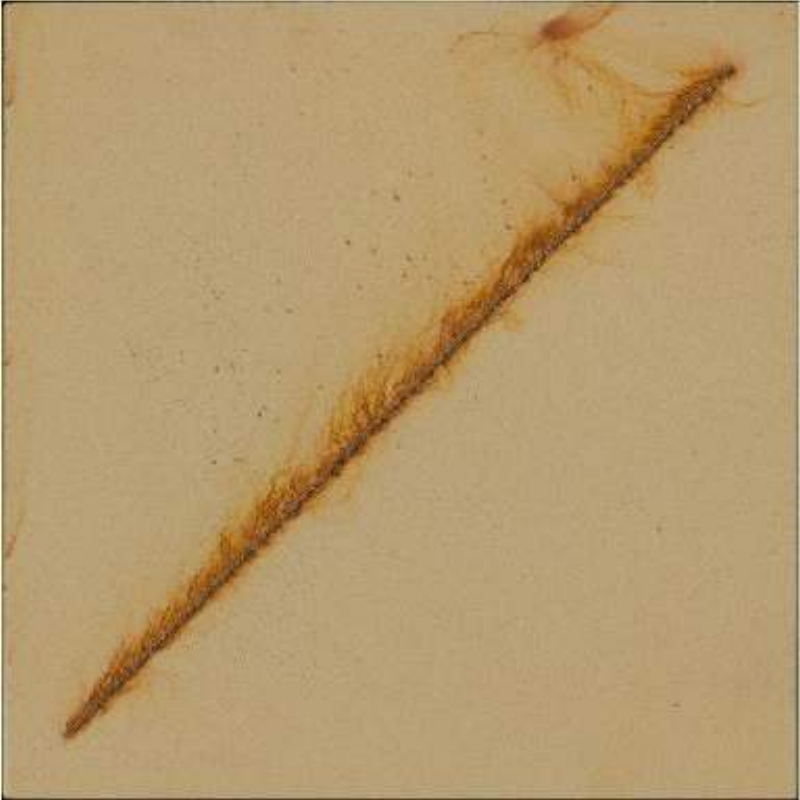}
\vspace{-5mm}
  \put(450,50){\makebox[0pt][c]{%
    \parbox{0.65\columnwidth}{%
      \centering
      \textbf{Examine the provided input image and answer the following questions about the corrosion 9 category it depicts.} 
    }%
  }}  
\end{overpic}  \\

\midrule
\textbf{Query prompt}  & \textbf{GPT-4 Turbo with vision output} \\
\midrule
\endhead
\textcolor{black}{What is the primary intent of the image?} & This image provides a detailed visual depiction of corrosion damage that has been categorized as Level 9 based on standardized criteria for assessing and cataloging corrosion severity.
\\

\midrule
\textcolor{black}{Can you describe the overall color scheme of the image?} & The image displays a color scheme dominated by shades of brown against a cream or light tan background, which is typical for visualizing the rust and corrosion on metal surfaces under magnification.
\\

\midrule
\textcolor{black}{What is the main corroded feature present in the image?} & The main corroded feature present in the image appears to be a linear, rust-like mark indicative of corrosion, often seen as a result of corrosive processes on metal surfaces. It shows a concentrated area of material degradation with color changes that suggest rusting or pitting, which are common signs of corrosion damage.
\\

\midrule
\textcolor{black}{How does the corrosion manifest around this scribed line?} & The corrosion around the scribed line manifests as a concentrated area of discoloration, likely indicating rust, which appears to have spread from the line itself. The edges of the line show some feathering, suggesting that the corrosion has progressed from the initial point of damage into the surrounding material. This pattern is indicative of a corrosive process where the protective layer has been breached, allowing environmental factors to attack the exposed metal, leading to progressive degradation.
\\

\midrule
\textcolor{black}{Given the context of corrosion ratings, how does a rating of 9 compare to other ratings in terms of corrosion severity?} & In the context of corrosion ratings, a rating of 9 typically indicates minimal corrosion, as the scale often runs from 0 (severe corrosion) to 10 (no corrosion). Therefore, a rating of 9 would suggest only slight corrosion, with the substrate mostly intact and little to no rust or deterioration present.
\\
 
\midrule
\textcolor{black}{Is the corrosion consistent throughout the scribed line?} & The corrosion does not appear to be uniform throughout the scribed line. The variation in color intensity and spread indicates that some areas along the line are more affected than others, showing signs of irregular and uneven corrosion progression.
\\
 
\midrule
\textcolor{black}{How does the background of the image appear in comparison to the corroded area?} & The background of the image, in comparison to the corroded area, appears to be relatively uniform and intact, with a light tan or cream color. This contrasts with the corroded area, which shows a darker, brownish discoloration along the scribed line, highlighting the presence and impact of the corrosion.
\\
 
\midrule
\textcolor{black}{If one were to run their fingers across the image, what differences in texture might they expect between the corroded scribe and the background?} & If one were to run their fingers across the image, the corroded scribe would likely feel rougher and more irregular compared to the background. The background, appearing smooth and intact, would offer a contrastingly even surface. The corroded area, due to its degradation and potential pitting or flaking, would give a tactile sense of unevenness and texture variation.
\\[-0.75em]

\end{tabularx}
\end{tcolorbox}
\end{adjustwidth}

\section{KTH-TIPS}
\begin{adjustwidth}{-0.15cm}{-0.15cm} 
\begin{tcolorbox}[colback=white!5!white,colframe=black!75!black]% [width=10cm]
\vspace{-5mm}
\begin{tabularx}{1.0\textwidth}{bss}
\caption{The table shows question-answer pairs about a microscopic image of the corduroy category from the KTH-Tips dataset, created with GPT-4 Turbo with vision. In summary, GPT-4 helps generate detailed instruction-tuning dataset about these microscopic images for training smaller multimodal models(SMMs), to serve as a domain-specific vision-language assistant for enterprise adoption.}\label{tab:tab14}\\ [-0.75em]
\toprule
\begin{overpic}[width=0.15\columnwidth]{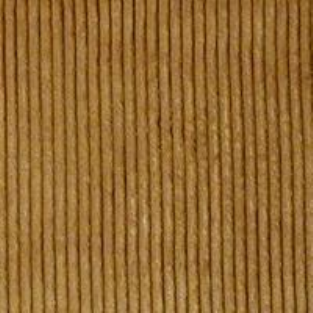}
\vspace{-5mm}
  \put(450,50){\makebox[0pt][c]{%
    \parbox{0.65\columnwidth}{%
      \centering
      \textbf{Examine the provided input image and answer the following questions about the  corduroy category it depicts.} 
    }%
  }}  
\end{overpic}  \\

\midrule
\textbf{Query prompt}  & \textbf{GPT-4 Turbo with vision output} \\
\midrule
\endhead
\textcolor{black}{What type of material is depicted in the image?} & The image you've provided appears to depict corduroy, which is a textile with a distinctive pattern, characterized by its raised ridges, or ``wales", that run in parallel lines. Corduroy is typically made from tufted cords, sometimes exhibiting a channel (bare to the base fabric) between the tufts. The width of the wales varies and is often used to distinguish different types of corduroy. It's a durable fabric and is commonly used for clothing and upholstery.
\\

\midrule
\textcolor{black}{What color does the corduroy fabric primarily exhibit?} & The image depicts corduroy fabric with a tan or beige color. This is a common color for corduroy, making it versatile for various applications in fashion and furnishings.
\\

\midrule
\textcolor{black}{Describe the texture and pattern noticeable on the corduroy?} & Corduroy fabric has a distinctive texture of raised parallel lines called wales that run the length of the material, creating a soft, plush, and velvety feel with visual and tactile interest.
\\

\midrule
\textcolor{black}{Are these raised lines uniformly distributed throughout the fabric?} & Yes, in the image of the corduroy fabric, the raised lines, or wales, appear to be uniformly distributed throughout, maintaining consistent spacing and parallel alignment, which is typical of corduroy.
\\

\midrule
\textcolor{black}{How does light interact with the fabric's texture?} & The light on the fabric accentuates the corduroy's ribbed texture, creating highlights and shadows along the parallel wales. This interplay showcases the material's soft sheen on the raised ridges, while the grooves appear darker from cast shadows. The contrast showcases the tactile quality.
\\
 
\midrule
\textcolor{black}{Describe the spacing between the ribs?} & In the provided image, the spacing between the ribs of the corduroy fabric appears regular and even, consistent with the characteristic look of corduroy. Each rib is separated by a narrow groove, which is uniform throughout, giving the fabric its distinctive textured pattern.
\\
 
\midrule
\textcolor{black}{How does the fabric's texture vary between the raised lines and the gaps in between?} &  The corduroy fabric has a textured surface with pronounced, plush wales separated by flatter, smoother gaps, creating a distinctive tactile and visual quality.
\\
 
\midrule
\textcolor{black}{Are there any visible imperfections or signs of wear on the corduroy?} &  Based on the image provided, there are no visible imperfections or signs of wear on the corduroy. The fabric appears to be in good condition, with the texture and color uniformly consistent across the surface.
\\
 
\midrule
\textcolor{black}{From which angle or perspective is the fabric being observed?} &  The fabric is observed from a close-up, top-down perspective, allowing a detailed view of the corduroy's texture and pattern.
\\ 

\midrule
\textcolor{black}{Is the image a full representation of the fabric or just a section?} & 
The image appears to capture just a section of the corduroy fabric, focusing on its texture and pattern in detail rather than showing the entire piece.
\\[-0.75em]

\end{tabularx}
\end{tcolorbox}
\end{adjustwidth}

\begin{adjustwidth}{-0.15cm}{-0.15cm} 
\begin{tcolorbox}[colback=white!5!white,colframe=black!75!black]% [width=10cm]
\vspace{-5mm}
\begin{tabularx}{1.0\textwidth}{bss}
\caption{The table shows question-answer pairs about a microscopic image of the linen category from the KTH-Tips dataset, created with GPT-4 Turbo with vision. In summary, GPT-4 helps generate detailed instruction-tuning dataset about these microscopic images for training smaller multimodal models(SMMs), to serve as a domain-specific vision-language assistant for enterprise adoption.}\label{tab:tab15}
\\ [-0.75em]
\toprule
\begin{overpic}[width=0.15\columnwidth]{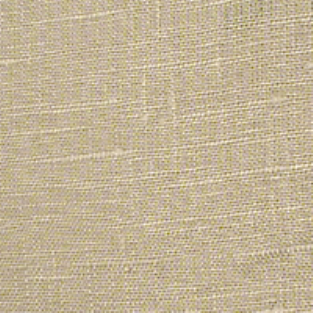}
\vspace{-5mm}
  \put(450,50){\makebox[0pt][c]{%
    \parbox{0.65\columnwidth}{%
      \centering
      \textbf{Examine the provided input image and answer the following questions about the linen category it depicts.} 
    }%
  }}  
\end{overpic}  \\

\midrule
\textbf{Query prompt}  & \textbf{GPT-4 Turbo with vision output} \\
\midrule
\endhead
\textcolor{black}{What is the primary material displayed in the image?} & The primary material displayed in the image appears to be a woven fabric with a relatively uniform and tight weave, which is characteristic of linen. Linen is known for its distinct, slightly textured weave, which is visible in the image. The even, crisscross pattern typical of linen is created by the plain weave, or 'linen weave', where each weft thread crosses over and under each warp thread.
\\

\midrule
\textcolor{black}{Can you identify the primary color of the linen fabric in the image?} &  The primary color of the linen fabric in the image is a neutral, beige tone. This color is typical for linen and is often associated with its natural, unbleached state. The variations in the color, such as the slight differences in shading and the presence of darker and lighter areas, contribute to the textured appearance of the fabric.
\\

\midrule
\textcolor{black}{Describe the texture or pattern visible on the linen?} & The linen fabric has a tight, slightly irregular weave pattern with natural variations in thread thickness, creating a distinctive tactile texture and a rich, dimensional appearance.
\\

\midrule
\textcolor{black}{How would you describe the spacing between the woven threads?} & The spacing between the woven threads appears to be consistent and tight, indicative of a high-quality linen with a fine weave.
\\

\midrule
\textcolor{black}{Does the linen appear soft or rigid based on the visible texture?} & Based on the visible texture, the linen appears to have a soft, supple texture rather than a rigid one, as indicated by the slight natural irregularities and the drape that can be inferred from the flow of the weave.
\\
 
\midrule
\textcolor{black}{Are there any observable sheen or reflective properties on the linen?} &  In the provided image, the linen does not exhibit a pronounced sheen or reflective properties. The fabric appears matte, which is typical for linen, known for its natural, lusterless finish.
\\
 
\midrule
\textcolor{black}{How consistent is the texture throughout the displayed section?} & The texture appears consistent throughout the displayed section, with a uniform weave pattern and even spacing, characteristic of high-quality linen fabric. There are no significant variations or defects visible, suggesting a homogeneous material quality across the section shown.
\\
 
\midrule
\textcolor{black}{ Based on the visible attributes, how would you describe the tactile sensation of the fabric?} &  Based on the visible attributes, the fabric likely has a crisp yet supple tactile sensation, with a textured surface that could provide a slight roughness to the touch due to the natural variations in the weave. However, it would not be excessively coarse; linen typically feels cool and comfortable against the skin.
\\
 
\midrule
\textcolor{black}{Are there any discernible imperfections or unique characteristics in the fabric?} &  From the image, the linen fabric appears quite uniform with no significant imperfections. It displays the natural and subtle variations in thread size that are characteristic of linen, which give it a unique texture. These are not flaws but rather inherent qualities of the fabric that enhance its natural aesthetic.
\\[-0.75em]

\end{tabularx}
\end{tcolorbox}
\end{adjustwidth}

\section{NEU-SDD}
\begin{adjustwidth}{-0.15cm}{-0.15cm} 
\begin{tcolorbox}[colback=white!5!white,colframe=black!75!black]% [width=10cm]
\vspace{-5mm}
\begin{tabularx}{1.0\textwidth}{bss}
\caption{The table displays question-answer pairs related to microscopic images of the scratches category from the NEU-SDD dataset within the field of defect detection, especially for surface defects. These pairs were created using GPT-4 Turbo with vision. In summary, GPT-4 assists in generating a detailed instruction-tuning dataset for training smaller multimodal models (SMMs) to serve as domain-specific vision-language assistants for enterprise adoption.}
\label{tab:tab16}\\ [-0.75em]
\toprule
\begin{overpic}[width=0.15\columnwidth]{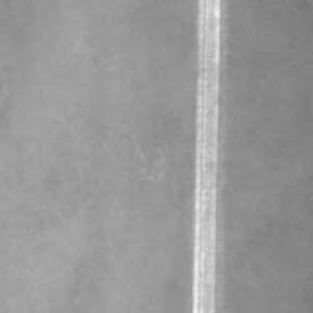}
\vspace{-5mm}
  \put(450,50){\makebox[0pt][c]{%
    \parbox{0.65\columnwidth}{%
      \centering
      \textbf{Examine the provided input image and answer the following questions about the scratches category it depicts.} 
    }%
  }}  
\end{overpic}  \\

\midrule
\textbf{Query prompt}  & \textbf{GPT-4 Turbo with vision output} \\
\midrule
\endhead
\textcolor{black}{What category does this texture belong to?} & The texture in the image belongs to the scratches category, which is characterized by linear grooves or marks on the surface of hot-rolled steel plates.
\\

\midrule
\textcolor{black}{Describe the central feature in the image?} & The main feature is a linear scratch on the material's surface. It is a distinct, narrow, continuous indentation contrasting the surrounding area. Running vertically with a uniform width, it is lighter than the base material, revealing variances in depth or texture. Such defects typically result from mechanical abrasion or contact with sharp, hard objects.
\\

\midrule
\textcolor{black}{How would you describe the contrast between the scratch and the surrounding texture?} & The bright, linear scratch contrasts significantly with the uniform, granular, and darker surrounding texture, making the defect stand out.
\\

\midrule
\textcolor{black}{What color tones dominate the image?} & A grayscale image with shades of gray ranging from black to white. The background is a mid-gray tone while the scratch is represented by a lighter, almost white shade indicating higher reflectance. No color hues since it's not a color image; the gray tone variations provide the visual information.
\\

\midrule
\textcolor{black}{Can you define the orientation of the scratch?} & The scratch has a vertical orientation, running top to bottom with a consistent linear path. It creates distinct contrast against the more textured background. Its alignment is vertical relative to the image edges
\\
 
\midrule
\textcolor{black}{How would you describe the edges of the scratch?} & The edges of the scratch are straight and clearly defined, creating a distinct contrast with the surrounding surface texture. They appear abrupt rather than gradual, suggesting a sudden change in surface texture and depth.
\\
 
\midrule
\textcolor{black}{Is the scratch in the center or towards the side of the image?} & The scratch is positioned towards the center of the image, running vertically and occupying a central location within the frame of view.
\\
 
\midrule
\textcolor{black}{Are there multiple scratches or just a single prominent one?} & Based on the uploaded image, there is just a single prominent scratch visible running vertically near the center of the image. There are no other distinct scratches observable in the grayscale image provided.
\\
 
\midrule
\textcolor{black}{Is the rest of the image's texture smooth or grainy?} & The rest of the image's texture is grainy, with a speckled pattern that indicates a rough or irregular surface characteristic of the material under observation.
\\ 

\midrule
\textcolor{black}{How does the brightness of the scratch compare to the overall image?} & The scratch is notably brighter than the overall image. It stands out as a light, almost white line against the darker, grainier background of the rest of the image, indicating a significant contrast in brightness.
\\

\midrule
\textcolor{black}{If you were to touch the scratch, would it feel raised or recessed?} & The brighter appearance of the scratch in the grayscale image suggests it is likely a recessed, indented area compared to the surrounding surface.
\\[-0.75em]

\end{tabularx}
\end{tcolorbox}
\end{adjustwidth}

\begin{adjustwidth}{-0.15cm}{-0.15cm} 
\begin{tcolorbox}[colback=white!5!white,colframe=black!75!black]% [width=10cm]
\vspace{-5mm}
\begin{tabularx}{1.0\textwidth}{bss}
\caption{The table displays question-answer pairs related to a microscopic image from the patches category within the NEU-SDD dataset, generated using GPT-4 Turbo with vision. In summary, GPT-4 assists in producing a detailed instruction-tuning dataset for training smaller multimodal models (SMMs), aiming to serve as a domain-specific vision-language assistant for enterprise adoption.}\label{tab:tab17}\\ [-0.75em]
\toprule
\begin{overpic}[width=0.15\columnwidth]{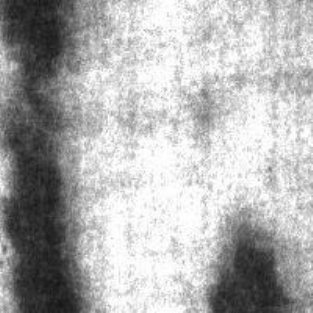}
\vspace{-5mm}
  \put(400,50){\makebox[0pt][c]{%
    \parbox{0.65\columnwidth}{%
      \centering
      \textbf{Examine the provided input image and answer the following questions about the patches category it depicts.} 
    }%
  }}  
\end{overpic}  \\

\midrule
\textbf{Query prompt}  & \textbf{GPT-4 Turbo with vision output} \\
\midrule
\endhead
\textcolor{black}{What category does this texture belong to?} & This texture appears to be a type of surface defect and it matches the characteristics of patches category. Typically, patches in the context of surface defects on hot-rolled steel plates are characterized by areas where the surface has irregularities that are distinct from the base material. These can be due to various factors such as contamination, uneven cooling, or material inconsistencies during the manufacturing process.
\\

\midrule
\textcolor{black}{Describe the dominant features in the image?} & The image is a grayscale representation of a surface defect on a steel plate, likely falling under the 'patches' category. It exhibits varied gray intensity, non-uniform and irregular patterns, and blurred or distinct edges. The shape and distribution of these features are crucial for classification, but a precise description cannot be provided without visual access to the image.
\\

\midrule
\textcolor{black}{How would you characterize the contrast between the patches and the surrounding area?} & The contrast in the image likely shows patches as darker or lighter areas against the base metal, with potentially sharp or blurred boundaries that make them stand out from the surrounding surface.
\\

\midrule
\textcolor{black}{What color tones are prevalent in the image?} & The image is grayscale, featuring shades from black to white, without any color tones.
\\

\midrule
\textcolor{black}{Describe the shape and arrangement of the patches?} & This grayscale image from the NEU-SDD dataset showcases surface defects categorized as `patches' on hot-rolled steel. The monochromatic view displays varying shades of gray, indicating textural differences on the material. The irregular patches appear randomly distributed without regular patterns. The grayscale nuances denote surface irregularities; precise categorization would require material science expertise and analysis.
\\
 
\midrule
\textcolor{black}{How does the texture within the patches compare to the rest of the image?} & The provided image likely depicts patches as areas with distinct textures and contrasts compared to the rest of the surface, indicative of material defects which require specialized analysis for accurate classification.
\\
 
\midrule
\textcolor{black}{Would you say the patches are centralized or more dispersed?} & The patches in the image are not clearly discernible with the given resolution, making it difficult to determine if they are centralized or dispersed.
\\
 
\midrule
\textcolor{black}{Do the patches have well-defined boundaries?} &  The image provided is too low in resolution and lacks clear definition, making it challenging to determine if the patches have well-defined boundaries. For a precise assessment, a higher-resolution image and specialized analysis would be required.
\\
 
\midrule
\textcolor{black}{Describe the granularity of the background texture?} & The background texture in the image appears to be quite fine and consistent, suggesting a high degree of granularity that is typical for grayscale images representing material surfaces. This granularity could be indicative of the surface quality or the resolution of the imaging technique used.
\\ 

\midrule
\textcolor{black}{In a real-world scenario, how might such patches form on materials?} & Patches on materials may form due to corrosion, wear, inconsistent heat treatment, coating failures, material defects, repair work, or environmental damage, affecting the material's integrity and performance.
\\

\midrule
\textcolor{black}{Does the image lean more towards being bright, dark, or neutral overall?} & The provided image appears to be neutral overall, with a balanced distribution of both light and dark areas, without leaning significantly towards being overly bright or dark.
\\[-0.75em]

\end{tabularx}
\end{tcolorbox}
\end{adjustwidth}

\begin{table*}[!htb]
    \caption{The table presents a sample of electron microscope images, along with their corresponding framework-generated and ground-truth answers, and similarity evaluation metrics for an open-ended VQA task. The task focuses on identifying specific material surface defects and their impact on performance. We evaluate the quality of the machine-generated answers using evaluation metrics such as BLEU-2, ROUGE-L, and METEOR to measure their similarity to the ground-truth answers.   
    }
     \vspace{3mm}
      \centering 
       \begin{tabular}{|>{\centering\arraybackslash}m{2cm}|m{5cm}|m{5cm}|m{1.75cm}|}
        \hline 
        Image & Ground Truth &  Framework-Generated Answer & BLEU-2 /  \ ROGUE-L / \ METEOR \\ \hline
        \includegraphics[width=2cm,height=2cm,keepaspectratio]{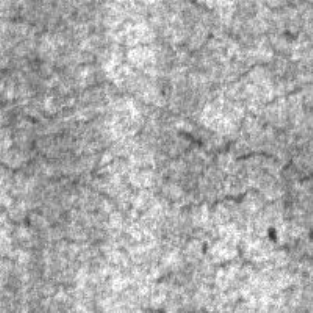} & The image shows a surface with crazing, which can compromise material strength and lead to premature failure under stress. & The image depicts a surface with crazing, which could weaken the material strength and lead to early failure under stress & \begin{tabular}[c]{@{}c@{}}0.659\\ 0.769\\ 0.831\end{tabular} \\ \hline
        \includegraphics[width=2cm,height=2cm,keepaspectratio]{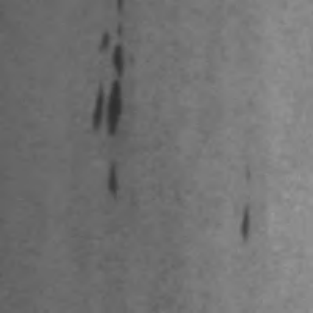} & The image likely shows a material defect known as inclusion, which can create weak points and reduce the uniformity and strength of the material. & The photo probably displays an imperfection referred to as an inclusion, which can make fragile spots and decrease the consistency and robustness of the material. & \begin{tabular}[c]{@{}c@{}}0.271\\ 0.449\\ 0.504\end{tabular}  \\ \hline
        \includegraphics[width=2cm,height=2cm,keepaspectratio]{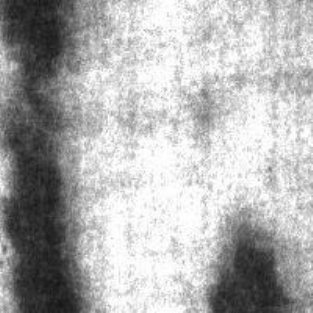} & The image depicts a defect known as patches, which can lead to heterogeneous material properties and potentially reduce the material's overall durability and performance. & The photo exhibits a flaw termed patches, which could result in nonuniform material traits and perhaps reduce the overall toughness and function of the material. & \begin{tabular}[c]{@{}c@{}}0.183\\ 0.400\\ 0.354\end{tabular} \\ \hline
        \includegraphics[width=2cm,height=2cm,keepaspectratio]{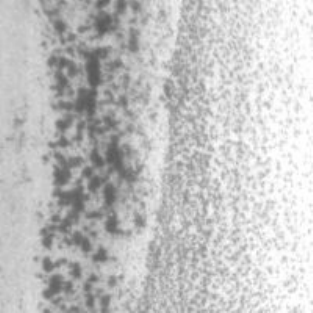} & The image appears to show a clustering of material defects known as "pitting," which can compromise the material's integrity and lead to failure through stress concentration. & The image likely exhibits clustering of flaws called pitting, which could undermine the material's soundness and cause failure via concentration of tension. & \begin{tabular}[c]{@{}c@{}}0.245\\ 0.480\\ 0.370\end{tabular}  \\ \hline
        \includegraphics[width=2cm,height=2cm,keepaspectratio]{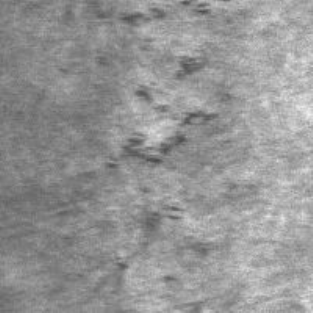} & The image seems to exhibit rolled-in scale, a defect which can create weak spots and reduce a material's overall strength and fatigue resistance. & The image appears to show rolled-in scale, a flaw that can generate weak areas and lessen the complete durability and wear tolerance of a material. & \begin{tabular}[c]{@{}c@{}}0.235\\ 0.471\\ 0.566\end{tabular}  \\ \hline
        \includegraphics[width=2cm,height=2cm,keepaspectratio]{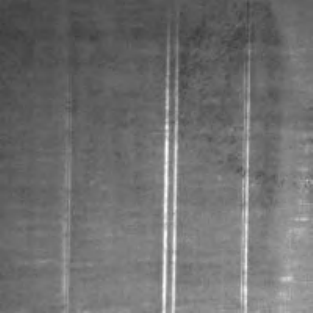} & The image displays scratches, a surface defect that can act as stress concentrators, potentially leading to reduced fatigue life and increased susceptibility to crack propagation. & The image shows scratches, a surface imperfection which could act as stress concentrators, potentially causing shortened fatigue life and magnified vulnerability to crack spreading. &\begin{tabular}[c]{@{}c@{}}0.500\\ 0.612\\ 0.638\end{tabular} \\ \hline
        \end{tabular}    
        \label{tab:onvqa1}
\end{table*}

\begin{table*}[!htb]
    \caption{The table showcases sample electron microscope images with corresponding framework-generated and ground-truth answers for surface texture analysis (smoothness, roughness, imperfections). We evaluate generated answer quality using similarity metrics (BLEU-2, ROUGE-L, METEOR) to ground truth.}
     \vspace{3mm}
      \centering 
         \begin{tabular}{|>{\centering\arraybackslash}m{2cm}|m{5cm}|m{5cm}|m{1.75cm}|}
        \hline 
        Image & Ground Truth & Framework-Generated Answer & BLEU-2 /\ ROGUE-L /\ METEOR \\ \hline
        \includegraphics[width=2cm,height=1.75cm,keepaspectratio]{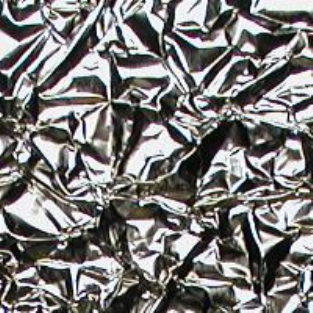} & The material is crinkled and reflective, with a rough, uneven texture and no significant imperfections. & The material appears crumpled and shiny, displaying a jagged, uneven surface without any major imperfections visible. & \begin{tabular}[c]{@{}c@{}}0.144\\ 0.387\\ 0.246\end{tabular} \\ \hline
        \includegraphics[width=2cm,height=1.75cm,keepaspectratio]{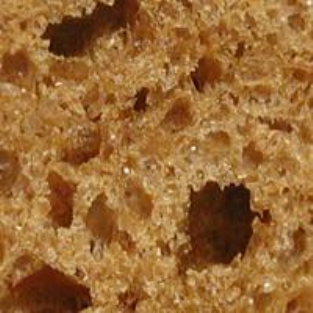} & The material is porous with a rough texture, featuring many small cavities and no obvious imperfections. & The substance has a rough texture and is full of tiny pores, with no visible flaws. & \begin{tabular}[c]{@{}c@{}}0.171\\ 0.375\\ 0.300\end{tabular}  \\ \hline
        \includegraphics[width=2cm,height=1.75cm,keepaspectratio]{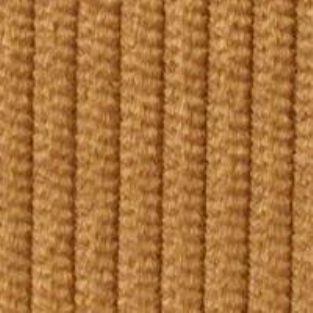} & The material has a textured, ribbed surface with a consistent pattern, indicative of a woven or knitted fabric, likely rough to the touch, with no discernible imperfections. & The material features a textured, ribbed surface with a uniform pattern, suggesting a woven or knitted fabric, likely rough to touch, and free of noticeable imperfections & \begin{tabular}[c]{@{}c@{}}0.593\\ 0.717\\ 0.669\end{tabular}  \\ \hline
        \includegraphics[width=2cm,height=1.75cm,keepaspectratio]{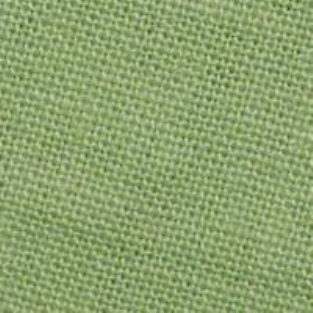} & The material has a fine, woven texture, likely with a slight roughness due to its knit, and appears uniform without visible imperfections. & The material features a fine, woven texture, likely displaying a slight roughness due to its knit, and exhibits a uniform appearance without noticeable imperfections. & \begin{tabular}[c]{@{}c@{}}0.626\\ 0.735\\ 0.810\end{tabular}  \\ \hline
        \includegraphics[width=2cm,height=1.75cm,keepaspectratio]{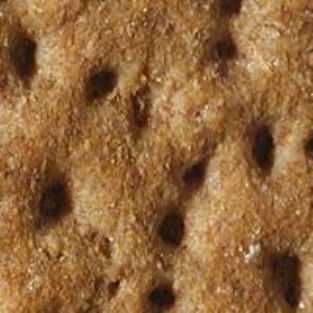} & The material is porous with a grainy texture, featuring holes and a coarse surface, with no notable imperfections aside from its natural texture. & The material exhibits a porous structure with a grainy texture, characterized by holes and a rough surface, having no significant imperfections other than its inherent texture. & \begin{tabular}[c]{@{}c@{}}0.372\\ 0.783\\ 0.548\end{tabular}  \\ \hline
        \includegraphics[width=2cm,height=1.75cm,keepaspectratio]{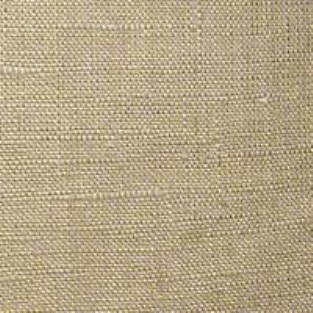} & The material has a fine, tightly woven texture with a uniform appearance, likely smooth to the touch, and shows no apparent imperfections. & The material displays a fine, tightly woven texture with a consistent appearance, likely feeling smooth to the touch, and exhibits no discernible imperfections. & \begin{tabular}[c]{@{}c@{}}0.653\\ 0.612\\ 0.852\end{tabular} \\ \hline
        \includegraphics[width=2cm,height=1.75cm,keepaspectratio]{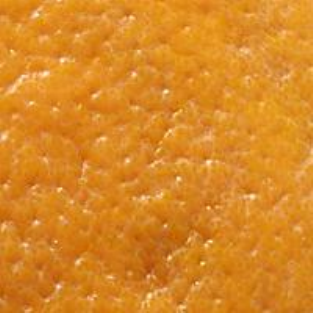} & The material has a bubbled, uneven texture that suggests roughness, with a glossy sheen and no obvious imperfections beyond its inherent textured surface. & The material exhibits a bubbled, uneven texture, indicating roughness, featuring a glossy finish, and displaying no noticeable imperfections beyond its inherent textured surface. & \begin{tabular}[c]{@{}c@{}}0.533\\ 0.800\\ 0.667\end{tabular}  \\ \hline
        \includegraphics[width=2cm,height=1.75cm,keepaspectratio]{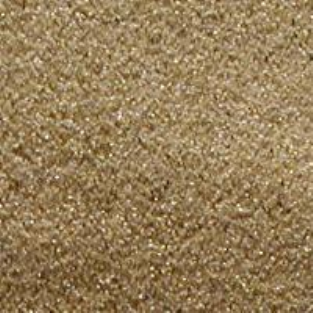} & The material has a granular, sandy texture that would likely be rough to the touch, with a speckled appearance and no significant imperfections visible. & The material displays a granular, sandy texture, likely feeling rough to the touch, featuring a speckled pattern, and showing no significant imperfections. & \begin{tabular}[c]{@{}c@{}}0.465\\ 0.739\\ 0.598\end{tabular} \\ \hline
        \includegraphics[width=2cm,height=1.75cm,keepaspectratio]{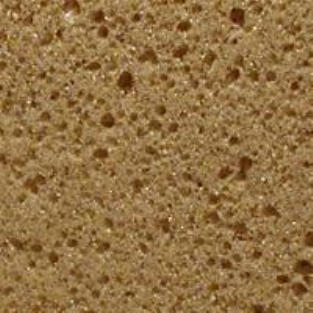} & The surface is porous and spongy with a rough texture, featuring numerous small holes and a coarse feel, without specific imperfections beyond its natural structure. & The surface exhibits a porous and spongy texture, feeling rough, with numerous tiny holes, and a coarse touch, displaying no specific imperfections beyond its natural structure. & \begin{tabular}[c]{@{}c@{}}0.749\\ 0.739\\ 0.686\end{tabular}  \\ \hline
        \includegraphics[width=2cm,height=1.75cm,keepaspectratio]{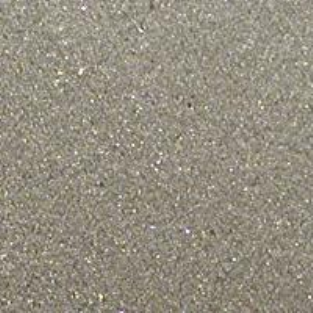} & The surface has a fine, gritty texture with a subtle sparkle, suggesting a slightly rough feel, and shows a uniform appearance with no clear imperfections. & The surface displays a fine, gritty texture with a slight sparkle, implying a somewhat rough touch, and exhibits a uniform look with no apparent imperfections. & \begin{tabular}[c]{@{}c@{}}0.462\\ 0.739\\ 0.707\end{tabular} \\ \hline
        \end{tabular}    
        \label{tab:onvqa2}
\end{table*}

\begin{table*}[!htb]
    \caption{The table showcases sample electron microscope images used in an open-ended VQA task focused on metal corrosion. It compares framework-generated answers to ground-truth responses, along with metrics like BLEU-2, ROUGE-L, and METEOR to assess their similarity.}
      \centering
      \vspace{3mm}
       \begin{tabular}{|>{\centering\arraybackslash}m{2cm}|m{5cm}|m{5cm}|m{1.75cm}|}
        \hline 
        Image & Ground Truth & Framework-Generated Answer & BLEU-2 /\ ROGUE-L / \ METEOR \\ \hline
        \includegraphics[width=2cm,height=1.75cm,keepaspectratio]{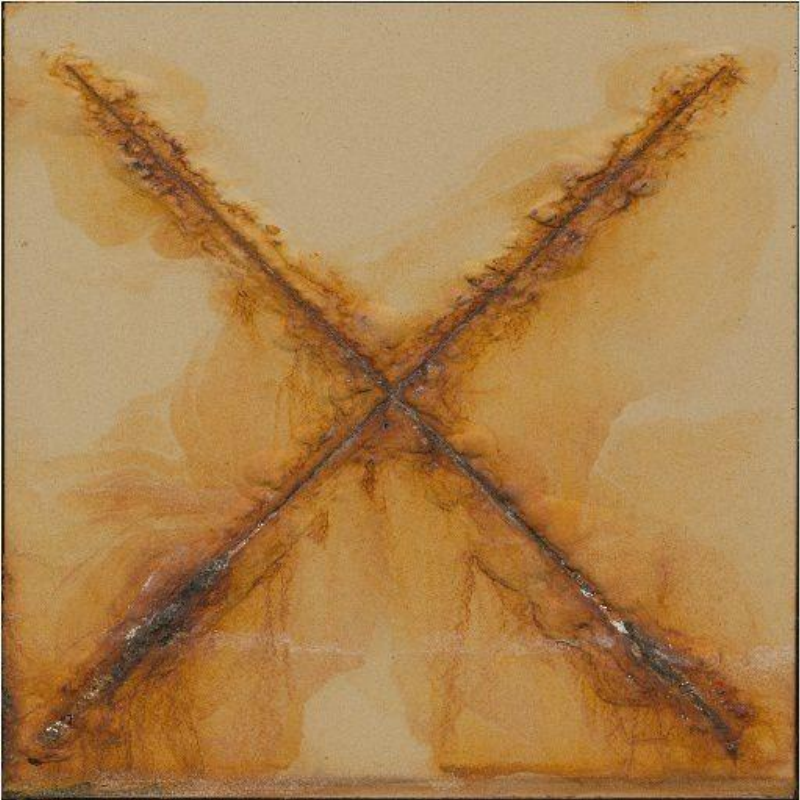} & The extent of corrosion shown in the image suggests advanced corrosion with widespread areas of rust, indicating prolonged exposure to a corrosive environment. & The image reveals extensive corrosion, suggesting severe corrosion with widespread rust, indicating prolonged exposure to a corrosive atmosphere. & \begin{tabular}[c]{@{}c@{}}0.424\\ 0.585\\ 0.569\end{tabular} \\ \hline
        \includegraphics[width=2cm,height=1.75cm,keepaspectratio]{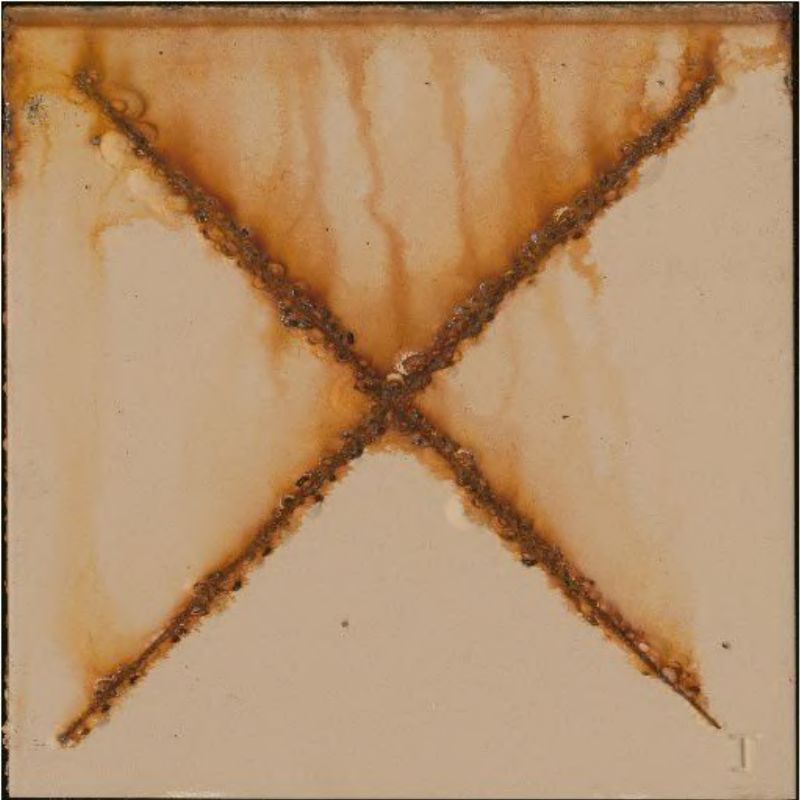} & The image shows localized corrosion concentrated along the metal's edges, with some spread from the central point of contact, indicating moderate corrosion. & Theimage reveals localized corrosion along the metal edges, with some spreading from the central contact point, suggesting moderate corrosion & \begin{tabular}[c]{@{}c@{}}0.491\\ 0.744\\ 0.675\end{tabular}  \\ \hline
        \includegraphics[width=2cm,height=1.75cm,keepaspectratio]{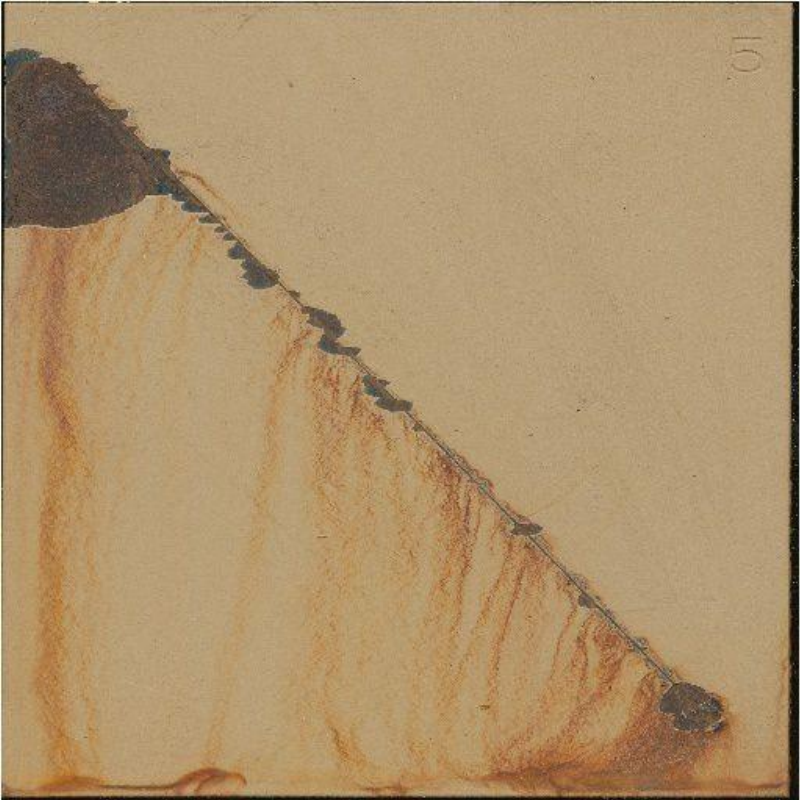} & The corrosion is localized at one corner with streaking patterns, suggesting early to moderate stages of corrosion, likely from moisture exposure. & Corrosion is localized in one corner with streaking patterns, indicating early to moderate corrosion stages, likely due to moisture exposure. & \begin{tabular}[c]{@{}c@{}}0.516\\ 0.732\\ 0.607\end{tabular}  \\ \hline
        \includegraphics[width=2cm,height=1.75cm,keepaspectratio]{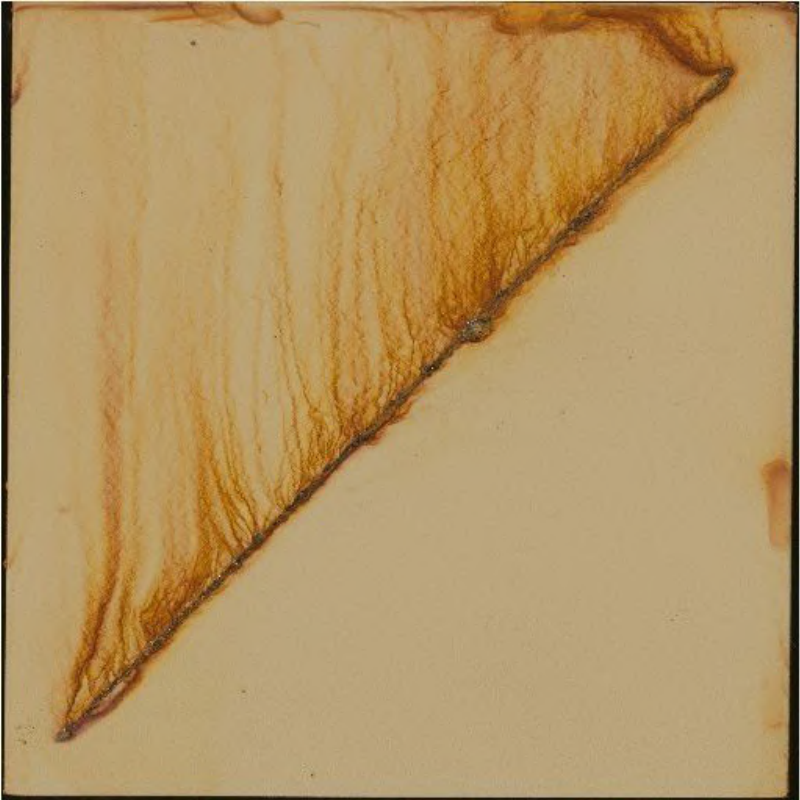} & The corrosion appears as a well-defined edge with some spreading, indicative of moderate progression. & Corrosion appears as a distinct edge with some spread, suggesting moderate progression. & \begin{tabular}[c]{@{}c@{}}0.466\\ 0.667\\ 0.640\end{tabular}  \\ \hline
        \includegraphics[width=2cm,height=1.75cm,keepaspectratio]{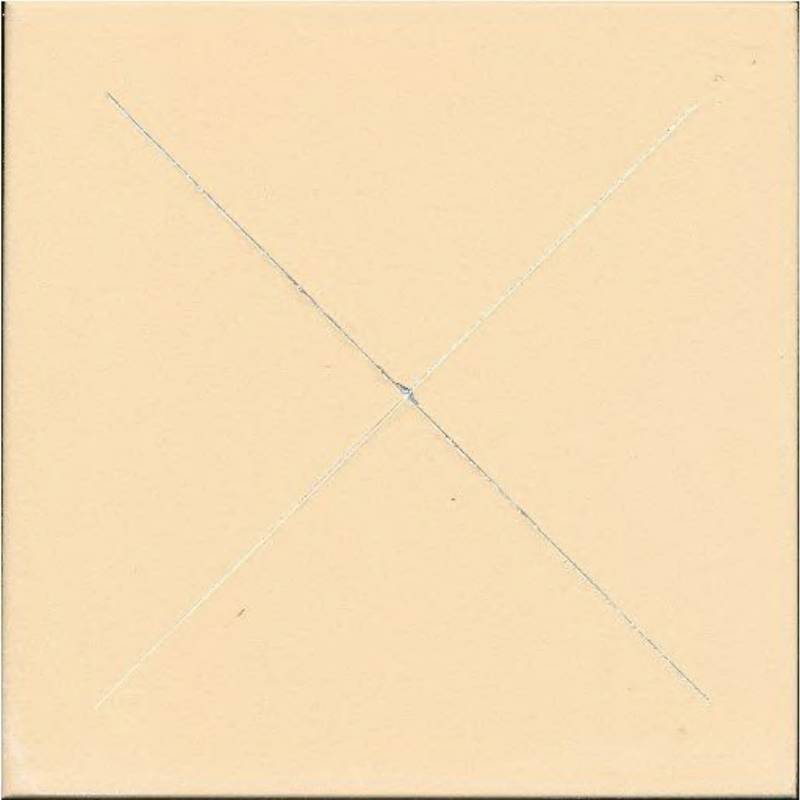} & There's minimal to no visible corrosion on the metal in this image, indicating it's in good condition or the coating is effective. & There's little to no visible corrosion on the metal in this picture, suggesting it's in good shape or the coating works well. & \begin{tabular}[c]{@{}c@{}}0.645\\ 0.750\\ 0.722\end{tabular}  \\ \hline
        \end{tabular}   
        \label{tab:onvqa3} 
\end{table*}

\end{document}